\documentclass{article}

\usepackage{arxiv}

\usepackage[utf8]{inputenc}
\usepackage[T1]{fontenc}
\usepackage{hyperref}
\usepackage{url}
\usepackage{booktabs}
\usepackage{amsfonts}
\usepackage{amsmath,amssymb}
\usepackage{nicefrac}
\usepackage{microtype}
\usepackage{graphicx}
\usepackage[numbers,sort&compress]{natbib}
\usepackage{algorithm}
\usepackage{algorithmic}
\usepackage{multirow}
\usepackage{array}
\usepackage{lscape}
\usepackage{xcolor}
\definecolor{barcol}{HTML}{4C78A8}       
\usepackage{colortbl}                    
\usepackage{eso-pic}     
\usepackage{calc}        
\usepackage[framemethod=tikz]{mdframed}  
\usepackage{float}                       
\usepackage{subcaption}                  
\usepackage{listings}                    
\definecolor{abstractbg}{RGB}{232, 240, 253}  
\definecolor{abstractcode}{RGB}{180, 50, 70}  
\definecolor{abstractlink}{RGB}{20, 90, 170}  
\definecolor{msblue}{HTML}{0078D4}
\definecolor{msred}{HTML}{F25022}
\definecolor{msgreen}{HTML}{7FBA00}
\definecolor{msyellow}{HTML}{FFB900}
\newcommand{\reelword}{\textcolor{msred}{R}\textcolor{msgreen}{e}\textcolor{msblue}{e}\textcolor{msyellow}{l}}

\definecolor{revision}{RGB}{200,0,0}
\definecolor{revisionb}{RGB}{0,0,200}
\newif\ifshowrevisions
\showrevisionstrue

\renewenvironment{abstract}{}{}

\providecommand{\keywords}[1]{}
\renewcommand{\keywords}[1]{}

\providecommand{\cmark}{\textcolor{teal}{\checkmark}}
\providecommand{\xmark}{\textcolor{red!65!black}{\ensuremath{\times}}}
\providecommand{\dnmark}{\textcolor{gray}{\rule[0.45ex]{1.3ex}{0.13ex}}}

\InputIfFileExists{gallery_runs/table1_cells.real.tex}{}{}
\InputIfFileExists{ablation_runs/abl_cells.tex}{}{}
\title{\LARGE ResearchStudio-\reelword: Automate the Last Mile of Research\\ from Paper to Poster, Video, and Blog}

\makeatletter
\renewcommand{\@toptitlebar}{%
  \noindent\hspace*{-0pt}\rule{\dimexpr\textwidth+0pt\relax}{2pt}%
  \vskip 0.3in
  \vskip -\parskip
}
\renewcommand{\@maketitle}{%
  \vbox{%
    \hsize\textwidth
    \linewidth\hsize
    \vskip 0.1in
    \@toptitlebar
    \centering
    {\LARGE\sffamily\bfseries \@title\par}%
    \vskip 0.15in
    \textsc{\undertitle}%
    \def\And{%
      \end{tabular}\hfil\linebreak[0]\hfil%
      \begin{tabular}[t]{c}\ignorespaces%
    }
    \def\AND{%
      \end{tabular}\hfil\linebreak[4]\hfil%
      \begin{tabular}[t]{c}\ignorespaces%
    }
    \begin{tabular}[t]{c}\ignorespaces\@author\end{tabular}%
    \vskip 0.25in \@minus 0.05in \center{\@date}\vskip 0.15in
  }
}
\makeatother

\author{%
Lingao Xiao\textsuperscript{1,2*} \quad Yalun Dai\textsuperscript{1,3*} \quad Yangyu Huang\textsuperscript{1*\ddag} \quad Qihao Zhao\textsuperscript{3} \quad Wenshan Wu\textsuperscript{1} \quad Hugo He\textsuperscript{$\flat$} \\
Ruishuo Chen\textsuperscript{4} \quad Jin Jiang\textsuperscript{5} \quad Qianli Ma\textsuperscript{6} \quad Jiahuan Zhang\textsuperscript{7} \quad Xin Zhang\textsuperscript{1} \quad Ying Xin\textsuperscript{1} \quad Yang Ou\textsuperscript{1} \\
Yan Xia\textsuperscript{1} \quad Scarlett Li\textsuperscript{1} \quad Longbo Huang\textsuperscript{4} \quad Zhipeng Zhang\textsuperscript{6} \quad Yang He\textsuperscript{2,8\ddag} \quad Yap Kim Hui\textsuperscript{3\ddag} \quad Yan Lu\textsuperscript{1} \\[2pt]
{\small\color{black!55}\textsuperscript{1}Microsoft Research \quad \textsuperscript{2}National University of Singapore \quad \textsuperscript{3}Nanyang Technological University} \\
{\small\color{black!55}\textsuperscript{4}Tsinghua University \quad \textsuperscript{5}Peking University \quad \textsuperscript{6}Shanghai Jiao Tong University \quad \textsuperscript{7}Westlake University \quad \textsuperscript{8}CFAR, A*STAR}%
}

\renewcommand{\headeright}{}
\providecommand{\undertitle}{}
\renewcommand{\undertitle}{}

\begin{document}
\newcommand{\monthyear}{\ifcase\month\or
  January\or February\or March\or April\or May\or June\or
  July\or August\or September\or October\or November\or December\fi
  \space\number\year}
\newlength{\dateboxwidth}
\settowidth{\dateboxwidth}{\normalsize\monthyear}
\AddToShipoutPictureBG*{%
  \AtPageUpperLeft{%
    \put(64pt,-108pt){\includegraphics[height=96pt]{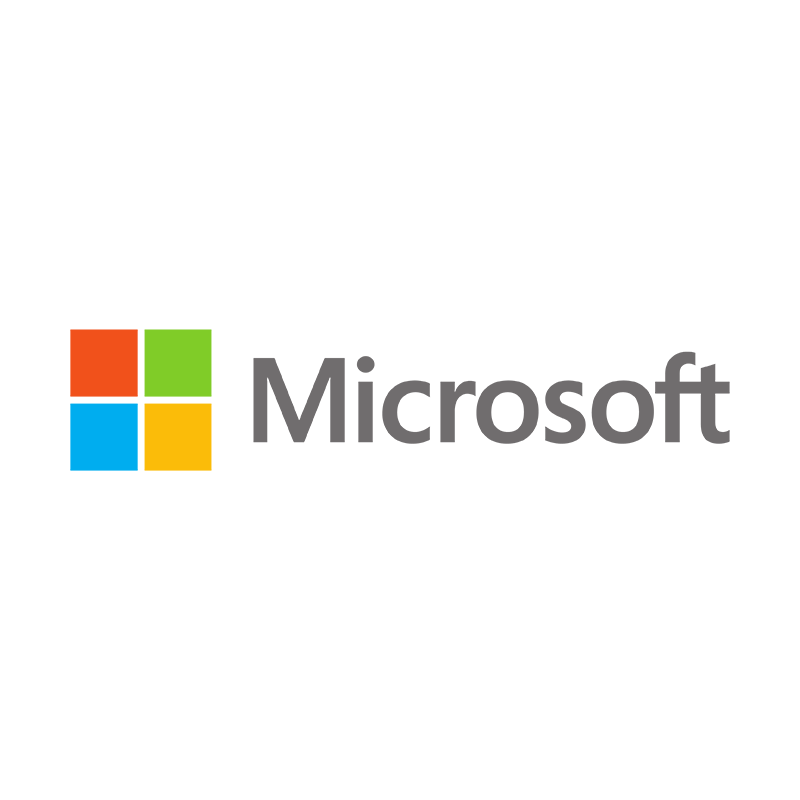}}%
    \put(\dimexpr\paperwidth-72pt-\dateboxwidth\relax,-72pt){\normalsize\monthyear}%
  }%
}
\date{}
\maketitle

\renewcommand{\thefootnote}{}
\makeatletter
\long\def\@makefntext#1{\noindent #1}
\makeatother
\hypersetup{hidelinks}
\footnotetext{\textsuperscript{*}Equal contribution. \\ \textsuperscript{$\flat$}The author and maintainer of \href{https://github.com/hugohe3/ppt-master}{ppt-master}. \\ \textsuperscript{\ddag}Corresponding author: \href{mailto:yanghuan@microsoft.com}{yanghuan@microsoft.com}, \href{mailto:he_yang@a-star.edu.sg}{he\_yang@a-star.edu.sg}, \href{mailto:EKHYap@ntu.edu.sg}{EKHYap@ntu.edu.sg}}
\addtocounter{footnote}{-1}

\begin{abstract}
\begin{mdframed}[backgroundcolor=abstractbg,
                 linewidth=0pt,
                 roundcorner=12pt,
                 leftmargin=12pt,
                 rightmargin=12pt,
                 innerleftmargin=16pt,
                 innerrightmargin=16pt,
                 innertopmargin=16pt,
                 innerbottommargin=16pt,
                 skipabove=8pt,
                 skipbelow=8pt]
\begingroup
\let\origtexttt\texttt
\renewcommand{\texttt}[1]{{\color{abstractcode}\origtexttt{#1}}}
\hypersetup{urlcolor=abstractlink}
\hypersetup{hidelinks}

Despite growing automation, turning a paper into a coherent poster, talk video, and blog piece often remains a labor-intensive last mile.
Recent systems increasingly generate multiple dissemination formats, but a practical workflow must also keep the outputs editable in native tools and bound into one navigable deliverable for revision and reuse.
We present \textbf{ResearchStudio-Reel}, a native-editable dissemination workspace that binds its three artifacts into one interactive deliverable at the experience level, implemented as five skills executable in Claude Code and Codex: one \textbf{shared} extractor, three \textbf{editable} artifact generators, and one \textbf{interactive} convergence layer.
A shared asset bundle feeds a PowerPoint poster and video deck, plus a bilingual Word blog; rather than re-rendering the paper into a fourth format, Paper2Reel converges these already-produced artifacts at the \emph{experience level}, binding poster regions, video segments, and blog passages into one interactive viewer. Artifact-specific release checks make this delivery contract testable, and Paper2Poster additionally uses a measured-fill loop.
On the Paper2Poster benchmark, our Claude Code configuration achieves the best scores among automated systems on all three aesthetic sub-criteria and the best or tied-best scores on two of three information sub-criteria.
Under two VLM judges, it exceeds the authors' posters in average aesthetics ($3.56$ vs.\ $3.03$) and wins on overall quality on 74 and 95 of the 100 papers under the two judges. The full pipeline additionally packages the native-editable source artifacts and their aligned viewer.

\vspace{4pt}\noindent\textbf{Project:}~\href{https://aka.ms/ResearchStudio}{\color{abstractlink}https://aka.ms/ResearchStudio}
\endgroup
\end{mdframed}
\end{abstract}


\begin{figure}[H]
\centering
\vspace{-10pt}
\begin{subfigure}[b]{0.33\linewidth}
  \centering
  \includegraphics[height=3.3cm]{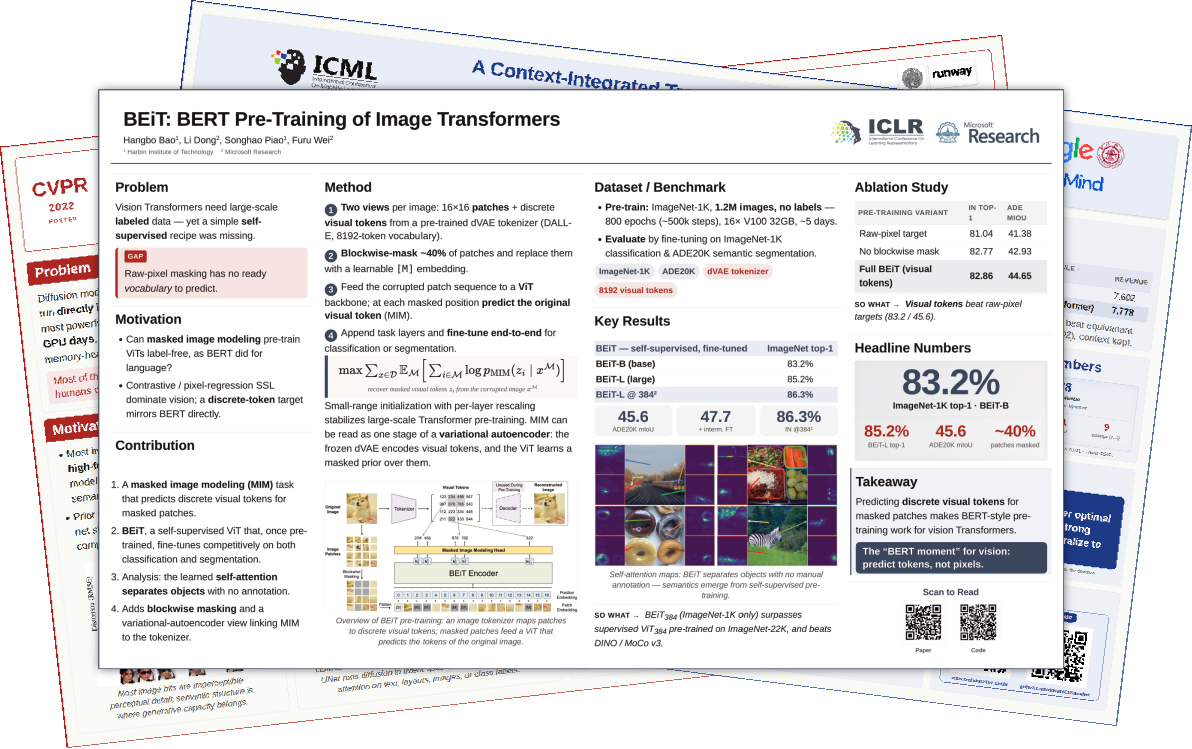}
  \caption{Poster (\textsc{Paper2Poster})}\label{fig:hero-poster}
\end{subfigure}
\hfill
\begin{subfigure}[b]{0.33\linewidth}
  \centering
  \includegraphics[height=3.3cm]{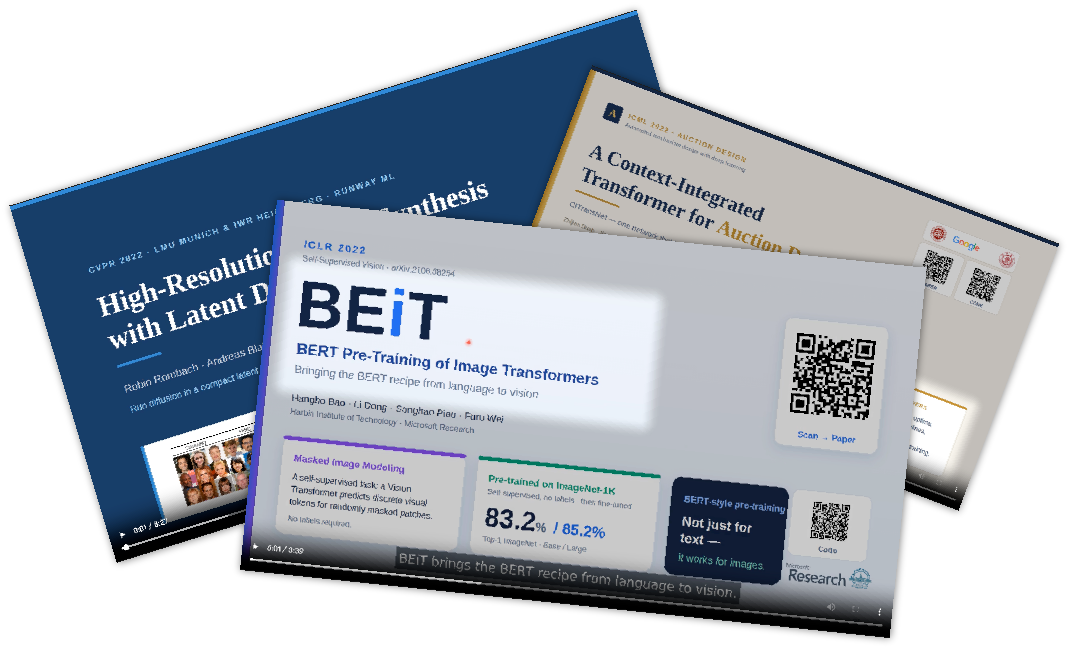}
  \caption{Talk video (\textsc{Paper2Video})}\label{fig:hero-video}
\end{subfigure}
\hfill
\begin{subfigure}[b]{0.33\linewidth}
  \centering
  \includegraphics[height=3.3cm]{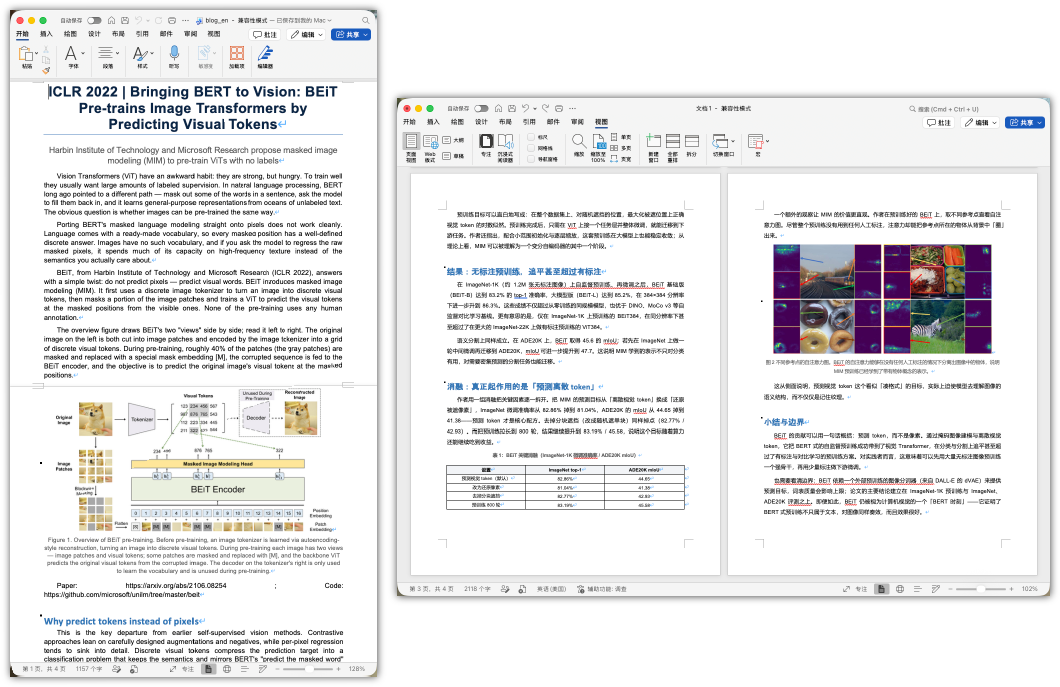}
  \caption{Bilingual blog (\textsc{Paper2Blog})}\label{fig:hero-blog}
\end{subfigure}
\caption{\textbf{Three editor-ready artifacts from one paper.} From a single accepted PDF, ResearchStudio-Reel produces a print-ready conference poster, a narration-aligned talk video, and a bilingual blog spread, shown here across example papers. Each artifact package includes a native-editable source file (PowerPoint for the poster and the video deck, Word for the blog), and the three are cross-linked into one navigable surface. A single shared \texttt{paper2assets} extraction (Figure~\ref{fig:pipeline}) grounds the artifacts in the same extracted content rather than re-deriving the paper three times.}
\label{fig:hero}
\end{figure}

\section{Introduction}\label{sec:introduction}

The dissemination layer of research is structurally separate from the paper itself and consumes time at exactly the moment authors have the least to spare. This layer covers three artifacts: the poster you stand next to at a conference session, the talk video that lives on the venue's virtual track or on the lab's YouTube channel, and the blog piece shared through social media to introduce the paper to a broader audience. Each involves figure selection, layout, narration, or audience-appropriate writing, and authors often prepare these artifacts in the days immediately after acceptance, soon after finishing the camera-ready. The stakes extend past the original author: these artifacts can also support an industrial research lab's public-facing communications, a graduate course's paper-of-the-week briefing pack, and a multilingual research organization's cross-region outreach. A workable last-mile pipeline could therefore benefit users well beyond a single author.

A fast-growing automation literature targets exactly this layer: paper-to-poster systems \citep{pang2026paper2poster, zhang2025postergen, sun2025p2p}, paper-to-video pipelines \citep{zhu2025paper2video, liang2026videoagent, liu2025preacher}, and paper-to-blog authoring and long-form summarization \citep{radensky2024posts,li2025hera,devi2025long,bian2024gosum}; we survey four buckets in \S\ref{sec:related}. Much of this literature remains artifact-specific, while PaperX and concurrent OmniPresent increasingly support multi-artifact presentation suites \citep{yu2026paperx,ma2026omnipresent}. This progress shifts the practical question from whether several formats can be generated to whether the resulting artifacts remain native-editable and navigable. Three gaps motivate our design. \textbf{(G1) Shared grounding.} Combining artifact-specific systems may repeat figure, caption, and metadata extraction, leaving cross-artifact references to be reconciled by the user. \textbf{(G2) Native editability.} A multi-format suite does not by itself guarantee native-editable source files for every artifact in the author's PowerPoint and Word workflow. \textbf{(G3) Experience-level convergence.} Parallel or consistently rendered outputs do not by themselves bind the finished artifacts into one interactive surface where a reader can jump from a poster region to the matching video segment and blog passage for revision and reuse.

We respond by treating the last mile as a native-editable dissemination workspace whose artifacts converge at the experience level, implemented through five \emph{skills} in Claude Code~\citep{anthropic2025claudecode} and Codex~\citep{openai2025codex}: the shared extractor \textbf{Paper2Assets}, the generators \textbf{Paper2Poster}, \textbf{Paper2Video}, and \textbf{Paper2Blog}, and the convergence layer \textbf{Paper2Reel}. Paper2Assets addresses \textbf{G1} by extracting one bundle with stable section IDs, figure handles, and claim anchors. Native PowerPoint and Word deliverables address \textbf{G2}, letting authors revise the poster, video deck, and blog without regenerating them. Paper2Reel addresses \textbf{G3} by converging the three finished artifacts at the experience level, binding poster sections, video segments, captions, and blog passages into one interactive, navigable deliverable. Artifact-specific release checks validate package and layout properties, while Paper2Poster's measured-fill loop supplies a categorical render gate.

These requirements remain important after the first render. Authors may need to correct a number, replace a figure, or revise the framing without rerunning an entire generator; when three artifacts are edited independently, their content and navigation can drift even if they were initially produced from the same source. We therefore treat native source files and the alignment record as first-class deliverables rather than incidental intermediates.

\begin{figure}[t]
\centering
\includegraphics[width=\linewidth]{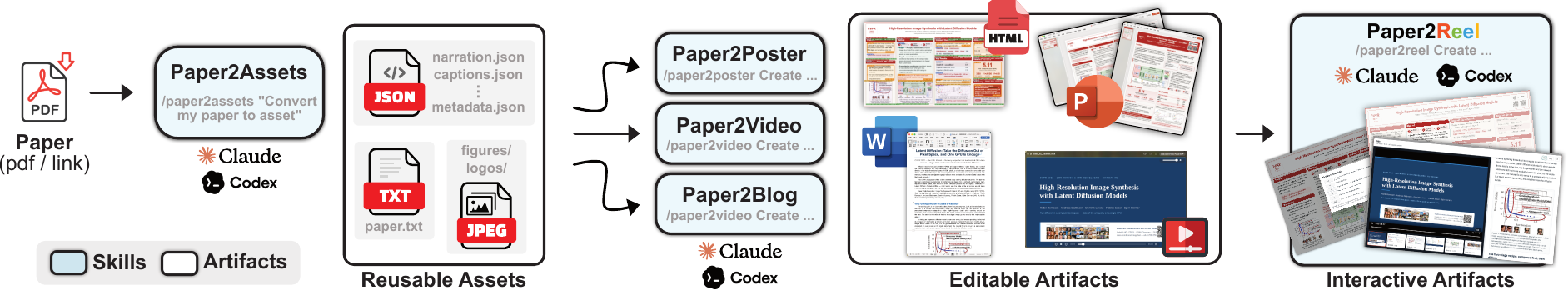}
\caption{\textbf{The ResearchStudio-Reel pipeline.} One PDF in, three editor-ready artifacts out, with one shared extraction stage in the middle. A single \texttt{Paper2Assets} pass produces the bundle that Paper2Poster, Paper2Video, and Paper2Blog each consume verbatim, and Paper2Reel binds the three into one navigable surface. Sharing the same section identifiers, figure handles, and claim anchors keeps the artifacts mutually cross-referenced rather than disjoint.}
\label{fig:pipeline}
\end{figure}

Three technical ingredients support this implementation: (i) Claude Code \citep{anthropic2025claudecode} and Codex \citep{openai2025codex} provide skill runtimes for agent-driven workflows \citep{anthropic2025skills, anthropic2024toolUse}; (ii) deterministic primitives support document and media transformations, including headless Chromium for HTML$\to$PDF \citep{playwright}, LibreOffice plus ffmpeg for slides$\to$video \citep{libreoffice, ffmpeg}, and python-docx for editorial \texttt{.docx} \citep{pythondocx}; (iii) Edge TTS \citep{edgetts} provides narration synthesis.
Figure~\ref{fig:hero} shows what the system produces, three editor-ready artifacts from one paper PDF, and Figure~\ref{fig:pipeline} sketches the pipeline behind them. 

Our contributions are as follows:
\begingroup
\setlength{\leftmargini}{1.2em}
\setlength{\topsep}{10pt}
\setlength{\partopsep}{0pt}
\setlength{\parskip}{0pt}
\begin{itemize}\itemsep=1.2pt\parsep=0pt
  \item We implement \textbf{ResearchStudio-Reel}, a native-editable dissemination workspace that turns one paper PDF into a print-ready conference poster, a narrated talk video, and a bilingual blog through composable skills.
  \item We introduce Paper2Reel, an experience-level convergence layer whose interactions connect poster regions, video segments, and blog passages in one navigable HTML surface.
  \item Its delivery contract combines one shared Paper2Assets bundle with native-editable source files: PowerPoint for the poster and video deck, and Word for the bilingual blog.
  \item The generated posters achieve the best scores among automated systems on all three aesthetic sub-criteria and the best or tied-best scores on two of three information sub-criteria on the Paper2Poster benchmark, \emph{exceeding the authors' own} by ${\sim}17.5\%$ ($3.56$ vs.\ $3.03$) on aesthetics under two VLM judges and winning the overall score on 74 and 95 of the 100 papers under the two judges; the generated video ships with narration-aligned highlight transitions, burned-in and sidecar subtitles, and target-duration control; and the generated blog ships in two languages with layout-aware DOCX repair.
\end{itemize}
\endgroup

\section{Skills}\label{sec:instantiations}

This section instantiates the pattern. Five skills compose the system: \textbf{Paper2Assets} is the shared upstream extraction layer the three generators consume; \textbf{Paper2Poster}, \textbf{Paper2Video}, and \textbf{Paper2Blog} are the three artifact generators; and \textbf{Paper2Reel} is the interactive convergence layer that turns the three deliverables into a single navigable presentation surface. The generator subsections describe each artifact package and its quality gate. The two infrastructural skills, Paper2Assets and Paper2Reel, describe the shared data layer and the final interaction layer.

\subsection{Paper2Assets}\label{sec:p2a}

Paper2Assets reads a paper PDF once and produces a single bundle, the shared stage in Figure~\ref{fig:pipeline}, that every downstream generator consumes verbatim: the full body text with page breaks preserved, the detected figure captions, a per-figure manifest, the cleaned figure images, the paper's metadata (title, authors, institutes, venue, and the paper and code links), a structured nine-section summary of the paper, and best-effort institution logos and QR codes for those links. A second pass adds an inventory manifest with the source PDF's checksum, the summary parsed into per-section records with stable identifiers, and a narration clip list in reading order, so the non-poster generators never have to re-parse anything.

\textbf{Pipeline.} Extraction runs as a short sequence of steps: pull the text and captions and crop each figure with a column-aware margin; synthesize the metadata from the first page and, for arXiv papers, the abstract page; write the nine-section summary (Problem, Motivation, Contribution, Method, Dataset/Benchmark, Key Result, Ablation, Headline Numbers, Takeaway), where each section carries a short essential entry, a supplementary entry held in reserve, and a spoken audio script; fetch the institution logos best-effort from Wikimedia Commons (the official mark linked in each institute's English Wikipedia infobox, filtered to \texttt{logo}/\texttt{seal}/\texttt{wordmark}-type files and resolved to the full-resolution upload) and the venue logo from Wikidata, verifying the downloaded bytes and, when an institute or venue does not resolve, hiding that titlebar slot rather than showing a placeholder; render the QR codes; and finally emit the canonical record files the downstream skills read. Each step is a separate idempotent script writing its own output, so an improved or failed stage re-runs in isolation without re-extracting the whole paper, and the reading-order narration list is fixed once here so the poster, video, and blog all narrate the paper in the same sequence.

\textbf{Figure cleanup.} The load-bearing and most time-consuming stage is the figure-cleanup chain, and it runs once on every extracted figure rather than being redone, generator by generator, each time a downstream picks a figure to render. A deterministic prefix strips chrome residue, baked-in caption strips, and uniform white margins to handle the easy cases; a visual-AI step then judges a tight bounding box, a fresh-context sub-agent verifier independently re-reads the original against the proposed crop, and only a clean pass commits, splitting the image when one raster packs two independent figures. Every mode is designed to be idempotent, and the raw extract is backed up once before any crop, so re-runs can start from the preserved source.

The bundle is the only interface between Paper2Assets and the rest of the system: downstream skills read it and never re-open the PDF, and a stable figure-naming invariant supports consistent figure references across the three artifacts without any cross-skill coordination, so the figure the poster's method card embeds is the one the video's matching slide cites and the blog's evidence map references. This single-owner extraction is the skill's distinctive contribution. Many artifact-specific paper-to-X systems (\S\ref{sec:related}) re-derive figure crops, caption alignment, and metadata inside their own renderer (the duplicated-extraction problem labeled G1 in \S\ref{sec:introduction}), whereas folding that work, and especially the expensive figure cleanup, into one shared owner lets the downstream artifacts reuse the same figure references without repeating cleanup, reducing the risk of each renderer either repeating the verifier loop or shipping visible defects such as body text leaking into a crop or a caption baked into the image.

\subsection{Paper2Poster}\label{sec:p2p}

\begin{figure}[!b]
\centering
\includegraphics[width=\linewidth]{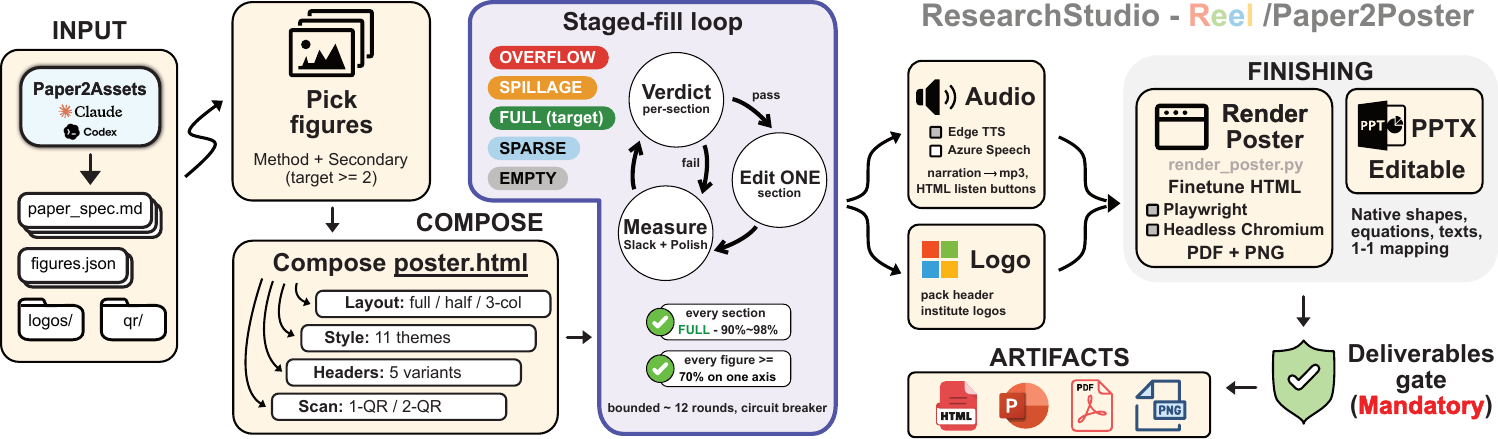}
  \caption{\textbf{The Paper2Poster pipeline.} A Paper2Assets bundle (paper spec, cleaned figures, logos, QR) drives an agent that picks the Method plus secondary figures and composes a self-contained \texttt{poster.html} along four axes: column layout, visual style, title-band header, and the Scan-to-Read block. A staged-fill loop then measures each section (slack~+~polish) and edits one section per round until every panel reads \texttt{FULL} (90--98\%) and every figure is large enough on one axis. Narration audio and header logos are packed in, and the converged page is rendered to PDF and PNG and to an editable, native-shape PowerPoint, released only through a mandatory deliverables gate.}

\label{fig:p2p-pipeline}
\end{figure}

Paper2Poster turns the shared asset bundle from Paper2Assets into a print-ready conference poster, without re-reading the source PDF. A single run yields the poster in four coupled formats (a self-contained web page, a print-resolution PDF at the exact canvas size, a thumbnail image, and a natively-editable PowerPoint file), together with optional per-section narration audio.

\subsubsection{Design Requirements}

A conference poster is not a summary that can be stretched or trimmed at will; it is a single fixed-size page that must stay readable, look deliberately designed, and remain editable. The engineering that took the most iteration was not the visual design but five recurring requirements, and the rest of this section answers them in order. \textbf{(A1) Composition without a template explosion.} Covering the real variety of posters with a bank of fixed templates means a combinatorial blow-up across column layout, visual style, header arrangement, and the bottom QR block, which is brittle to extend and still tends to make every poster look alike. \textbf{(A2) A fill loop that converges.} The band in which a section reads full is narrow, yet a single discrete text edit moves a section well past it, so the natural refine loop oscillates between too-empty and over-full for dozens of rounds unless its step size and stopping rule are engineered deliberately. \textbf{(A3) A page too large to re-read.} The working poster is one HTML file of around a hundred kilobytes; re-reading it on every refinement round floods the agent's context window and triggers a compaction that erases the loop's own progress, and emitting it through the model's output channel overflows the per-turn token cap and kills the run outright. \textbf{(A4) Figures that fill their cards.} In any column that holds a figure, the figure absorbs whatever vertical room the text leaves, so ordinary text edits do not move it and a figure stranded as a small stamp has to be resized through the one lever that actually changes its box. \textbf{(A5) A faithful editable export.} A poster the author can reopen requires converting the rendered page into native PowerPoint shapes, including native equation objects, rather than pasting in a flat image.

\subsubsection{Our Solution}

\textbf{Composition over fixed templates~(A1).} Rather than maintain a combinatorial set of fixed templates, the poster is assembled at build time from four orthogonal axes: the column \emph{layout} (full, half, or three-column), the visual \emph{style} (one of a family of interchangeable themes, each a self-contained CSS file that adds a new look with no other code change), the title-band \emph{header} (one of five arrangements of the venue logo, institution logos, and QR codes), and the internal layout of the bottom \emph{Scan-to-Read} block. Only the layout is fixed by a hard rule (the Method figure's shape routes a wide or full-column figure to a merged-column grid and everything else to a half-width default), while the remaining axes are sampled reproducibly, so a batch of posters gets a stable spread without a brittle ``pick a random style'' step and the agent never debates layout family or theme but only layout \emph{fill}, where its budget is best spent. A companion rule that every section carry a distinct visual widget keeps adjacent sections from collapsing into an undifferentiated wall of text.

\textbf{A staged fill loop~(A2).} The poster starts as a lean draft holding only each section's essential text and a small set of selected figures (always the Method diagram and at least one key-result or teaser visual), and a measured-fill loop grows it to a full page. The loop is deliberately discrete rather than a continuous optimization. On each pass the page is rendered in a headless browser and every section is assigned a scalar $\mathrm{fullRatio} = h_{\mathrm{content}}/h_{\mathrm{card}}$, the painted content height (the bottom edge of the section's lowest rendered element, read from \texttt{getBoundingClientRect}) divided by the inner height of its card; this ratio is quantized into one of five categorical verdicts, and the verdict alone selects the remediation. The five bands and their moves are: \textbf{(1)}~\texttt{EMPTY} ($\mathrm{fullRatio} < 0.70$), a load-bearing underfill, remediated by appending a withheld supplementary paragraph or promoting an optional section into the column; \textbf{(2)}~\texttt{SPARSE} ($0.70$ to $0.90$), a visible underfill, remediated by \emph{polishing up}: padding the existing prose or enlarging a widget until the content reaches the card floor; \textbf{(3)}~\texttt{FULL} ($0.90$ to $0.98$), the target band, left untouched; \textbf{(4)}~\texttt{SPILLAGE} ($0.98$ to $1.10$), content just past the target band, remediated by \emph{polishing down}: tightening prose until it re-enters the band; and \textbf{(5)}~\texttt{OVERFLOW} ($> 1.10$), content visibly past the border, remediated by dropping a supplementary paragraph or an entire optional section. The Method figure is retuned on the same pass under a separate figure-fill gate that requires it to paint at least $90\%$ of its card on one axis, and the loop terminates only when every section reads \texttt{FULL} \emph{and} no figure trips that gate. Because termination is a categorical fixed point rather than a learned-score plateau, and each iteration logs which move was applied to which section, a poster that fails to converge is diagnosable rather than a ``the agent did something'' mystery. Three mechanisms damp oscillation across the narrow \texttt{FULL} band: each move is sized by the signed pixel delta the measurement reports rather than guessed at, the loop refuses to re-apply a move that already overshot the band on a given section, and an on-disk round counter trips a circuit breaker that ships the best-measured state rather than grinding indefinitely.

\begin{figure}[!b]
\centering
\includegraphics[width=0.95\linewidth]{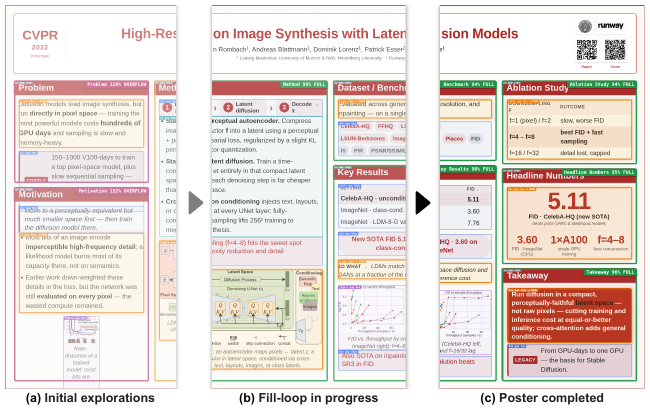}
\caption{\textbf{The staged-fill loop, visualized on the Latent Diffusion Models poster.} A debug overlay boxes every section and colors it by fill verdict (red / amber for \texttt{EMPTY} / \texttt{SPARSE}, green for \texttt{FULL}, orange / magenta for \texttt{SPILLAGE} / \texttt{OVERFLOW}), annotated with its fill percentage. \emph{(a) Initial explorations:} the freshly composed draft is uneven, with several underfilled sections and small figures. \emph{(b) Fill-loop in progress:} the loop measures each section and edits one per round, so cards pass through \texttt{OVERFLOW} and \texttt{SPARSE} as content is added or trimmed. \emph{(c) Poster completed:} the loop stops once every section reads \texttt{FULL} (90--98\%) and every figure is large enough, yielding the shipped poster.}
\label{fig:fill-loop}
\end{figure}

\textbf{A loose gate, then a render-time expand.} The fill gate is deliberately loose: a section counts as full at 90\% of its card rather than 98\%, because the last few percent are exactly where a discrete-edit loop oscillates, and tightening the gate toward 98\% markedly increases the number of refinement rounds, sharply so past $0.94$, for little visual gain. To recover those percent without paying that cost, a single render-time pass then stretches every under-filled card toward 98\% by growing the whitespace between its rows, so the shipped poster reads visually full while the loop still converges against the cheap 90\% gate. The expand is constrained: figures are not resized, and any card whose growth would push the fixed-canvas layout taller is reverted, to avoid moving column bottoms. It is baked into the poster once, so the web page, the PDF, the PNG, and the editable PowerPoint all show the same filled layout.

\textbf{Editing the page without re-reading it~(A3).} The refinement loop never pulls the whole poster into context. Each measurement returns, next to the per-section verdict, the verbatim source of exactly the sections that are off-target, so the agent edits those snippets in place and the hundred-kilobyte file never enters its window or forces a compaction that would erase what the loop has already tried. The file is likewise generated indirectly and never printed back through the output channel, so neither the input nor the output token budget is spent on the bulk of the page.

\textbf{Sizing figures by their one real lever~(A4).} Because a figure column's height does not respond to its text, the loop sizes each figure through its height cap rather than through the surrounding prose, and holds every figure to a hard floor on its card so that ordinary text edits can never leave it stranded as a small stamp. This yields two regimes with distinct levers: a figure sitting below its cap with whitespace beneath it is enlarged by raising the cap, while a crowded column is relieved by trimming text, since the figure never shrinks past its floor. That floor is one of the fill loop's own termination conditions, so the loop cannot converge while any figure remains below it.

\textbf{The editable PowerPoint bridge~(A5).} A headless browser renders the converged page to a PDF and a PNG thumbnail, and the same run also emits an editable PowerPoint, giving both poster traditions at once: the web page as the visual ground truth and PowerPoint as what conference attendees and lab editors actually revise. Rather than rasterize the finished page and reverse-engineer shapes from pixels the way a generic PDF-to-PowerPoint converter does, the bridge rebuilds each slide directly from the live DOM, reading every element's on-screen position, size, and style and turning it into the native PowerPoint object its role calls for: a text block becomes an editable text frame that keeps its bold and italic emphasis, a figure becomes a replaceable picture, an equation stays a native editable equation rather than a flattened image, and a section card becomes a styled shape. Because the layout and its meaning are read from the DOM rather than guessed from pixels, the author can fix a typo, swap a figure, or recolor a card in PowerPoint and re-export without re-running the fill loop, turning ``a poster the system designed'' into one the author can continue editing.

Figure~\ref{fig:p2p-pipeline} summarizes the pipeline, and Figure~\ref{fig:fill-loop} shows the staged-fill loop bringing an example poster to convergence.

\subsubsection{Author Preference Alignment}

Several of the poster's defaults are not technical necessities but choices matched to what real posters and venues do, so the output reads like something an author would have made rather than a generic render. Author ground-truth posters tend to blow up their institution logos, so the header packs every logo to one shared, generous height instead of a timid uniform small mark, and caps that shared height when a paper lists many institutions so the title band absorbs the rest rather than overflow. Most author posters for figure-heavy papers use a wide three-column grid, so once a paper contributes more than three high-signal figures the layout switches to three equal columns whose wider tracks hold bigger figures. The page is sized to the format each venue expects, a wide sixty by thirty-six inch landscape for ICML, NeurIPS, and CVPR and an A0 portrait for the ACL family, rather than one fixed shape. Academic posters are overwhelmingly light, so the default is a pale title band over white section cards with the accent colour used sparingly, while deep saturated backgrounds are held back. Finally, the Scan-to-Read QR is treated as furniture: it is rendered only for the paper and code links that exist, and dropped when the header is already crowded or the section would render too wide and flat for a lone QR, rather than shipped adrift in empty space.

\subsubsection{Quality Gates}

Like the other generators, a poster ships only through a mandatory deliverables gate that refuses to release it until several conditions hold together. Convergence is the fill loop's own exit target, enforced during the loop rather than at release: the loop terminates normally only once every section lands in the full band, between ninety and ninety-eight percent of its card, with no panel left visibly empty or overflowing, and only once every figure fills at least ninety percent of its card on one of its axes, so no figure reads as a small stamp in an empty card. Because a discrete edit loop cannot always seat every section in-band, an on-disk circuit breaker (\S\ref{sec:p2p}) bounds the iteration count and, on non-convergence, ships the best-measured state marked degraded, with the render-time whitespace expand recovering the residual fill; the release gate does not re-litigate the fill verdict. An optional reading-comprehension gate, Reader-Reconstruction Preference (RRP), adds a second axis: it accepts an edit only if a held-out reader model can still answer questions about the paper from the poster alone, providing a check against aesthetic edits that reduce answerability on the reconstruction questions. A final deterministic check confirms that the rendered PDF matches the intended fixed-size canvas before any file reaches the user, a non-advisory gate, since a model will otherwise sometimes skip the render on the rationalization that the other gates passed. Only when all four artifacts are present and non-truncated and this canvas check passes are the HTML file, the PDF, the PNG, and the editable PowerPoint released as one bundle.

\subsection{Paper2Video}\label{sec:p2v}

\begin{figure}[!b]
\centering
\includegraphics[width=\linewidth]{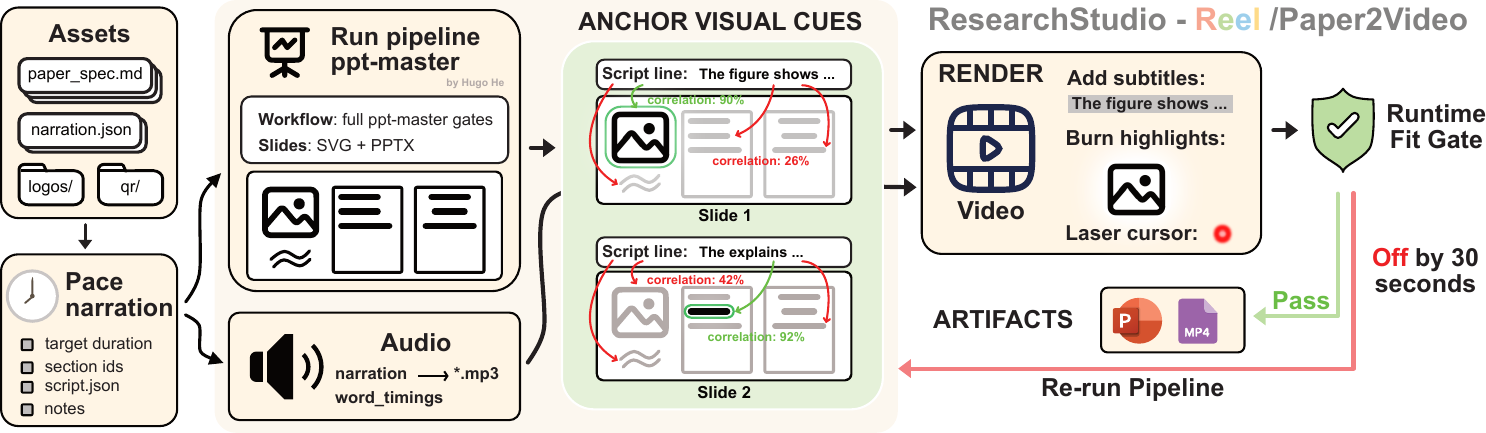}
\caption{\textbf{Paper2Video overview.} The skill reuses the Paper2Assets bundle, plans narration and duration, delegates deck authoring to the full \texttt{ppt-master} workflow, synthesizes aligned audio and captions, renders visual attention cues, and packages the editable deck with captioned and no-subtitle videos. The root deliverables are \texttt{video.pptx}, \texttt{video.mp4}, and \texttt{video\_no\_subtitles.mp4}, while timelines, captions, audio clips, and visual cues stay under \texttt{assets/}, the auditable intermediates directory.}
\label{fig:p2v-overview}
\end{figure}

Paper2Video turns the shared Paper2Assets bundle into a synchronized video package. It uses the full \texttt{ppt-master} workflow \citep{pptmaster2025} for deck authoring, so slide design is treated as a dependency rather than a separate claim. The skill's role is to turn that deck and the shared paper assets into an auditable media bundle: a planned narration script, section-level audio, captions, visual attention cues, two MP4 variants, and a timeline that Paper2Reel can navigate. A completed run writes three user-facing deliverables at the bundle root: \texttt{video.pptx}, \texttt{video.mp4}, and \texttt{video\_no\_subtitles.mp4}. Intermediate files stay under \texttt{assets/}, including narration audio, subtitle sidecars, rendered slide frames, duration reports, visual-cue plans, timelines, and QA reports.

Table~\ref{tbl:p2v-capabilities} compares this media contract against representative presentation, video, and paper-to-video systems. The comparison is capability-based, not score-based: it asks which objects the workflow produces and checks.

\subsubsection{Design Requirements}

A good slide deck does not by itself make a reliable paper video. Once the deck leaves the slide authoring workflow, the system still has to synchronize time, sound, captions, highlights, and downstream navigation. We therefore focus on five media-level requirements. \textbf{(B1) Duration must be planned before rendering.} A target-length video cannot be made safely by cutting the final MP4, clipping audio, or applying a large global speed-up after rendering. The narration has to be shaped before TTS, then checked again after the measured render. \textbf{(B2) The viewer needs attention guidance, not only slide playback.} Research slides often contain multiple figures, tables, and formula blocks. If the narration discusses one region while the whole slide remains static, the viewer has to guess where to look. \textbf{(B3) Captions have two delivery contexts.} A shareable MP4 should carry readable burned-in subtitles, while the interactive reel needs a clean video source so its own caption toggle does not create double subtitles. \textbf{(B4) The video has to remain addressable after export.} Downstream tools should not recover section boundaries by scraping pixels from the MP4 or guessing from slide order. The exported video needs a sidecar contract that keeps sections, audio windows, captions, slides, and visual cues aligned. \textbf{(B5) Media failures need deterministic checks.} Missing audio, empty subtitles, malformed highlight boxes, timeline drift, blank frames, and unsafe duration fixes are easy to miss by eye during a long generation run, so they need hard package gates.

\subsubsection{Our Solution}

\textbf{Narration and duration planning~(B1).} Paper2Video first converts the shared section narration into a video script with stable section ids. When the user requests a target length, the planner estimates the script before TTS and asks for semantic rewrites when the text is too long or too short. After rendering, a duration report compares the measured MP4 length with the plan. Small residual errors can be repaired by a bounded speech-rate plan. Large errors return to narration rewriting, so the target length is reached through content planning rather than by truncating the final video.

\textbf{Deck authoring through \texttt{ppt-master}.} The editable deck is produced by the full \texttt{ppt-master} route. Paper2Video passes the shared paper assets, the section script, and optional visual-anchor requirements into that workflow, then uses the exported PPTX as the render source. The top-level \texttt{video.pptx} is therefore not a disposable intermediate. It is a user-facing artifact that an author can reopen and edit.

\textbf{Narration-aligned visual highlights~(B2).} Before final rendering, Paper2Video can turn the script into a visual-cue requirement file. The deck workflow attaches semantic anchors to the visible objects named by the narration, such as a figure panel, equation block, table row, or method card. A cue resolver then combines the script, word timings, slide geometry, and authored anchors to produce \texttt{visual\_cues.json}. The production renderer uses normalized geometry, so the same cue plan survives different video resolutions. Its default delivery style is \texttt{spotlight\_laser}, which softly emphasizes the target region and marks the current focus point.

\textbf{Audio, captions, and MP4 variants~(B3).} The renderer synthesizes one narration clip per script section and writes subtitle sidecars from the same timing model. It first renders \texttt{video\_no\_subtitles.mp4}, which preserves the slide frames, narration audio, and visual highlights without burned-in text. It then burns the subtitle layer into \texttt{video.mp4}. The two MP4 files share the same slide frames, audio alignment, and highlight timing, and the only difference is whether subtitles are burned in.

\textbf{Timeline and bundle contract~(B4).} Paper2Video writes \texttt{timeline.json} as the canonical sidecar for downstream navigation. Each timeline entry maps a paper section or narration chunk to its audio window, subtitle cues, slide frame, and accepted visual cue. Paper2Reel reads this sidecar directly when it connects poster sections, slide thumbnails, captions, and video seek points. The final package is therefore not only a compressed video. It is an editable and navigable media bundle.

\textbf{Deterministic media checks~(B5).} The media-level failure modes that are easy to miss during a long generation run are not left to visual inspection: missing audio, empty or doubled subtitles, malformed or word-sized highlight boxes, timeline drift, blank or overflowing frames, and unsafe post-hoc duration fixes are each caught by a single mandatory package gate rather than trusted to the eye. That gate is the release contract for the bundle, and its checks are detailed in the Quality Gates below.

\subsubsection{Quality Gates}

\begin{figure}[!b]
\centering
\vspace{-0.2cm}
\includegraphics[width=0.9\linewidth]{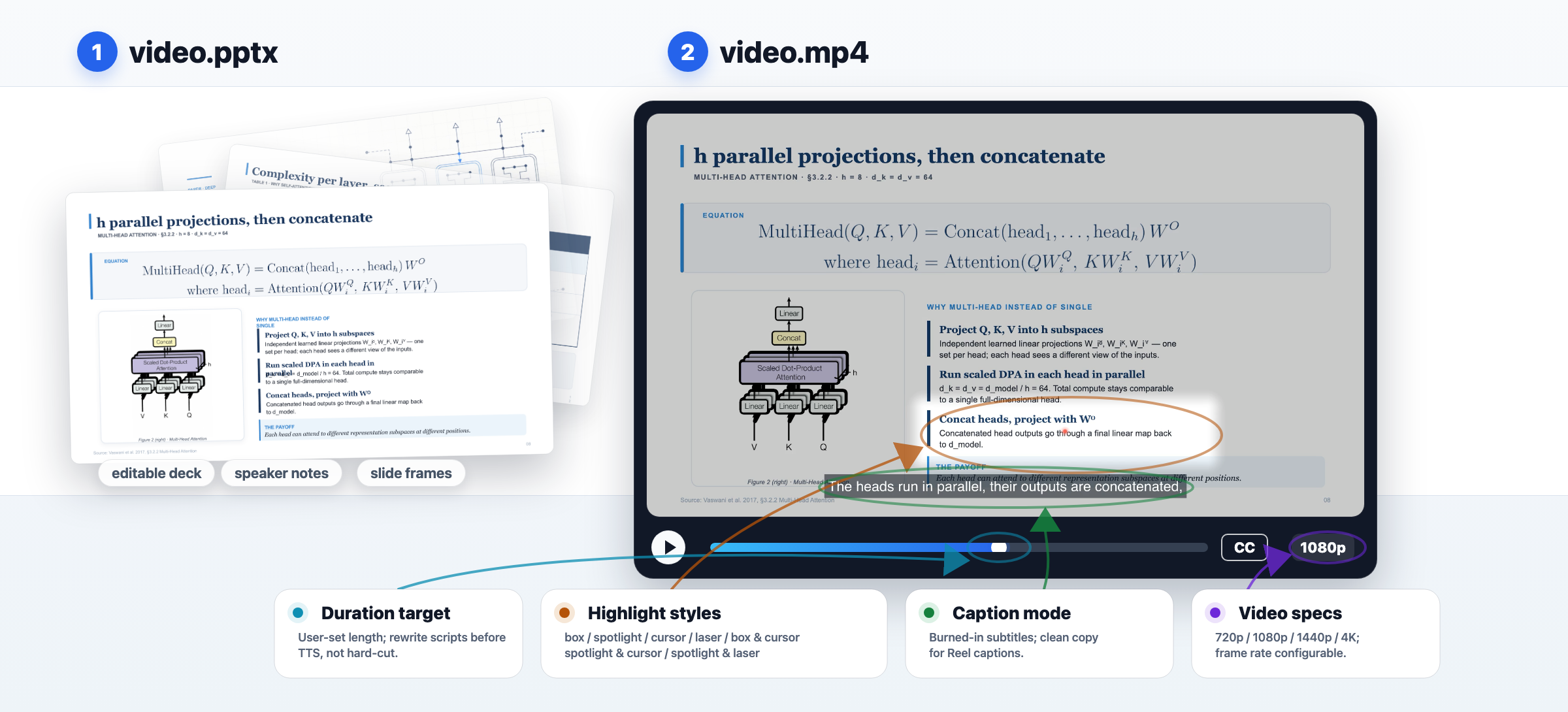}
\caption{
\textbf{Paper2Video deliverable showcase.} The figure pairs the editable \texttt{video.pptx} with the captioned \texttt{video.mp4} and summarizes the user-facing controls checked by the package gate: target duration, highlight style, caption mode, and video specification.
}
\label{fig:p2v-showcase}
\vspace{-0.5cm}
\end{figure}

The final package gate is \texttt{check\_video\_package.py}. It checks the acceptance questions that matter to a user and to downstream skills. The gate verifies that \texttt{video.pptx}, \texttt{video.mp4}, and \texttt{video\_no\_subtitles.mp4} exist in the expected locations, that the MP4 files are playable, that the final video has an audio stream, and that the measured duration is inside the requested tolerance when duration control is enabled. For subtitle delivery, it requires a non-empty subtitle sidecar, requires both the raw no-subtitle MP4 and the final subtitled MP4, checks that the final MP4 is not byte-identical to the raw render, and warns if subtitle burn-in changes the duration too much.

For visual attention, the gate requires \texttt{visual\_cues.json}, the cue plan, and the timeline when highlighted delivery is requested. It checks cue coverage, normalized geometry, semantic target ids, timing within the matching audio segment, and consistency with the accepted cue plan. It also rejects word-sized highlight boxes for final presentation use, because the highlight should point to a meaningful module, card, row, figure part, or bullet group rather than a tiny phrase. The same gate parses the PPTX and rendered frames to catch blank frames, severe text overflow, text-image overlap, over-shrunk visual content, and cropped slide content. For duration control, it requires a TTS rate plan and rejects unsafe rate changes, so target length cannot be achieved by an uncontrolled speed change.

These checks make the video package testable rather than merely generated. A passing package answers the concrete questions raised by the workflow: the video has sound, subtitles are delivered in the correct variant, highlights are present and timed when requested, duration control stays within tolerance, and \texttt{timeline.json} keeps audio, captions, slides, and highlights aligned for Paper2Reel. Figure~\ref{fig:p2v-showcase} visualizes the deliverables and options these checks cover.

\subsection{Paper2Blog}\label{sec:p2b}

Paper2Blog turns the shared Paper2Assets bundle into a bilingual editorial package for human editors. A completed run writes two root deliverables: the Chinese article, cast as a WeChat public-account piece, and the English article, cast as a research-blog piece. The two documents share the same facts, figures, numbers, claims, and source links, but they are written for different reading habits. Their outlines, previews, reports, and figure dependencies stay in the intermediates directory, so the bundle root remains limited to the deliverables and the manifest.

Table~\ref{tbl:p2b-capabilities} positions this article contract against representative scientific summarizers and research assistants. The comparison focuses on whether the tool produces an editable, grounded blog artifact rather than only an answer or note.

\subsubsection{Design Requirements}

Paper2Blog is not just a bilingual summarizer. It has to produce two editable articles that agree on the science while reading naturally in different editorial contexts. We found five requirements to be central. \textbf{(C1) One evidence base for two languages.} The Chinese and English articles may use different openings, pacing, and examples, but they cannot drift on results, method names, figure references, or source links. \textbf{(C2) Register must be controlled during writing.} The Chinese version should read like a restrained WeChat public-account article, while the English version should read like a neutral research blog for technical readers. This cannot be checked reliably after the fact by a deterministic script, so it has to be specified before drafting. \textbf{(C3) Figures need article-level selection and placement.} Paper2Assets provides reusable figures, but a blog should not embed every available figure. It needs a small shared figure set, placed near the paragraph that explains it, with article-specific captions in each language. \textbf{(C4) The deliverable is a Word article, not plain text.} Editors need Word files with embedded images, stable fonts, and fixed filenames, so the article opens ready to revise rather than as plain text to reformat. \textbf{(C5) Word layout has to be editor-ready.} Correct prose is not enough if the rendered document leaves large blank pages, images shrunk to thumbnails, or an orphan word or two stranded on a paragraph's final line; pagination and image fit have to be checked on the document as a rendered visual artifact, not only as text.

\subsubsection{Our Solution}

\begin{figure}[!b]
\centering
  \includegraphics[width=\linewidth]{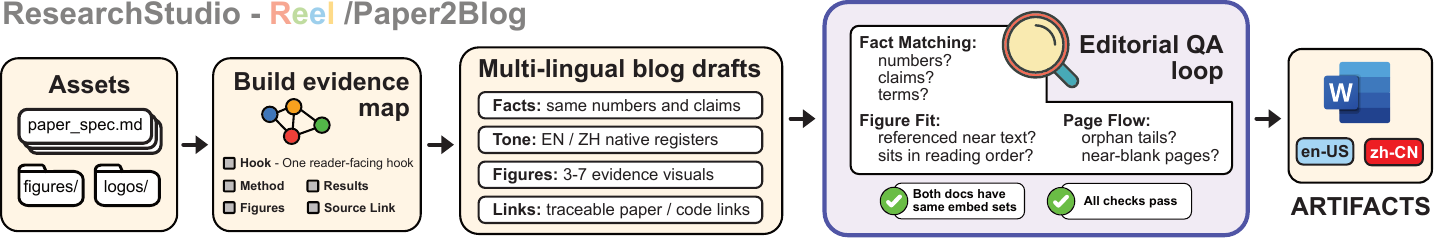}%
\caption{\textbf{Paper2Blog pipeline.} The skill reuses the Paper2Assets bundle, builds one shared evidence map, selects a shared article figure set, writes two language-specific outlines and DOCX files, and runs a strict package gate. The root deliverables are the Chinese article and the English article. Outlines, previews, and QA reports stay in the skill's auditable intermediates directory.}
\label{fig:p2b-pipeline}
\end{figure}

\textbf{Shared evidence map~(C1).} Paper2Blog first builds one evidence map from the shared bundle. The map records the paper's hook, problem, method components, main claims, quantitative results, limitations, source links, and figure roles. Both language drafts read from this map. If a hard fact such as a code link, DOI, venue, affiliation, or acceptance status is missing from the inputs, the draft omits it instead of guessing.

\textbf{Language-specific outlines and voice~(C2).} The two articles are written separately rather than translated sentence by sentence. Before drafting, the skill reads a language-specific editorial style guide. The Chinese outline targets a restrained public-account register, with necessary English technical terms kept and explained when useful. The English outline targets a neutral research-blog register. The gate later checks hard consistency, but the register itself is controlled at generation time through the style guide and the separate outlines.

\textbf{Shared figure set for article evidence~(C3).} Paper2Blog selects only the figures that help the article explain the paper. The same selected set is used in both languages, and each figure is placed next to the section that prepares the reader to understand it. This is not a second figure-extraction stage. It is a selected-figure review for article use: the skill checks that chosen figures are readable after DOCX resizing, that article captions do not fight with leftover source captions, and that image placement does not create obvious layout imbalance.

\textbf{DOCX assembly and bundle contract~(C4).} The assembler writes the Chinese article and the English article with fixed filenames, embedded media, captions in the matching language, source links, and editor-friendly fonts. The fixed root names are part of the contract, since downstream upload, zip, CMS, and Paper2Reel tooling can find the documents without parsing paper titles or non-ASCII filenames. The outlines, rendered previews, crop records, and QA report stay in the intermediates directory. Figure~\ref{fig:p2b-pipeline} summarizes the package flow.

\textbf{Layout treated as a rendered artifact~(C5).} Beyond writing the prose, the assembler and its gate inspect the DOCX the way an editor sees it on the page. Image fit after resizing, pagination, and orphan tails are read from the document's internal structure and, in strict mode, from rendered page images, so a near-blank page, a stranded thumbnail, or a lone trailing word is caught before delivery rather than left for the editor to find. The specific layout checks are detailed in the Quality Gates below.

\subsubsection{Quality Gates}

\begin{figure}[!b]
\vspace{-0.2cm}
\centering
\includegraphics[width=0.9\linewidth]{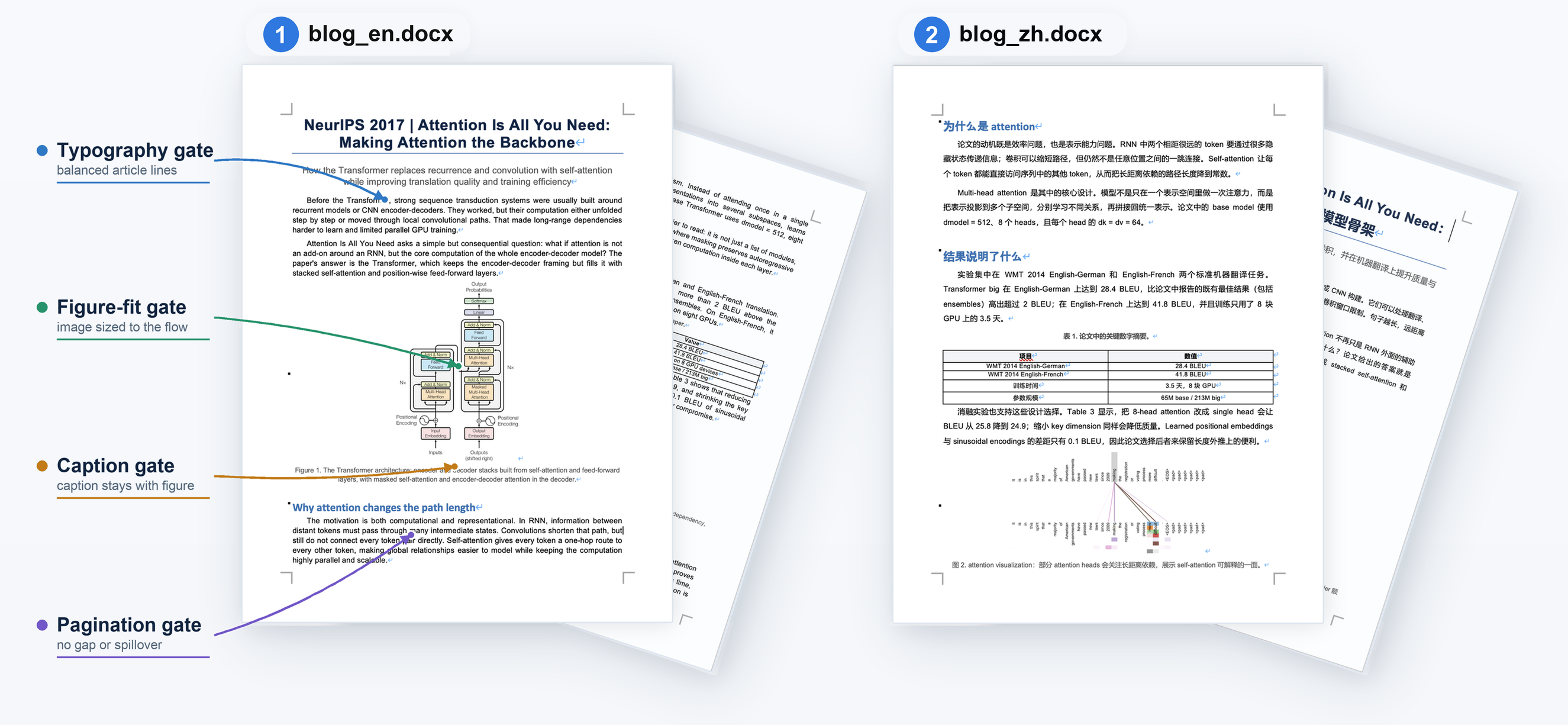}
\caption{\textbf{Paper2Blog DOCX showcase.} The figure shows the two required Word deliverables, the English article and the Chinese article, together with the layout checks discussed in the text: typography balance, figure fit, caption placement, and pagination risk.}
\label{fig:p2b-showcase}
\end{figure}

The final gate is the package gate. It first reads both Word packages directly. The gate checks that the Chinese article and the English article exist, that they are readable Word files, that they contain enough article text, that images are embedded rather than linked, and that placeholder text such as leftover editing markers or Chinese placeholder phrases is absent. It also checks font declarations, covering a Latin body font and a Chinese fallback font.

The bilingual checks focus on facts that can be tested deterministically. The gate compares the number of embedded images, their identity and order when available, figures are recorded in the outlines, extracted numeric claims, and technical-looking terms. These checks are not to judge writing style. Instead, they catch concrete failures that cause the two native articles to diverge, such as a figure appearing in only one language version but being dropped by the other.

The layout checks treat DOCX as a visual artifact. From the document's internal structure, the gate flags underfilled images, images that move to the next page while a moderate resize could fit them, and likely orphan tails such as one English word or a few Chinese characters left alone on the final line of a paragraph. In strict mode, it renders both documents to page images, then inspects them for near-blank pages, sparse pages, and large bottom whitespace on non-final pages. A passing package is therefore not only bilingual and grounded. It is also close enough in Word layout for an editor to revise rather than rebuild.
Figure~\ref{fig:p2b-showcase} illustrates these DOCX-level checks on the two deliverables.


\subsection{Paper2Reel}\label{sec:connect}

Paper2Reel is the convergence and presentation layer of ResearchStudio-Reel. It does not replace the poster, video, or blog generators. Instead, it reads their completed deliverables and builds a self-contained interactive viewer around them. The root deliverable is the viewer page. A second root file, the alignment record, records how poster sections, slide thumbnails, video times, captions, and blog blocks line up. The native poster, slide, video, and document files remain downloadable, while the reel adds a section-level reading surface on top of them.

\begin{figure}[!b]
\vspace{-0.2cm}
\centering
\includegraphics[width=1.0\linewidth]{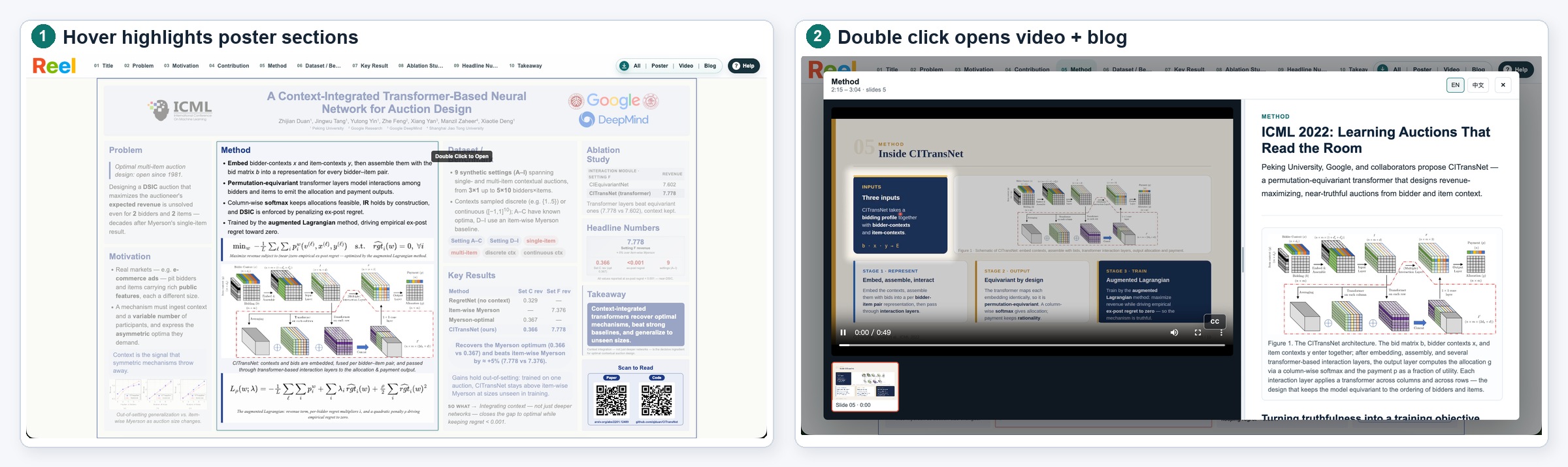}
\caption{\textbf{Paper2Reel interaction showcase.} (1) Hovering a poster section highlights the active block while neighboring sections fade. (2) Double-clicking the section opens the synchronized modal, with video playback, caption control, and slide thumbnail on the left, a language switch in the header, and the aligned blog passage on the right.}
\label{fig:p2r-showcase}
\end{figure}

\subsubsection{Interaction Design}

The viewer is poster-first because the poster is the most compact map of the paper. The first screen is therefore the generated poster, not a dashboard or a tabbed report. Moving over a poster section gives immediate hover feedback, and double-clicking a section opens the section modal. The title area opens a full-paper modal. The top menu, help panel, audio control, and downloads stay out of the way until called through the UI or keyboard shortcuts.

The section modal is the main reading unit. It places video on the left and blog content on the right, with a draggable splitter between them. The video pane uses the subtitle-free video source for playback, while captions come from caption sidecars controlled by the viewer's subtitle toggle. This keeps the interactive caption switch useful and avoids double subtitles. Slide thumbnails sit below the video and seek to the corresponding time when clicked. Direct progress-bar seeking uses the same timing contract. The blog pane shows the matching article block and supports both English and Chinese content from Paper2Blog.

This design makes the reel a presentation surface rather than a fourth generated summary. A reader can start from the poster, enter the section they care about, watch the matching video segment, read the matching blog passage, switch language, inspect slide context, and download the native artifacts without leaving the bundle.
Figure~\ref{fig:p2r-showcase} illustrates the two interaction states that carry most of the reader workflow.

\subsubsection{Content Alignment}

The load-bearing object is the alignment sidecar. Each canonical section id maps to the corresponding poster block, one or more slide targets, video start and end times, subtitle tracks, slide thumbnails, and blog blocks. The viewer reads this sidecar instead of guessing from filenames, scraping the poster, or inferring times from the video. As a result, a poster block, a video segment, a slide thumbnail, and a blog passage become different views of the same paper section.

Paper2Reel also uses the bundle contract to handle incomplete inputs. If the user invokes the skill on a PDF, an arXiv link, or a partial bundle, it inspects which upstream deliverables are missing and completes those stages through the full Paper2Assets, Paper2Poster, Paper2Blog, and Paper2Video workflows before assembling the viewer. This preserves the same section ids and asset paths across the whole package, rather than allowing a reel-specific shortcut to drift away from the native deliverables.

The final bundle keeps the original artifacts intact. Poster files are copied into a poster folder, video and caption assets into a media folder, slide frames into a slides folder, blog blocks and images into a blog folder, and download bundles into a downloads folder. The top level remains limited to the root deliverables, including the interactive viewer, the alignment record, and the package manifest.

\subsubsection{Quality Gates}

Paper2Reel is validated as an interactive browser artifact, not only as a set of files. The static part of the package gate checks that the interactive viewer, the alignment record, the package manifest, copied poster assets, media directories, slide frames, blog blocks, wordmark assets, and download bundles exist. It rejects stale tabbed-viewer markers, machine-local path leaks, backup files, missing poster resources, missing section slides, missing section clips, missing caption sidecars, missing English or Chinese blog blocks, and missing blog figures. It also verifies that the reel uses the raw pre-subtitle video as its playback source and caption sidecars as toggleable captions.

The browser part of the gate serves the bundle with a range-capable local server, the preview server used by the skill. This is necessary because video seeking and slide-thumbnail jumps depend on partial-content responses. The gate confirms this support before it exercises the page. It then opens the viewer in a headless browser and checks poster-first loading, hover behavior, section modal opening, title modal opening, split-pane layout, subtitle toggling, slide-thumbnail seeking, direct video seeking, top-menu layout, downloads, shortcut-driven controls, and blog rendering. A separate file-browser gate checks that the same viewer page can also open directly from disk when the bundle folder stays intact.

The result is section-level convergence. Paper2Poster, Paper2Video, and Paper2Blog remain independently editable deliverables, but Paper2Reel gives them one navigable surface in which poster sections, video moments, captions, slides, and bilingual blog passages stay aligned.

\section{Experiments}\label{sec:experiments}

ResearchStudio-Reel combines artifact generation with a native-editable, section-aligned delivery contract, so its evaluation separates quantitative poster quality from capability coverage of the full pipeline. We anchor the quantitative comparison on Paper2Poster because its benchmark directly matches our poster output and includes author-created references, scoring our output against prior automated poster systems, single-shot frontier LLMs, and the authors' posters under the same protocol. We report poster quality and pipeline coverage together in Table~\ref{tbl:main}.

\textbf{Experiment setup.} We compare three families of system on the Paper2Poster benchmark \citep{pang2026paper2poster}, all originating from the same benchmark papers and scored with the same evaluation pipeline. \emph{Single-shot} baselines prompt a frontier vision-language model once to emit an entire poster in one generation: Claude-4.8 Opus, GPT-5.5, and Gemini-3.1 Pro. \emph{Poster Pipeline} systems instead run a multi-step agentic procedure: the prior Paper2Poster~\cite{pang2026paper2poster} tool, P2P~\cite{sun2025p2p}, and PosterGen~\cite{zhang2025postergen} under their released configurations, and our ResearchStudio-\reelword{} in two settings that hold the skill fixed while swapping the agent harness and its generator model. The \textbf{Claude Code} setting runs the skills inside Claude Code \citep{anthropic2025claudecode} driven by \texttt{claude-opus-4.8}; the \textbf{Codex} setting runs the identical skills inside Codex driven by \texttt{gpt-5.5}. Reporting both allows us to examine whether the workflow transfers across two model/harness configurations, without isolating model effects from harness effects. The author ground-truth poster is the human reference.

\textbf{Benchmark.} The Paper2Poster benchmark contains 100 papers \citep{pang2026paper2poster}. We generate each poster end-to-end from the PDF and score it with two VLM judges, \texttt{claude-opus-4.8} and \texttt{gpt-5.5}, reusing Paper2Poster's own six-criterion aesthetic / information rubric and its PaperQuiz reading-comprehension probe verbatim; every poster is downscaled to $\leq$2560\,px so no source is advantaged by resolution. Each poster is also scored on its complete rendering: because a model can emit HTML that declares a page smaller than its own content (one configuration wrote a 48-inch \texttt{@page} over a 60-inch layout, which would crop about a fifth of the poster), we render every method at its true content size before scoring, so each system is judged on the full poster it produced rather than a cropped view. Aesthetic and Information are the means of their three sub-criteria on a 1 to 5 scale, reported separately for the Claude and GPT judges; PaperQuiz is answer accuracy, graded by exact match against the benchmark's own fixed question set.
We define a poster's \emph{overall score} as the mean of these six aesthetic and information sub-criteria (equivalently, the mean of the two Overall columns in Table~\ref{tbl:main}); PaperQuiz, being a percentage accuracy rather than a 1--5 rating, is excluded from it and reported separately. The per-paper \emph{win rate} is a head-to-head count computed from the per-paper scores, not read off the aggregate means in Table~\ref{tbl:main}. Fixing one judge, for each of the 100 benchmark papers we compare the overall score that judge assigns to our poster against the overall score it assigns to the author ground-truth poster for the same paper, and record a \emph{win} when ours is strictly higher (a loss when strictly lower, a tie otherwise); the win rate is the number of winning papers divided by 100. In every case the judges and the PaperQuiz reader see only the rendered poster image and never the source paper, so each system is credited solely for what its poster itself conveys rather than for prior knowledge of the work.

\begin{table}[!t]
\centering
\small
\setlength{\tabcolsep}{4pt}
\caption{\textbf{ResearchStudio-Reel vs.\ baselines and author ground-truth.} Poster quality tested on the Paper2Poster benchmark \citep{pang2026paper2poster}, broken into six aesthetic/information sub-criteria (1--5, higher is better) and PaperQuiz reading comprehension, alongside pipeline capability. Each quality cell is the mean of two VLM judges (\texttt{claude-opus-4.8} and \texttt{gpt-5.5}); per-judge breakdowns are in Appendix~\ref{app:judges} (Table~\ref{tbl:judge-perjudge}). $^{\S}$~PaperQuiz answer accuracy (\%): an AI reader answers fixed questions from the poster image and is graded by exact match. For capability, \cmark~/~\xmark~= emits~/~does not emit that format; \dnmark{}~= not provided by the benchmark, which only provides PDF posters. Best quality value per column among the systems in \textbf{bold} and second best \underline{underlined}; the author ground-truth (last row) is the human reference. $^{\dagger}$Each Overall is the mean of the three sub-criteria in its Aesthetic or Information block. $^{\ast}$Baselines are reproduced with best efforts and scored by us under Claude Code with \texttt{claude-opus-4.8}.}
\vspace{0.5em}
\label{tbl:main}
\resizebox{\linewidth}{!}{%
\begin{tabular}{l cccc cccc cc c cccc}
\toprule
& \multicolumn{4}{c}{\textbf{Aesthetic}} & \multicolumn{4}{c}{\textbf{Information}} & \multicolumn{2}{c}{\textbf{Quiz}} && \multicolumn{4}{c}{\textbf{Pipeline Artifacts}} \\
\cmidrule(lr){2-5}\cmidrule(lr){6-9}\cmidrule(lr){10-11}\cmidrule(lr){13-16}
System & Elem. & Engag. & Layout & Overall$^{\dagger}$ & Low & Logic & Cont. & Overall$^{\dagger}$ & Detail$^{\S}$ & Underst.$^{\S}$ && HTML & PPTX & Video & Blog \\
\midrule
\rowcolor{black!5}\multicolumn{16}{@{}l}{\emph{Single-shot}}\\
Claude-4.8 Opus~\citep{anthropic2026opus48} & 3.10 & 2.68 & 2.97 & 2.92 & 3.92 & \underline{4.01} & \underline{3.80} & \underline{3.91} & 69.86 & 93.67 && \dnmark & \dnmark & \dnmark & \dnmark \\
GPT-5.5~\citep{openai2026gpt55} & 3.12 & 2.94 & 3.40 & 3.15 & \textbf{3.99} & 4.00 & \textbf{3.91} & \textbf{3.97} & \underline{70.89} & 94.25 && \dnmark & \dnmark & \dnmark & \dnmark \\
Gemini-3.1 Pro~\citep{google2026gemini31pro} & 3.07 & 2.89 & 3.27 & 3.08 & \underline{3.95} & 4.00 & 3.65 & 3.87 & 65.36 & 93.59 && \dnmark & \dnmark & \dnmark & \dnmark \\
\rowcolor{black!5}\multicolumn{16}{@{}l}{\emph{Poster Pipeline}}\\
Paper2Poster Tool$^{\ast}$ \citep{pang2026paper2poster} & 2.86 & 2.30 & 3.00 & 2.72 & 3.46 & 3.41 & 3.34 & 3.40 & 69.95 & \underline{95.02} && \xmark & \cmark & \xmark & \xmark \\
P2P$^{\ast}$ \citep{sun2025p2p} & 2.68 & 2.34 & 3.15 & 2.72 & 3.92 & 3.78 & 3.70 & 3.80 & \textbf{75.40} & \textbf{95.60} && \cmark & \xmark & \xmark & \xmark \\
PosterGen$^{\ast}$ \citep{zhang2025postergen} & 2.93 & 2.23 & 2.98 & 2.71 & 3.80 & 3.76 & 3.37 & 3.64 & 57.45 & 91.85 && \xmark & \cmark & \xmark & \xmark \\
\textbf{\reelword{} (Codex)} & \underline{3.16} & \underline{3.03} & \underline{3.62} & \underline{3.27} & \underline{3.95} & 3.97 & 3.09 & 3.67 & 55.55 & 92.01 && \cmark & \cmark & \cmark & \cmark \\
\textbf{\reelword{} (Claude Code)} & \textbf{3.53} & \textbf{3.17} & \textbf{3.99} & \textbf{3.56} & \textbf{3.99} & \textbf{4.03} & 3.41 & 3.81 & 56.79 & 92.16 && \cmark & \cmark & \cmark & \cmark \\
\midrule
Author ground-truth & 3.13 & 2.60 & 3.35 & 3.03 & 3.45 & 3.85 & 3.50 & 3.60 & 54.73 & 91.40 && \dnmark & \dnmark & \dnmark & \dnmark \\
\bottomrule
\end{tabular}%
}
\vspace{-1cm}
\end{table}

\textbf{Results.} As shown in Table~\ref{tbl:main}, ResearchStudio-\reelword{} (Claude Code) holds the best score among automated systems on all three aesthetic sub-criteria and the best or tied-best on two of the three information sub-criteria; its Aesthetic Overall of $3.56$ is the best among the systems, above the authors' $3.03$. Per paper, it wins the overall score on 74 of the 100 papers under the Claude judge and 95 under the GPT judge, or $85\%$ and $100\%$ of the non-tied papers. Averaging the two Overall columns, Claude Code leads the table at $3.69$; the three single-shot LLMs ($3.41$ to $3.56$) and our Codex configuration ($3.47$) form a tight second tier, all ahead of the reproduced poster pipelines ($3.06$ to $3.26$). The single-shot posters organize content well and mainly lack visual polish, whereas Codex reaches the same tier from the opposite direction, pairing top-tier aesthetics with lower content coverage. On PaperQuiz the ranking inverts, a tension between raw text coverage and visual legibility that we unpack in the analysis below.

\textbf{Analysis.} The PaperQuiz ordering is close to the reverse of the aesthetic ordering, suggesting a tension between the two objectives. PaperQuiz rewards a poster that reproduces enough paper content for an AI reader to answer questions, which may favor density, whereas the aesthetic criteria may favor selective, legible layouts. This offers one possible explanation for why the strongest reading-comprehension systems score less well visually. P2P~\cite{sun2025p2p} leads both PaperQuiz splits because its full-height portrait canvas is packed with prose lifted almost verbatim from the paper, giving the reader the most raw text to draw on, at the cost of some of the lowest aesthetic scores in the table. The Paper2Poster tool is close behind on both splits for a related reason: its content is assembled directly from the benchmark's own question-and-answer extraction, so the questions are answerable almost by construction, again while its visual scores sit among the lowest. The author posters occupy the opposite corner: a human designer prunes the paper down to a few headline results, which reads cleanly and scores well on aesthetics but discards the detail a comprehension probe rewards, so the ground truth places last on both Quiz splits. ResearchStudio-\reelword{} sits deliberately between these poles. Its measured fill loop packs each column to a target density rather than to exhaustion, a design that coincides with the strongest aesthetic scores but lower PaperQuiz scores than the denser systems. The single-shot LLM baselines are strong on content, ranking near the top on both PaperQuiz and the information criteria, but only intermediate on aesthetics; unlike the full pipeline, they do not use iterative fill control. We caution that these aesthetic margins are not an objective verdict of quality: aesthetic appeal is inherently subjective and audience-dependent, and the sub-criteria are scored by VLM judges that stand in for, but do not equal, human design preference. A higher judge score therefore means these automated raters prefer the poster, not that human viewers would necessarily agree; the aesthetic numbers should be read as a proxy signal rather than proof that our posters are genuinely better-looking.

\begin{figure*}[t]
\centering
\begin{minipage}[t]{0.32\linewidth}\centering\fbox{\includegraphics[width=\dimexpr\linewidth-2\fboxsep-2\fboxrule\relax]{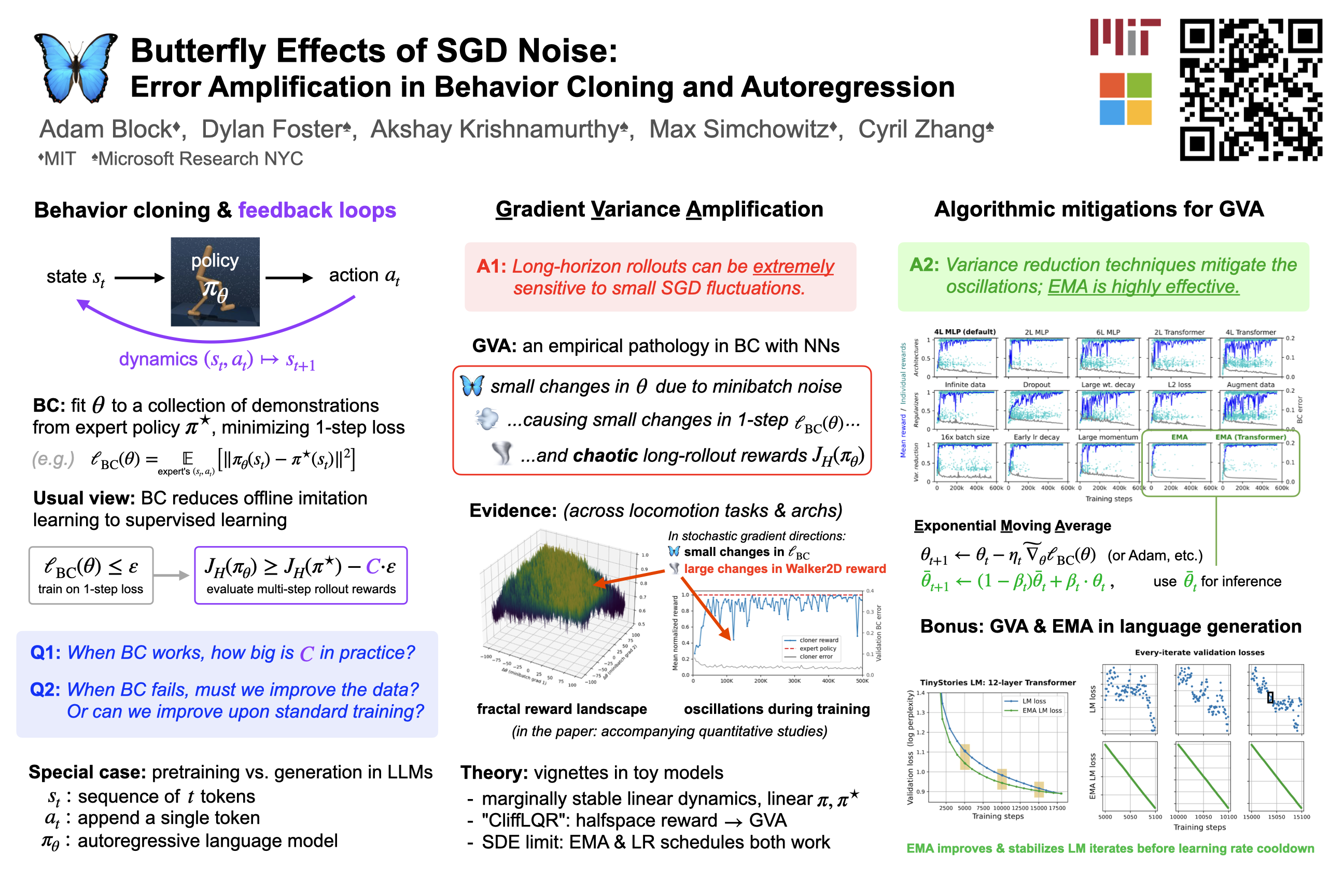}}\\[1.5pt]{\scriptsize Author ground-truth}\end{minipage}\hfill\begin{minipage}[t]{0.32\linewidth}\centering\fbox{\includegraphics[width=\dimexpr\linewidth-2\fboxsep-2\fboxrule\relax]{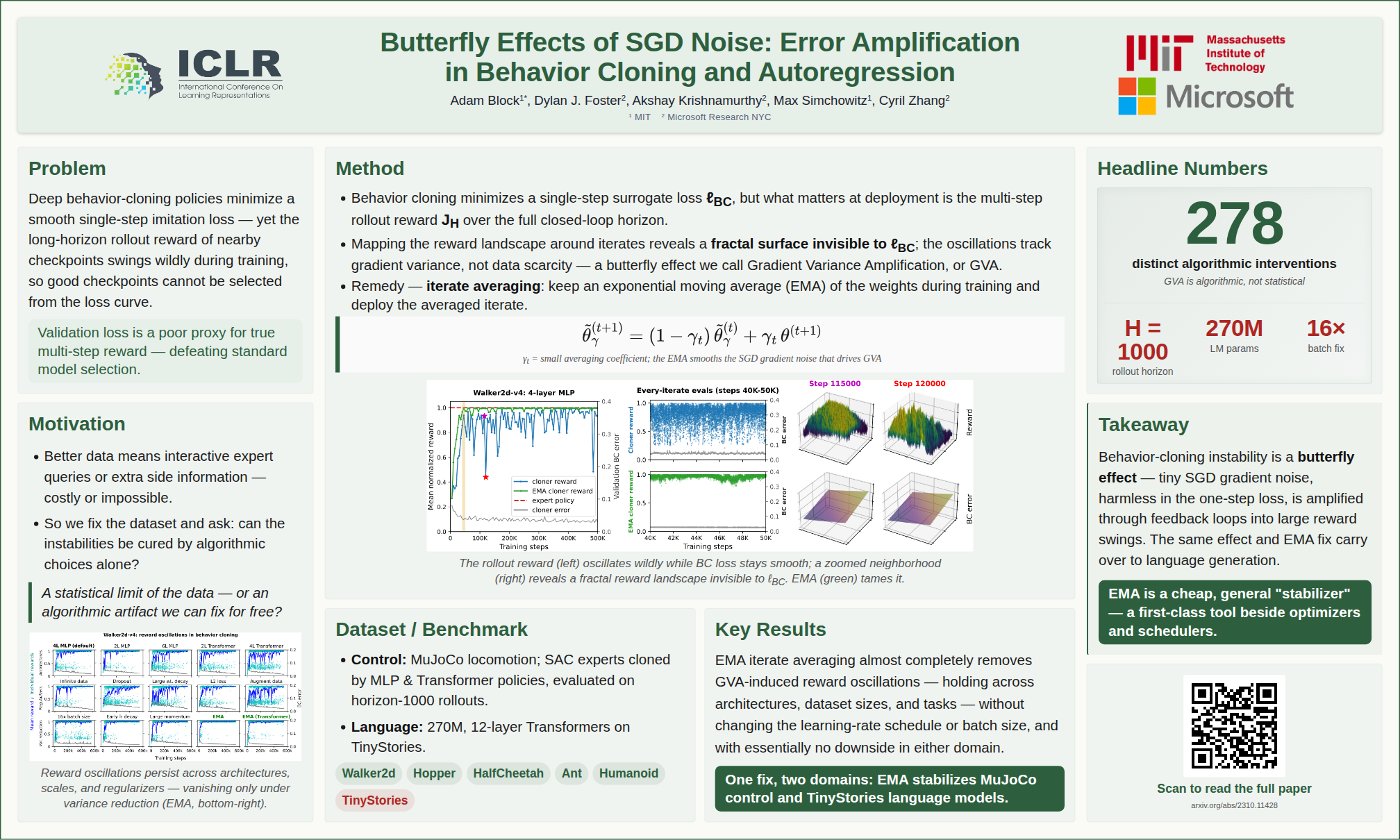}}\\[1.5pt]{\scriptsize Claude Code (\texttt{claude-opus-4.8})}\end{minipage}\hfill\begin{minipage}[t]{0.32\linewidth}\centering\fbox{\includegraphics[width=\dimexpr\linewidth-2\fboxsep-2\fboxrule\relax]{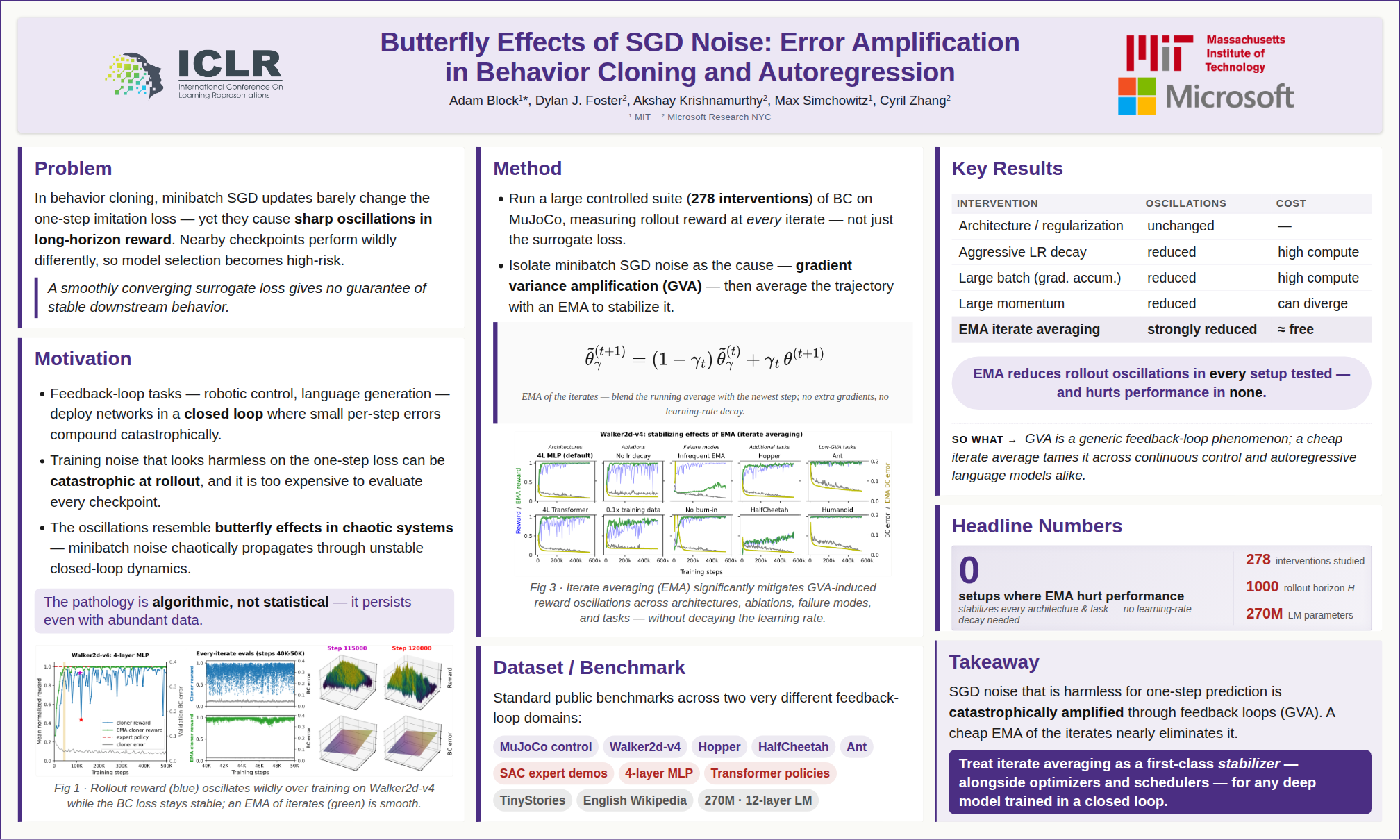}}\\[1.5pt]{\scriptsize Claude Code (\texttt{claude-opus-4.8}), \emph{max reasoning}}\end{minipage}\par\vspace{6pt}
\begin{minipage}[t]{0.32\linewidth}\centering\fbox{\includegraphics[width=\dimexpr\linewidth-2\fboxsep-2\fboxrule\relax]{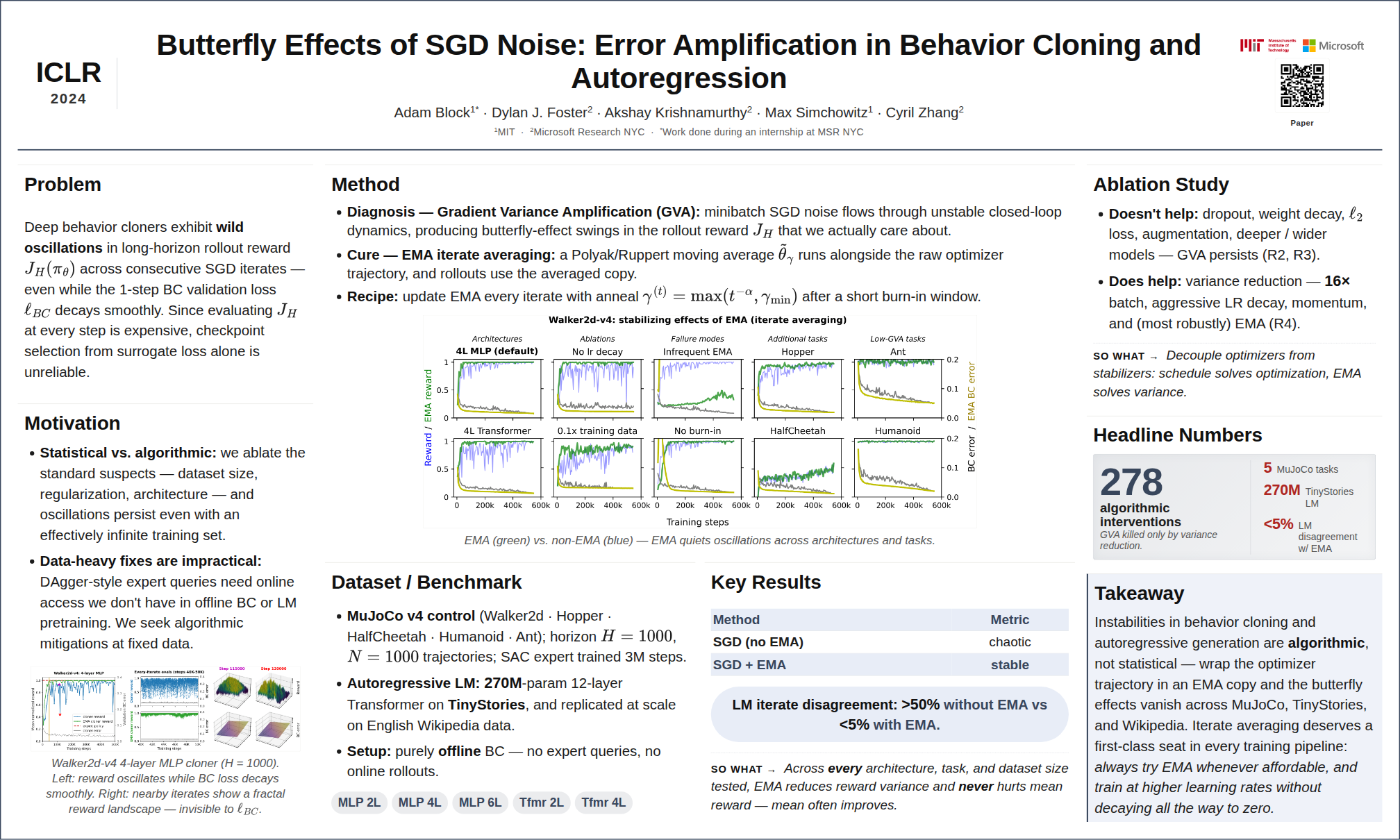}}\\[1.5pt]{\scriptsize Claude Code (\texttt{claude-opus-4.7})}\end{minipage}\hfill\begin{minipage}[t]{0.32\linewidth}\centering\fbox{\includegraphics[width=\dimexpr\linewidth-2\fboxsep-2\fboxrule\relax]{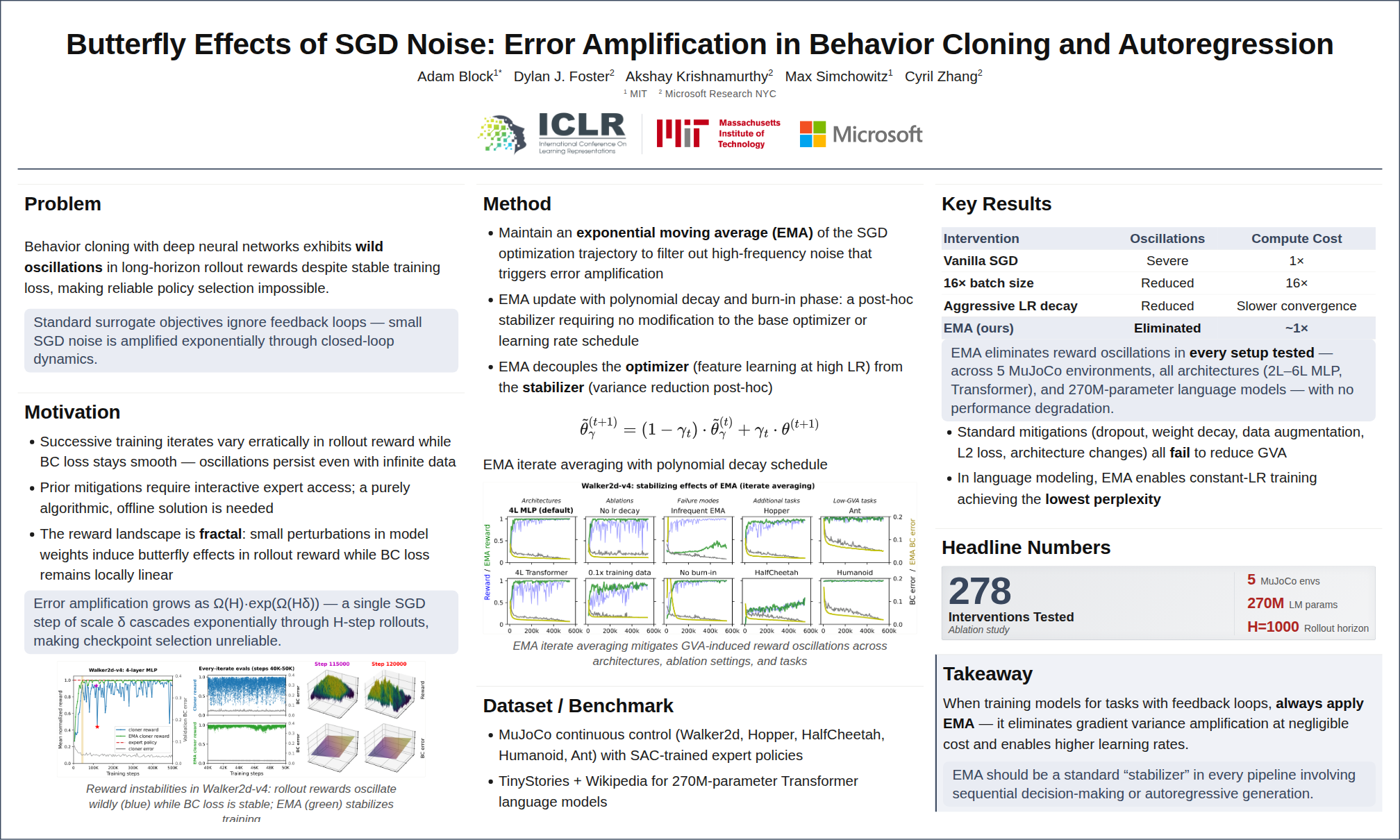}}\\[1.5pt]{\scriptsize Claude Code (\texttt{claude-opus-4.6})}\end{minipage}\hfill\begin{minipage}[t]{0.32\linewidth}\centering\fbox{\includegraphics[width=\dimexpr\linewidth-2\fboxsep-2\fboxrule\relax]{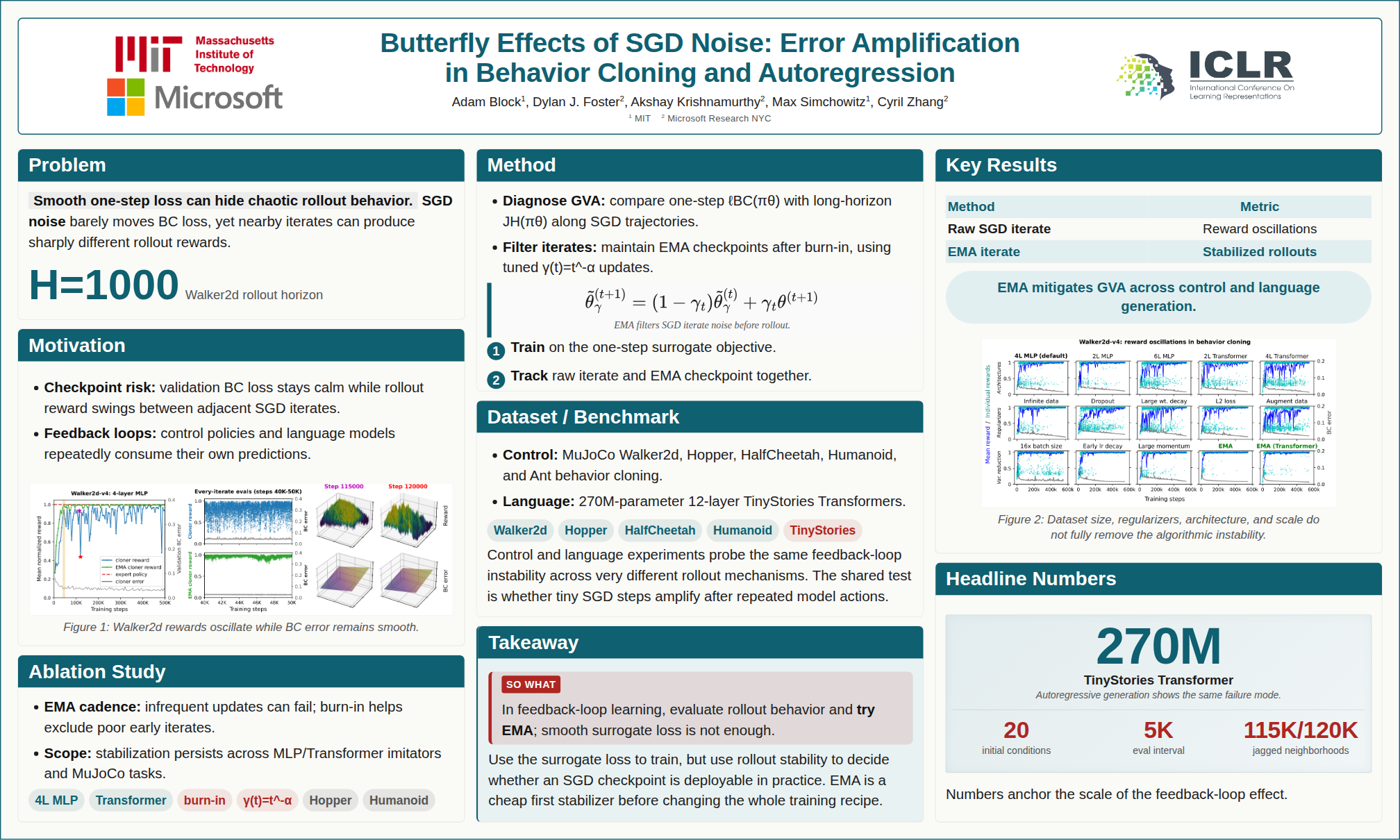}}\\[1.5pt]{\scriptsize Codex (\texttt{gpt-5.5})}\end{minipage}
\caption{Qualitative comparison across configurations. The author ground-truth (top left) is the human reference; the other five are all our method, holding the ResearchStudio-Reel skill, prompt, and pipeline fixed while varying the harness, base model, or reasoning effort. The examples differ in figure choice, phrasing density, and accent while each forms a full single page. All settings use high reasoning effort except the panel marked \emph{max reasoning} (top-right); the Codex panel runs \texttt{gpt-5.5} at its default effort.}
\label{fig:model-ablation}
\end{figure*}

\textbf{Controlled comparisons.} Two comparisons already present in Table~\ref{tbl:main} show how much the aesthetic result depends on the runtime configuration, not the skill alone. \emph{(i) The composition and fill loop, under Claude.} \texttt{claude-opus-4.8} prompted once for a full poster reaches an aesthetic mean of $2.92$ and a Layout score of $2.97$, whereas the same model wrapped in the composition step and the measured-fill loop, our Claude Code setting, reaches $3.56$ aesthetic and $3.99$ Layout; this gain of $0.64$ on aesthetics and over a full point on Layout is the payoff of the loop, although the comparison also changes prompting, the number of model calls, and inference budget and so does not isolate it. \emph{(ii) The same skill under a different harness and model.} Driving the identical skill with Codex and \texttt{gpt-5.5} still produces a strong poster: rendered at full size, the Codex configuration reaches an aesthetic mean of $3.27$ (per-judge $3.29$ and $3.25$), the second-highest aesthetic score in Table~\ref{tbl:main}, ahead of every single-shot baseline (the strongest, \texttt{gpt-5.5}, sits at $3.15$), every reproduced poster pipeline, and the author reference at $3.03$; it trails only the Claude Code configuration ($3.56$), by about $0.3$. The measured-fill loop therefore transfers across the harness and model swap, and the remaining margin reflects finer visual hierarchy in the Claude Code renders rather than a failure of the skill under Codex. Figure~\ref{fig:model-ablation} shows this qualitatively.

\textbf{Gap to human posters.} Matching or exceeding the aggregate judge scores does not mean the output is on par with an author's own poster. A human designer routinely adds material the source paper does not contain (a purpose-drawn method or overview diagram, or explanatory icons) to carry the narrative, whereas our pipeline is grounded in the extracted asset bundle and reuses only the figures and content that actually appear in the paper, so it does not fabricate such bespoke visuals. Author taste also spans a wide and legitimate range: some authors favour large type with concise text, others a dense information layout, and we make no claim to match any individual author's preference. We instead analyse each paper's figure and content profile and choose the layout that best fits it, aiming for a sensible middle ground rather than any single house style.

\begin{table}[t]
\centering
\begin{minipage}[t]{0.48\linewidth}
\centering
\footnotesize
\caption{\textbf{Paper2Video capability audit.} We compare if the resulting videos contain visual attention cues or cursor grounding, an editable PPTX deck, duration control, and narration audio.}
\label{tbl:p2v-capabilities}
\vspace{0.5em}
\setlength{\tabcolsep}{2.5pt}
\resizebox{\linewidth}{!}{%
\begin{tabular}{@{}lcccc@{}}
\toprule
System & Visual Cue & PPTX & Duration & Audio \\
\midrule
\rowcolor{black!5}\multicolumn{5}{@{}l}{\emph{Products}}\\
\quad Deck Tools$^{1}$ & \xmark & \cmark & \xmark & \xmark \\
\quad Video Tools$^{2}$ & \xmark & \xmark & \cmark & \cmark \\
\rowcolor{black!5}\multicolumn{5}{@{}l}{\emph{Research Papers}}\\
\quad VideoAgent~\citep{liang2026videoagent} & \xmark & \cmark & \xmark & \cmark \\
\quad PaperTalker~\citep{zhu2025paper2video} & \cmark & \xmark & \cmark & \cmark \\
\quad Preacher~\citep{liu2025preacher} & \cmark & \xmark & \cmark & \cmark \\
\quad PresentAgent~\citep{shi2025presentagent, wu2026presentagent2} & \xmark & \cmark & \cmark & \cmark \\
\midrule
\rowcolor{barcol!15}\textbf{ResearchStudio-\reelword}  & \cmark & \cmark & \cmark & \cmark  \\
\bottomrule
\end{tabular}%
}
\end{minipage}
\hfill
\begin{minipage}[t]{0.497\linewidth}
\centering
\footnotesize
\caption{\textbf{Paper2Blog capability audit.} We compare whether the reviewed paper or product documentation explicitly reports paper input, blog-style summarization, embedded figures, editable DOCX output, and layout checks.}
\label{tbl:p2b-capabilities}
\vspace{0.5em}
\setlength{\tabcolsep}{2.5pt}
\resizebox{\linewidth}{!}{%
\begin{tabular}{@{}lcccc@{}}
\toprule
System & Layout & Figures & DOCX & Summary \\
\midrule
\rowcolor{black!5}\multicolumn{5}{@{}l}{\emph{Products}}\\
\quad Semantic Scholar TLDR$^{3}$  & \xmark & \xmark & \xmark & \cmark \\
\quad Research Assistants$^{4}$  & \xmark & \xmark & \cmark & \cmark \\
\quad Scholarcy$^{5}$ & \xmark & \xmark & \cmark & \cmark \\
\quad NotebookLM$^{6}$  & \xmark & \xmark & \xmark & \cmark \\
\rowcolor{black!5}\multicolumn{5}{@{}l}{\emph{Research Papers}}\\
\quad Papers-to-Posts~\citep{radensky2024posts} & \xmark & \xmark & \xmark & \cmark \\
\quad PaperX~\citep{yu2026paperx} & \xmark & \cmark & \xmark & \cmark \\
\midrule
\rowcolor{barcol!15}\textbf{ResearchStudio-\reelword}  & \cmark & \cmark & \cmark & \cmark \\
\bottomrule
\end{tabular}%
}
\end{minipage}
\end{table}

\begingroup
\renewcommand{\thefootnote}{}
\footnotetext{\textsuperscript{1} AI presentation generators, e.g., \href{https://gamma.app/}{Gamma} and \href{https://www.canva.com/create/ai-presentations/}{Canva}. \textsuperscript{2} AI avatar/video generators, e.g., \href{https://www.synthesia.io/}{Synthesia} and \href{https://www.heygen.com/}{HeyGen}. \textsuperscript{3} Semantic Scholar's short paper-summary feature. \textsuperscript{4} Paper search and reading tools, e.g., \href{https://elicit.com/}{Elicit}, \href{https://scispace.com/}{SciSpace}, and \href{https://consensus.app/}{Consensus}. \textsuperscript{5} PDF summary and flashcard tool. \textsuperscript{6} Google's source-grounded notebook assistant.}
\endgroup

\textbf{Capability coverage.} Tables~\ref{tbl:p2v-capabilities} and~\ref{tbl:p2b-capabilities} audit the two video/blog generators against representative commercial products and prior research systems. The systems differ in what they can \emph{emit}: the video and blog systems in this audit generally focus on a single artifact, whereas ResearchStudio-Reel emits all three dissemination artifacts, exposes the shared intermediate \emph{artifacts} bundle, includes a native-editable source file in each artifact package, and stitches the three into one navigable surface. The poster scores in Table~\ref{tbl:main} rate rendered quality for a single artifact, whereas this audit documents the breadth of the delivery contract; it does not measure post-generation editing effort or the usability benefit of the aligned viewer.

\begin{table*}[t]\centering\scriptsize
\setlength{\tabcolsep}{4.5pt}\renewcommand{\arraystretch}{1.2}
\caption{\textbf{Per-stage breakdown of the full ResearchStudio pipeline}, mean over 5 papers on \texttt{claude-opus-4-8}. ``Input'' is distinct content tokens (fresh $+$ cache writes); ``Cache'' is cached-context re-reads, billed at ${\sim}$10\% of fresh input; ``Output'' is prorated generated tokens. Bars give each stage's share of the pipeline total, and the \textbf{bold} stage is each skill's heaviest. Each skill runs as an isolated subprocess, so its one-time context load (reading the paper and its figures) is folded into its first stage. A full four-artifact bundle from one PDF takes ${\sim}89$ minutes and ${\sim}2.6$M input / ${\sim}276$K output tokens per paper.}
\label{tbl:pipeline-breakdown}
\begin{tabular}{@{}ll r r r l r r l@{}}
\toprule
Skill & Stage & Time\,(min) & Turns & Input Tokens\,(K) & Share & Cache\,(K) & Output Tokens\,(K) & Share \\
\midrule
\multirow{1}{*}{Paper2Assets} & \textbf{extract \& figures} & 8.6 & 112 & 461 & \makebox[20pt][r]{18\%}\,{\color{barcol}\rule{7.5pt}{5pt}} & 12{,}068 & 33 & \makebox[20pt][r]{12\%}\,{\color{barcol}\rule{5.0pt}{5pt}} \\
\midrule
\multirow{5}{*}{Paper2Poster} & compose & 5.1 & 43 & 191 & \makebox[20pt][r]{7\%}\,{\color{barcol}\rule{3.1pt}{5pt}} & 7{,}989 & 12 & \makebox[20pt][r]{4\%}\,{\color{barcol}\rule{1.8pt}{5pt}} \\
 & \textbf{fill loop} & 14.8 & 100 & 310 & \makebox[20pt][r]{12\%}\,{\color{barcol}\rule{5.1pt}{5pt}} & 25{,}185 & 48 & \makebox[20pt][r]{18\%}\,{\color{barcol}\rule{7.4pt}{5pt}} \\
 & render & 1.9 & 19 & 18 & \makebox[20pt][r]{$<$1\%}\,{\color{barcol}\rule{0.3pt}{5pt}} & 5{,}527 & 6 & \makebox[20pt][r]{2\%}\,{\color{barcol}\rule{0.9pt}{5pt}} \\
 & narration audio & 1.5 & 3 & 5 & \makebox[20pt][r]{$<$1\%}\,{\color{barcol}\rule{0.1pt}{5pt}} & 916 & 1 & \makebox[20pt][r]{$<$1\%}\,{\color{barcol}\rule{0.1pt}{5pt}} \\
 & \cellcolor{barcol!15} \emph{subtotal} & \cellcolor{barcol!15} \emph{23.3} & \cellcolor{barcol!15} \emph{166} & \cellcolor{barcol!15} \emph{523} & \cellcolor{barcol!15} \makebox[20pt][r]{20\%}\,{\color{barcol}\rule{8.6pt}{5pt}} & \cellcolor{barcol!15} \emph{39{,}616} & \cellcolor{barcol!15} \emph{67} & \cellcolor{barcol!15} \makebox[20pt][r]{24\%}\,{\color{barcol}\rule{10.2pt}{5pt}} \\
\midrule
\multirow{7}{*}{Paper2Video} & \textbf{script \& cue spec} & 4.7 & 49 & 281 & \makebox[20pt][r]{11\%}\,{\color{barcol}\rule{4.6pt}{5pt}} & 5{,}385 & 20 & \makebox[20pt][r]{7\%}\,{\color{barcol}\rule{3.1pt}{5pt}} \\
 & deck (ppt-master) & 5.7 & 38 & 90 & \makebox[20pt][r]{4\%}\,{\color{barcol}\rule{1.5pt}{5pt}} & 6{,}665 & 16 & \makebox[20pt][r]{6\%}\,{\color{barcol}\rule{2.4pt}{5pt}} \\
 & visual cues & 1.3 & 16 & 21 & \makebox[20pt][r]{$<$1\%}\,{\color{barcol}\rule{0.3pt}{5pt}} & 2{,}925 & 6 & \makebox[20pt][r]{2\%}\,{\color{barcol}\rule{0.9pt}{5pt}} \\
 & narration audio & 3.4 & 21 & 33 & \makebox[20pt][r]{1\%}\,{\color{barcol}\rule{0.5pt}{5pt}} & 2{,}857 & 7 & \makebox[20pt][r]{3\%}\,{\color{barcol}\rule{1.1pt}{5pt}} \\
 & render \& mux & 9.2 & 38 & 43 & \makebox[20pt][r]{2\%}\,{\color{barcol}\rule{0.7pt}{5pt}} & 7{,}149 & 17 & \makebox[20pt][r]{6\%}\,{\color{barcol}\rule{2.6pt}{5pt}} \\
 & QA gate & 4.0 & 30 & 44 & \makebox[20pt][r]{2\%}\,{\color{barcol}\rule{0.7pt}{5pt}} & 6{,}315 & 14 & \makebox[20pt][r]{5\%}\,{\color{barcol}\rule{2.1pt}{5pt}} \\
 & \cellcolor{barcol!15} \emph{subtotal} & \cellcolor{barcol!15} \emph{28.5} & \cellcolor{barcol!15} \emph{193} & \cellcolor{barcol!15} \emph{512} & \cellcolor{barcol!15} \makebox[20pt][r]{20\%}\,{\color{barcol}\rule{8.4pt}{5pt}} & \cellcolor{barcol!15} \emph{31{,}297} & \cellcolor{barcol!15} \emph{80} & \cellcolor{barcol!15} \makebox[20pt][r]{29\%}\,{\color{barcol}\rule{12.2pt}{5pt}} \\
\midrule
\multirow{4}{*}{Paper2Blog} & \textbf{figures \& setup} & 6.6 & 47 & 524 & \makebox[20pt][r]{20\%}\,{\color{barcol}\rule{8.6pt}{5pt}} & 4{,}849 & 17 & \makebox[20pt][r]{6\%}\,{\color{barcol}\rule{2.6pt}{5pt}} \\
 & DOCX assembly & 0.1 & 2 & 4 & \makebox[20pt][r]{$<$1\%}\,{\color{barcol}\rule{0.1pt}{5pt}} & 336 & 1 & \makebox[20pt][r]{$<$1\%}\,{\color{barcol}\rule{0.1pt}{5pt}} \\
 & QA gate \& revision & 10.0 & 63 & 243 & \makebox[20pt][r]{9\%}\,{\color{barcol}\rule{4.0pt}{5pt}} & 11{,}062 & 47 & \makebox[20pt][r]{17\%}\,{\color{barcol}\rule{7.1pt}{5pt}} \\
 & \cellcolor{barcol!15} \emph{subtotal} & \cellcolor{barcol!15} \emph{16.7} & \cellcolor{barcol!15} \emph{112} & \cellcolor{barcol!15} \emph{771} & \cellcolor{barcol!15} \makebox[20pt][r]{30\%}\,{\color{barcol}\rule{12.6pt}{5pt}} & \cellcolor{barcol!15} \emph{16{,}247} & \cellcolor{barcol!15} \emph{64} & \cellcolor{barcol!15} \makebox[20pt][r]{23\%}\,{\color{barcol}\rule{9.8pt}{5pt}} \\
\midrule
\multirow{4}{*}{Paper2Reel} & \textbf{plan} & 3.2 & 24 & 198 & \makebox[20pt][r]{8\%}\,{\color{barcol}\rule{3.2pt}{5pt}} & 1{,}910 & 7 & \makebox[20pt][r]{2\%}\,{\color{barcol}\rule{1.0pt}{5pt}} \\
 & assemble & 2.7 & 16 & 26 & \makebox[20pt][r]{$<$1\%}\,{\color{barcol}\rule{0.4pt}{5pt}} & 1{,}359 & 4 & \makebox[20pt][r]{1\%}\,{\color{barcol}\rule{0.6pt}{5pt}} \\
 & QA gate & 6.2 & 52 & 77 & \makebox[20pt][r]{3\%}\,{\color{barcol}\rule{1.3pt}{5pt}} & 6{,}049 & 21 & \makebox[20pt][r]{8\%}\,{\color{barcol}\rule{3.2pt}{5pt}} \\
 & \cellcolor{barcol!15} \emph{subtotal} & \cellcolor{barcol!15} \emph{12.1} & \cellcolor{barcol!15} \emph{92} & \cellcolor{barcol!15} \emph{301} & \cellcolor{barcol!15} \makebox[20pt][r]{12\%}\,{\color{barcol}\rule{4.9pt}{5pt}} & \cellcolor{barcol!15} \emph{9{,}318} & \cellcolor{barcol!15} \emph{32} & \cellcolor{barcol!15} \makebox[20pt][r]{12\%}\,{\color{barcol}\rule{4.8pt}{5pt}} \\
\midrule
\multicolumn{2}{@{}l}{\textbf{Full pipeline}} & \textbf{89.2} & \textbf{675} & \textbf{2{,}568} & \makebox[20pt][r]{100\%}\,{\color{barcol}\rule{42.0pt}{5pt}} & \textbf{108{,}546} & \textbf{276} & \makebox[20pt][r]{100\%}\,{\color{barcol}\rule{42.0pt}{5pt}} \\
\bottomrule
\end{tabular}
\vspace{-0.5cm}
\end{table*}

\textbf{Operational profile.} We instrument every stage of the full pipeline with timestamped API traces; Table~\ref{tbl:pipeline-breakdown} reports per-stage wall-clock time and token usage, averaged over five sampled papers. The one-time Paper2Assets extraction is shared across every downstream skill. In these runs, Paper2Poster and Paper2Video have the largest generator subtotals: the measured-fill loop is Paper2Poster's longest stage, while Paper2Video spends most of its recorded time across deck construction, rendering and muxing, and QA. Paper2Blog is driven by its bilingual revision gate, and Paper2Reel has the smallest generator subtotal. Producing all four artifacts from one PDF takes about an hour and a half of summed stage runtime per paper, with token totals detailed in the caption; running the three generators in parallel can reduce elapsed wall-clock time. The Cache column reports cached-context re-reads separately from distinct input and output tokens. When the paper's LaTeX source is available, the pipeline can instead read the original figure files and captions, potentially reducing extraction time; the PDF path remains the default otherwise.

\section{Applications}\label{sec:applications}

Three application contexts drove the design of ResearchStudio-Reel and are the ones we expect it to serve first: the accepted-paper author working the camera-ready last mile, the lab or research organization wiring dissemination into its publication-intake pipeline, and the graduate course or reading group turning a paper list into weekly briefing packs. These three are not arbitrary: they deliberately span the dissemination spectrum, from a one-off, single-paper polish for the individual author, to a continuous per-submission feed for the organization, to a periodic many-paper batch for the classroom, so together they exercise the same pipeline at very different cadences and scales.

All use the same shared Paper2Assets extraction (\S\ref{sec:p2a}), native-editable PowerPoint and Word artifacts, and Paper2Reel viewer (\S\ref{sec:connect});
The differences below are therefore about \emph{who} runs the pipeline, \emph{which artifacts of the bundle they lean on}, and \emph{what they revise afterwards}, not about a different system per use case.

\subsection{The Author's Camera-ready Last Mile}\label{sec:app-camera}
The canonical case is an accepted-paper author in the narrow window between camera-ready and the venue. Within days they need a print-ready conference poster, a talk video for the virtual track or the lab's YouTube channel, and a public-facing blog piece to announce the work on social media. ResearchStudio-Reel collapses this into a single Claude Code session per paper: one extraction pass populates the Paper2Assets bundle, then the three generators run against it to emit an editable \texttt{.pptx}, a synchronized \texttt{video.pptx}$+$\texttt{.mp4} pair, and a bilingual Word \texttt{.docx} pair that the author can hand-revise rather than regenerate, all navigable through one Paper2Reel \texttt{reel.html} that doubles as a review surface the author can share with co-authors before shipping.

\subsection{Lab- and Org-level Scientific Communication}\label{sec:app-org}
A research lab or industrial-research group can wire ResearchStudio-Reel into its publication-intake pipeline so that every accepted paper auto-generates draft dissemination artifacts, which an editorial or communications team then polishes. The editable PowerPoint and Word outputs are load-bearing here: they turn the system into a draft-and-revise workflow rather than a black-box render, which is what makes hand-off to a non-author reviewer practical. The bilingual blog primitive (\S\ref{sec:p2b}) covers a Chinese-WeChat and an English-research-blog register from one shared evidence map, and the Paper2Reel viewer gives the comms team one interactive surface per paper to review before publication rather than three files to open in three tools.

\subsection{Educational and Pedagogical Reuse}\label{sec:app-edu}
A graduate course, a reading group, or a public-understanding-of-science outlet can feed a reading list through ResearchStudio-Reel to produce \emph{paper-of-the-week briefing packs}: an editable slide deck for lecture, a talk video for asynchronous students, and a Word blog piece for the course site or newsletter, all cross-linked through one Paper2Reel viewer that students can scrub between poster, video, and blog while reviewing. Because the artifacts are editable, an instructor can adjust framing, drop a slide, or splice in a course-specific example without losing the rest of the auto-generated structure, and because the shared Paper2Assets bundle is versioned per paper, re-running against a corrected or updated PDF re-issues the whole pack without manual re-extraction.

\section{Related Works}\label{sec:related}

Prior work spans artifact-specific generation for posters, slides and videos, and long-form articles, as well as increasingly unified presentation suites. We organize it by output type, then discuss the agent and skill mechanisms that support these workflows and the multi-artifact systems closest to ResearchStudio-\reelword{}.

\subsection{Paper-to-poster Systems}\label{sec:rw-poster}

Paper2Poster~\citep{pang2026paper2poster} introduces PosterAgent, a top-down visual-in-the-loop pipeline, together with a poster benchmark, VLM-based criteria, PaperQuiz, and editable PPTX output. PosterForest~\citep{choi2026posterforest} uses a hierarchical Poster Tree for joint content and layout reasoning; P2P~\citep{sun2025p2p} separates visual processing, content generation, and HTML poster assembly and contributes a fine-grained benchmark; and PosterGen~\citep{zhang2025postergen} assigns parsing, curation, layout, styling, and rendering to specialized agents, including explicit color and typography stages. EfficientPosterGen~\citep{tang2026efficientpostergen} combines semantic retrieval and token compression with deterministic layout-violation detection, while APEX~\citep{shi2026apex} addresses interactive editing of existing PPTX posters. SciPostLayout~\citep{tanaka2024scipostlayout} provides paired poster-layout data, and SciPostGen~\citep{inadumi2026scipostgen} studies paper-conditioned, retrieval-guided layout generation rather than full PPTX authoring. Any2Poster~\citep{vinaykumar2026any2poster} expands evaluation across source modalities and content domains. Concurrent work, PosterHarness~\citep{yang2026posterharness}, uses a placeholder-first contract and deterministic composition to make instruction following and source-figure provenance auditable. Chart-Plot~\citep{tang2026demonstrating} addresses the complementary problem of producing publication-ready academic charts. PosterVerse~\citep{liu2026posterverse} is adjacent work on general commercial poster generation with an HTML typography engine, not a scientific paper-to-poster system. These works use different output contracts and evaluation regimes: Paper2Poster and PosterGen provide PPTX output, APEX edits PPTX, P2P renders HTML, and several works use VLM scores for evaluation rather than as runtime release gates. Paper2Poster in ResearchStudio-Reel instead packages HTML, PDF, PNG, audio narration, and PPTX around a shared upstream asset bundle and a categorical render check.

\subsection{Paper-to-video and Slide-generation Systems}\label{sec:rw-video}

Slide generation progresses from PPSGen~\citep{hu2014ppsgen}'s learned sentence selection, through D2S~\citep{sun2021d2s}'s query-based retrieval and summarization and DOC2PPT~\citep{fu2022doc2ppt}'s multimodal content and layout generation, to SlideSpawn~\citep{kumar2024slidespawn}'s structure-aware paper summarization. PPTAgent~\citep{zheng2025pptagent} uses an edit-based workflow and introduces PPTEval, while SlideGen~\citep{liang2025slidegen} coordinates multimodal agents to produce editable scientific PPTX decks.
Building on slide generation, scientific presentation-video systems additionally coordinate visual progression, narration, captions, and timing.
PresentAgent~\citep{shi2025presentagent, wu2026presentagent2} generates synchronized slide frames, narration, and optional captions from long documents. VideoAgent~\citep{liang2026videoagent} combines PPTX slides, Manim animations, narration, subtitles, and user-specified duration and style. Paper2Video~\citep{zhu2025paper2video} contributes a benchmark of 101 papers paired with author videos and slides, and its PaperTalker system generates LaTeX-based slides with cursor grounding, subtitles, speech, and a talking head. Preacher~\citep{liu2025preacher} plans heterogeneous key scenes and synthesizes them with domain-specific rendering tools. AutoLectures~\citep{holmberg2025generating} instead starts from an existing deck and aligns narration with synchronized visual highlights. Paper2Video in ResearchStudio-Reel emphasizes a particular deliverable contract: an editable deck, captioned and subtitle-free videos, timed visual cues, and a section-addressable timeline retained for downstream navigation.

\subsection{Paper-to-blog and Long-form Summarization}\label{sec:rw-blog}

HERA~\citep{li2025hera} repackages and reorders event-related context for long-document summarization; PTSPI~\citep{devi2025long} aligns page-level source content with target summaries and learns page importance; and GoSum~\citep{bian2024gosum} uses discourse graphs and reinforcement learning for extractive scientific-document summarization. LongDPO~\citep{ping2025longdpo} addresses the related but distinct problem of improving general long-form generation through critique-augmented process supervision. Other work changes language or audience: ProjectMundo~\citep{kleidermacher2026science} translates scientific articles while preserving JATS XML structure, and research on scientific lay summaries also examines accessibility and accuracy \citep{blais2025bridging}. Papers-to-Posts~\citep{radensky2024posts} is the closest blog-authoring precedent: its interactive reverse source outline lets users control which paper details enter a research blog piece. Paper2Blog differs through editable Word output with layout-aware checks for pagination and figure placement, applied to paired Chinese and English articles.

\subsection{Agent Frameworks and Skills}\label{sec:rw-agents}

Agentic methods, frameworks, and APIs provide the execution substrate: ReAct~\citep{yao2022react} interleaves reasoning and actions; AutoGen~\citep{wu2023autogen} supports configurable multi-agent conversations; MetaGPT~\citep{hong2024metagpt} encodes role-based standard operating procedures; and function-calling and tool-use APIs expose external actions to models \citep{openai2023functions,anthropic2024toolUse}. Voyager~\citep{wang2023voyager} demonstrates an executable skill library for an embodied agent, while Skill-as-Pseudocode~\citep{li2026skill} rewrites prose skill libraries into typed pseudocode contracts. ResearchStudio-Reel runs on the Claude Code and Codex skill mechanisms \citep{anthropic2025claudecode,anthropic2025skills,openai2025codex}; the runtime itself is not our contribution. Adjacent evaluation work includes Cookie-Bench~\citep{yang2026cookie}, which uses agent-driven interaction to assess generated web applications. Auto-research studies evaluate end-to-end research agents and expose gaps between plausible manuscripts and experimental evidence \citep{zhang2026far}, while ResearchStudio-Idea applies a skill suite to the upstream task of evidence-grounded research ideation \citep{zhao2026researchstudio}.

\subsection{Positioning}\label{sec:rw-positioning}

PaperX~\citep{yu2026paperx} uses a Scholar DAG to generate PPTs, posters, and public-relations posts. Concurrent OmniPresent~\citep{ma2026omnipresent} generates poster, slide, and video suites from a centralized HTML representation with cross-modal verify-and-repair. We discuss OmniPresent qualitatively rather than include it as an experimental baseline.

Rather than claiming the first multi-artifact generator, ResearchStudio-Reel focuses on a native-editable, section-aligned dissemination workspace. It produces an editable PowerPoint poster, video deck, and bilingual Word blog from one shared asset bundle. A section-level alignment contract maps poster regions, video segments, and blog passages, which Paper2Reel binds into one interactive viewer. The distinction is native editability and experience-level convergence, supported by artifact-specific release checks.

\section{Future Work}\label{sec:future}

Two gaps matter more than broader coverage. First, our evaluation is proxy-bound: the aesthetic rubric and PaperQuiz pull in opposite directions, since a denser poster wins comprehension while a cleaner one wins aesthetics, and neither proxy measures whether a reader actually absorbs the work. The measured-fill loop optimizes a geometric density target, not understanding, so the honest next step is to close the loop on a controlled human reading-and-recall signal rather than on a proxy that a denser poster can always game. Second, one remaining gap to author posters is generative: because the pipeline reuses only figures that already exist in the paper, it cannot yet produce the bespoke method or overview diagrams that human designers may add to carry the narrative. Addressing this gap may require faithful figure synthesis, which reintroduces the very hallucination risk the categorical gates were built to suppress, and would push the gate discipline from layout onto the factual content of generated visuals.

\section{Conclusion}\label{sec:conclusion}

ResearchStudio-Reel reframes the last mile of research dissemination as a native-editable, section-aligned workspace. Implemented as five composable skills, it uses one Paper2Assets bundle to produce a native-editable PowerPoint poster and video deck, plus a native-editable bilingual Word blog, then binds their sections, slides, video times, captions, and passages through Paper2Reel.
On the 100-paper Paper2Poster benchmark, our posters achieve the best scores among automated systems on all three aesthetic sub-criteria and the best or tied-best scores on two of three information sub-criteria under two VLM judges, exceed the authors' own on aesthetics ($3.56$ vs.\ $3.03$), and win on overall quality on 74 and 95 of the 100 papers under the two judges; the capability audits in \S\ref{sec:experiments} further document native-editable source files and aligned deliverables across all three artifact packages.
This delivery-contract pattern may extend to other dissemination targets with native authoring formats, deterministic renderers, and testable package criteria, although its evaluative, generative, and architectural limits remain to be studied (\S\ref{sec:future}).
By keeping generated artifacts editable and connected after generation, ResearchStudio-Reel shifts the last mile from one-way automation toward a workflow authors can inspect, navigate, and revise.

\bibliographystyle{unsrtnat}
\bibliography{refs}

\clearpage
\appendix

\section{Limitations}\label{app:limit}

\textbf{Recurring failure modes.} Five failure modes recur in end-to-end runs, each with a one-sentence mitigation. \textbf{(L1) Figure-cleanup residue:} a caption strip or body-text slice baked into a raster survives Paper2Assets's deterministic prefix; the visual-AI decaption pass plus fresh-context verifier (\S\ref{sec:p2a}) re-crops before any downstream skill embeds the figure. \textbf{(L2) Fill-loop non-convergence:} the poster's discrete loop can oscillate across the 90--98\% band when no move in the closed catalogue lands a section in-band; the on-disk round counter and circuit breaker (\S\ref{sec:p2p}) ship the best-measured state and let the render-time whitespace expand recover the rest. \textbf{(L3) Slide--narration referential drift:} the script says ``Figure 3'' while the slide shows Figure 2 after a panel reorder; the shared alignment timeline of \S\ref{sec:p2v} lets the final QA gate reject the render. \textbf{(L4) Bilingual blog drift:} the two language drafts disagree on a numeric result, benchmark name, or affiliation; the shared evidence map plus layout-aware DOCX gate (\S\ref{sec:p2b}) cross-check numeric claims, technical terms, and figure order across the pair. \textbf{(L5) Voice-mismatched narration:} the default Edge TTS \citep{edgetts} voice reads a keynote-style deck flat; re-run the narration step (\S\ref{sec:p2v}) with a different voice, e.g.\ \texttt{en-US-GuyNeural} or \texttt{en-US-AriaNeural}.

\textbf{Domain scope.} The system is calibrated on ML, CV, and NLP venues, where paper layout, figure conventions, and the claim-evidence DAG are relatively uniform; transfer to biomedicine, physics, or design-heavy fields, where poster conventions and the typical evidence structure differ, is untested and each skill's move catalogue may need extending. Because the underlying primitives (PDF extraction, HTML layout, DOCX rendering, and Edge TTS) are domain-agnostic, we expect the skills-as-architecture pattern to generalize, but each new venue family should be validated end-to-end before deployment.

\textbf{Evaluation coverage.} Quantitative evaluation in current version is reported only for the poster: Table~\ref{tbl:main} uses the Paper2Poster benchmark under two VLM judges. Although Paper2Video provides a related video benchmark \citep{zhu2025paper2video}, we have not evaluated ResearchStudio-Reel under that protocol; the video and blog are compared only on capability coverage in Tables~\ref{tbl:p2v-capabilities} and~\ref{tbl:p2b-capabilities}. These audits document output contracts but do not measure human editing effort, fidelity after edits, or whether section-level navigation improves understanding. RRP (\S\ref{sec:p2p}) is currently an in-loop signal rather than a headline metric, and a third-party human-rater study across all three artifacts is left for future work.

\section{Ethics}\label{app:ethics}

\textbf{Provenance and disclosure.} AI-generated posters, videos, and blogs can amplify both correct understanding and confident misreadings of the underlying science, and a narrated talk video with burned-in subtitles is not obviously AI-produced at a glance. We therefore recommend that downstream deployments preserve provenance metadata (which skill and version was used, which Paper2Assets bundle checksum the artifact was rendered from, which poster theme and header arrangement or which TTS voice was picked, and which model backed the run), so readers can audit the generation chain and, where required, disclose AI involvement. Paper2Assets writes an inventory manifest with the source PDF's checksum and the per-step settings (\S\ref{sec:p2a}), and the Paper2Video QA record logs the deck, timeline, and narration configuration (\S\ref{sec:p2v}); together these form the attributable ancestry of any downstream artifact.

\textbf{Licensing and redistribution.} The \texttt{edge-tts} client accesses Microsoft Edge's online text-to-speech service \citep{edgetts}; deployments should verify the service terms applicable to their use. Per-figure copyright on extracted paper figures follows the source paper's license, which is typically arXiv-friendly but should be confirmed before redistribution, and institution and venue logos fetched by Paper2Assets follow their respective source licenses (Wikimedia Commons and Wikidata where available). ResearchStudio-Reel itself does not redistribute paper figures or logos: the user runs the pipeline on their own paper, and the generated artifacts inherit the source paper's license.

\section{Reproducibility}\label{app:repro}

\textbf{Code and dependencies.} ResearchStudio-Reel is released open-source under the MIT license at \url{https://aka.ms/ResearchStudio}, with each skill's \texttt{SKILL.md} as the source of truth for its workflow. The five skills pin their Python dependencies in per-skill \texttt{requirements.txt} files, and the top-level \texttt{install.sh} symlinks each skill into the host's skills directory, so \texttt{git pull} updates them all in place. System-level dependencies (\texttt{poppler-utils}, \texttt{libreoffice}, \texttt{ffmpeg}, and a headless Chromium) are listed in the README.

\textbf{Models and credentials.} Narration uses Edge TTS \citep{edgetts}, which needs no credentials; language-model access is picked up from the host runtime (\texttt{ANTHROPIC\_API\_KEY} for Claude Code, standard Codex configuration for Codex). Our own runs routed those credentials through a shared Copilot API proxy rather than the vendors' first-party endpoints, with jobs spread across several hosts; the model identifiers used throughout the paper (\texttt{claude-opus-4.8}, \texttt{gpt-5.5}, \texttt{gemini-3.1-pro}) name the underlying models as served by that proxy, and pointing the same skills at a first-party endpoint is a configuration change rather than a code change.

\textbf{Benchmark protocol.} Poster scoring in \S\ref{sec:experiments} reuses the Paper2Poster benchmark \citep{pang2026paper2poster} verbatim: the same 100-paper list, the same six-criterion aesthetic and information rubric, and the same PaperQuiz probe. Every poster is downscaled to at most $2560$\,px on the long edge and scored by both \texttt{claude-opus-4.8} and \texttt{gpt-5.5}; each Table~\ref{tbl:main} cell reports the mean of the two judges, with the per-judge breakdown in Appendix~\ref{app:judges} (Table~\ref{tbl:judge-perjudge}). Author ground-truth posters and the reproduced baselines (Paper2Poster tool, PosterGen, P2P) all go through the same downscale-and-score pipeline, and the three single-shot LLM baselines share one fixed prompt reproduced verbatim in Appendix~\ref{app:llm-prompt}, which controls for prompt differences across these three single-shot baselines. We do not report graded video or blog scores in this version; video and blog capability coverage is reported instead in Tables~\ref{tbl:p2v-capabilities} and~\ref{tbl:p2b-capabilities}, and applying external video or suite benchmarks is left to future work.

\section{Per-judge Benchmark Scores}\label{app:judges}

Table~\ref{tbl:main} in \S\ref{sec:experiments} reports each poster-quality cell as the mean of two VLM judges, one Claude and one GPT vision-language model. We average two judges from different model families to reduce dependence on any single evaluator. Table~\ref{tbl:judge-perjudge} unfolds that mean side by side: it gives, for each judge, the same six aesthetic and information sub-criteria together with the two PaperQuiz comprehension splits, so the unaveraged view of either judge is available for inspection. The judges yield similar aggregate rankings but differ on individual criteria; we therefore report both sets of scores. Their mean is used as a compact summary and should not be interpreted as eliminating evaluator bias.

\begin{table}[H]
\centering
\caption{\textbf{Per-criterion scores under each judge, side by side.} The two VLM judges whose mean forms Table~\ref{tbl:main}: Claude (\texttt{claude-opus-4.8}, left block) and GPT (\texttt{gpt-5.5}, right block). Aesthetic and Information sub-criteria on 1--5 (higher better); Quiz is $^{\S}$PaperQuiz answer accuracy (\%). Sub-criteria abbreviated El(ement), En(gagement), La(yout), Lo(w-level), L(o)g(ic), Co(ntent), Ov(erall), Det(ail), Und(erstanding); each Ov.\ column is the mean of the three sub-criteria in its block. Best value per column among the systems in \textbf{bold} and second best \underline{underlined}. $^{\ast}$Baseline systems are reproduced with best efforts; the Codex row is our skill under Codex + \texttt{gpt-5.5}.}
\label{tbl:judge-perjudge}
\vspace{0.5em}
\setlength{\tabcolsep}{3pt}
\scriptsize
\resizebox{\linewidth}{!}{%
\begin{tabular}{@{}l ccc c ccc c cc @{\hskip 5pt}|@{\hskip 5pt} ccc c ccc c cc@{}}
\toprule
& \multicolumn{10}{c}{\textbf{Claude judge} (\texttt{claude-opus-4.8})} & \multicolumn{10}{c}{\textbf{GPT judge} (\texttt{gpt-5.5})} \\
\cmidrule(lr){2-11}\cmidrule(lr){12-21}
& \multicolumn{4}{c}{Aesthetic} & \multicolumn{4}{c}{Information} & \multicolumn{2}{c}{Quiz\,(\%)} & \multicolumn{4}{c}{Aesthetic} & \multicolumn{4}{c}{Information} & \multicolumn{2}{c}{Quiz\,(\%)} \\
\cmidrule(lr){2-5}\cmidrule(lr){6-9}\cmidrule(lr){10-11}\cmidrule(lr){12-15}\cmidrule(lr){16-19}\cmidrule(lr){20-21}
System & El. & En. & La. & Ov. & Lo. & Lg. & Co. & Ov. & Det.$^{\S}$ & Und.$^{\S}$ & El. & En. & La. & Ov. & Lo. & Lg. & Co. & Ov. & Det.$^{\S}$ & Und.$^{\S}$ \\
\midrule
\rowcolor{black!10}\multicolumn{21}{@{}l}{\emph{Single-shot}}\\
Claude-4.8 Opus~\citep{anthropic2026opus48} & 3.16 & 2.71 & 2.94 & 2.94 & 3.98 & \underline{4.02} & \underline{3.93} & \underline{3.98} & 73.20 & 97.58 & 3.04 & 2.65 & 2.99 & 2.89 & 3.85 & \underline{3.99} & 3.67 & \underline{3.84} & 66.52 & 89.76 \\
GPT-5.5~\citep{openai2026gpt55} & \underline{3.17} & 2.96 & 3.49 & 3.21 & \underline{4.00} & 4.00 & \textbf{3.98} & \textbf{3.99} & \underline{73.24} & \textbf{98.04} & 3.07 & 2.92 & 3.31 & 3.10 & \textbf{3.98} & \textbf{4.00} & \textbf{3.83} & \textbf{3.94} & \underline{68.54} & 90.46 \\
Gemini-3.1 Pro~\citep{google2026gemini31pro} & 3.10 & 2.88 & 3.20 & 3.06 & \underline{4.00} & 4.00 & 3.91 & 3.97 & 67.98 & 97.42 & 3.03 & 2.89 & 3.34 & 3.09 & 3.90 & \underline{3.99} & 3.39 & 3.76 & 62.74 & 89.76 \\
\rowcolor{black!10}\multicolumn{21}{@{}l}{\emph{Poster Pipeline}}\\
Paper2Poster tool$^{\ast}$ \citep{pang2026paper2poster} & 2.90 & 2.49 & 3.00 & 2.80 & 3.63 & 3.53 & 3.64 & 3.60 & 72.71 & \textbf{98.04} & 2.83 & 2.11 & 3.00 & 2.65 & 3.30 & 3.30 & 3.03 & 3.21 & 67.18 & \underline{92.00} \\
P2P$^{\ast}$ \citep{sun2025p2p} & 2.73 & 2.47 & 3.23 & 2.81 & 3.97 & 3.83 & 3.70 & 3.83 & \textbf{77.73} & \underline{97.93} & 2.63 & 2.20 & 3.07 & 2.63 & 3.87 & 3.73 & \underline{3.70} & 3.77 & \textbf{73.13} & \textbf{93.27} \\
PosterGen$^{\ast}$ \citep{zhang2025postergen} & 2.93 & 2.26 & 2.96 & 2.72 & 3.96 & 3.85 & 3.63 & 3.81 & 62.15 & 97.11 & 2.93 & 2.19 & 3.00 & 2.71 & 3.63 & 3.67 & 3.11 & 3.47 & 52.81 & 86.59 \\
\textbf{\reelword{} (Claude Code)} & \textbf{3.53} & \textbf{3.32} & \textbf{3.99} & \textbf{3.61} & \textbf{4.01} & \textbf{4.05} & 3.56 & 3.87 & 60.82 & 96.90 & \textbf{3.53} & \textbf{3.01} & \textbf{3.99} & \textbf{3.51} & \underline{3.97} & \textbf{4.00} & 3.25 & 3.74 & 52.76 & 87.42 \\
\textbf{\reelword{} (Codex)} & 3.14 & \underline{3.13} & \underline{3.59} & \underline{3.29} & 3.98 & 3.98 & 3.19 & 3.72 & 60.30 & 97.20 & \underline{3.18} & \underline{2.93} & \underline{3.64} & \underline{3.25} & 3.91 & 3.95 & 2.99 & 3.62 & 50.80 & 86.82 \\
\midrule
Author ground-truth & 3.26 & 2.63 & 3.47 & 3.12 & 3.70 & 3.90 & 3.73 & 3.78 & 55.82 & 96.98 & 2.99 & 2.57 & 3.22 & 2.93 & 3.21 & 3.80 & 3.27 & 3.43 & 53.64 & 85.82 \\
\bottomrule
\end{tabular}}
\end{table}

\section{Poster Comparison Across Systems}\label{app:compare}
\begin{figure}[H]
\centering
\begin{minipage}[b]{0.35\linewidth}\centering
\includegraphics[width=\linewidth]{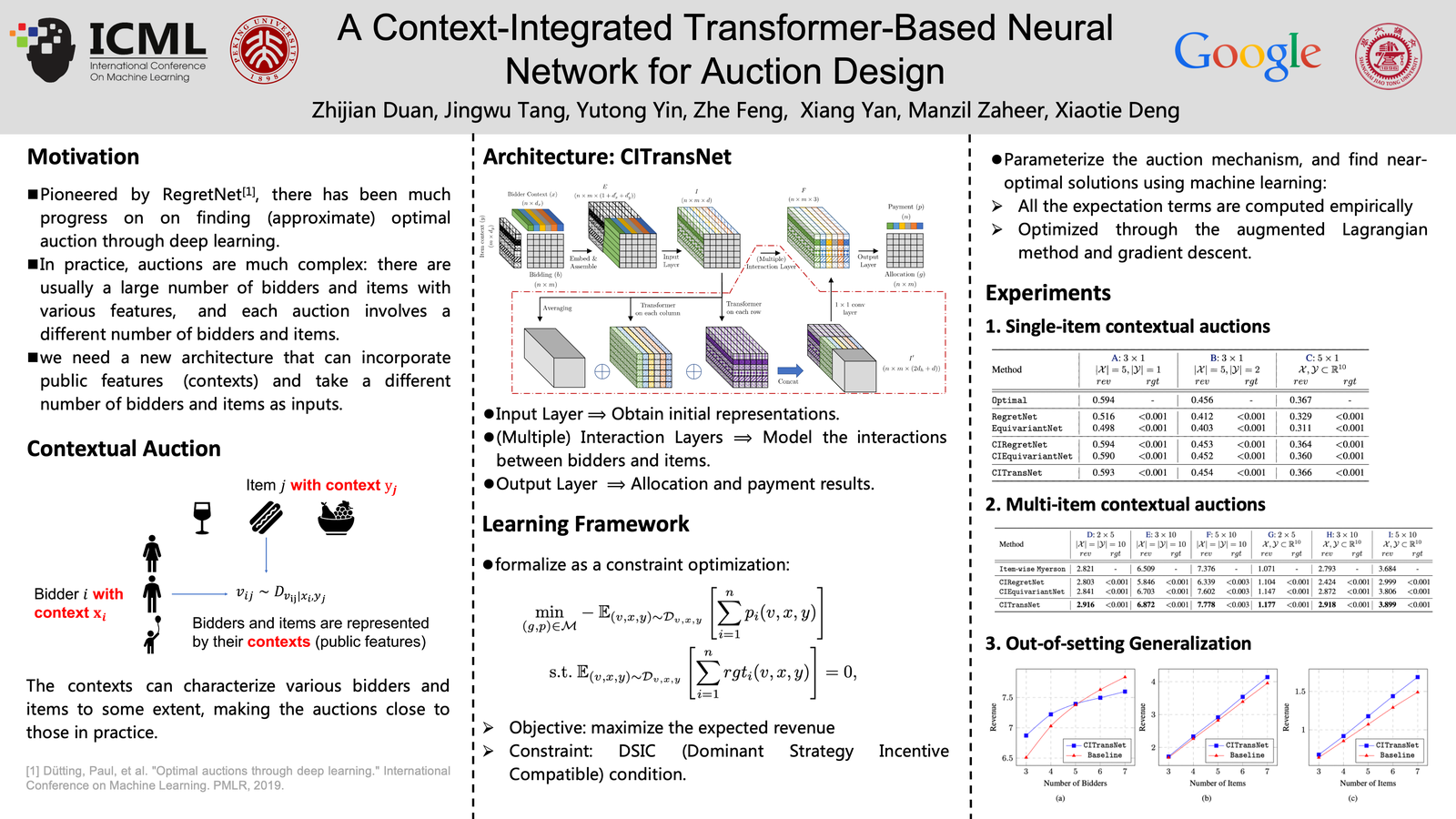}\\[1pt]{\scriptsize Human-made (from poster authors)}\\[5pt]
\includegraphics[width=\linewidth]{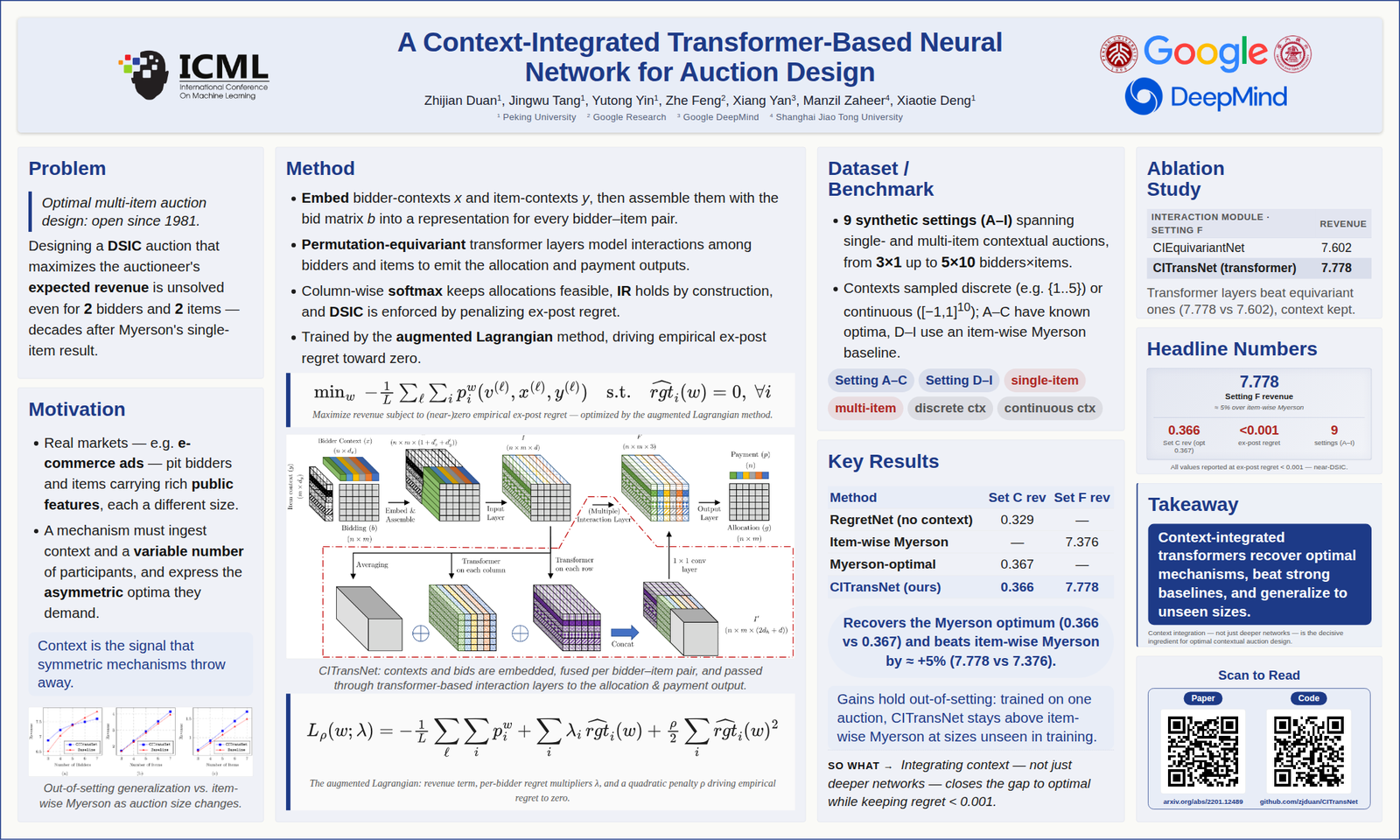}\\[1pt]{\scriptsize \textbf{ResearchStudio-\reelword ~(Ours)}}
\end{minipage}\hfill
\begin{minipage}[b]{0.28\linewidth}\centering
\includegraphics[width=\linewidth]{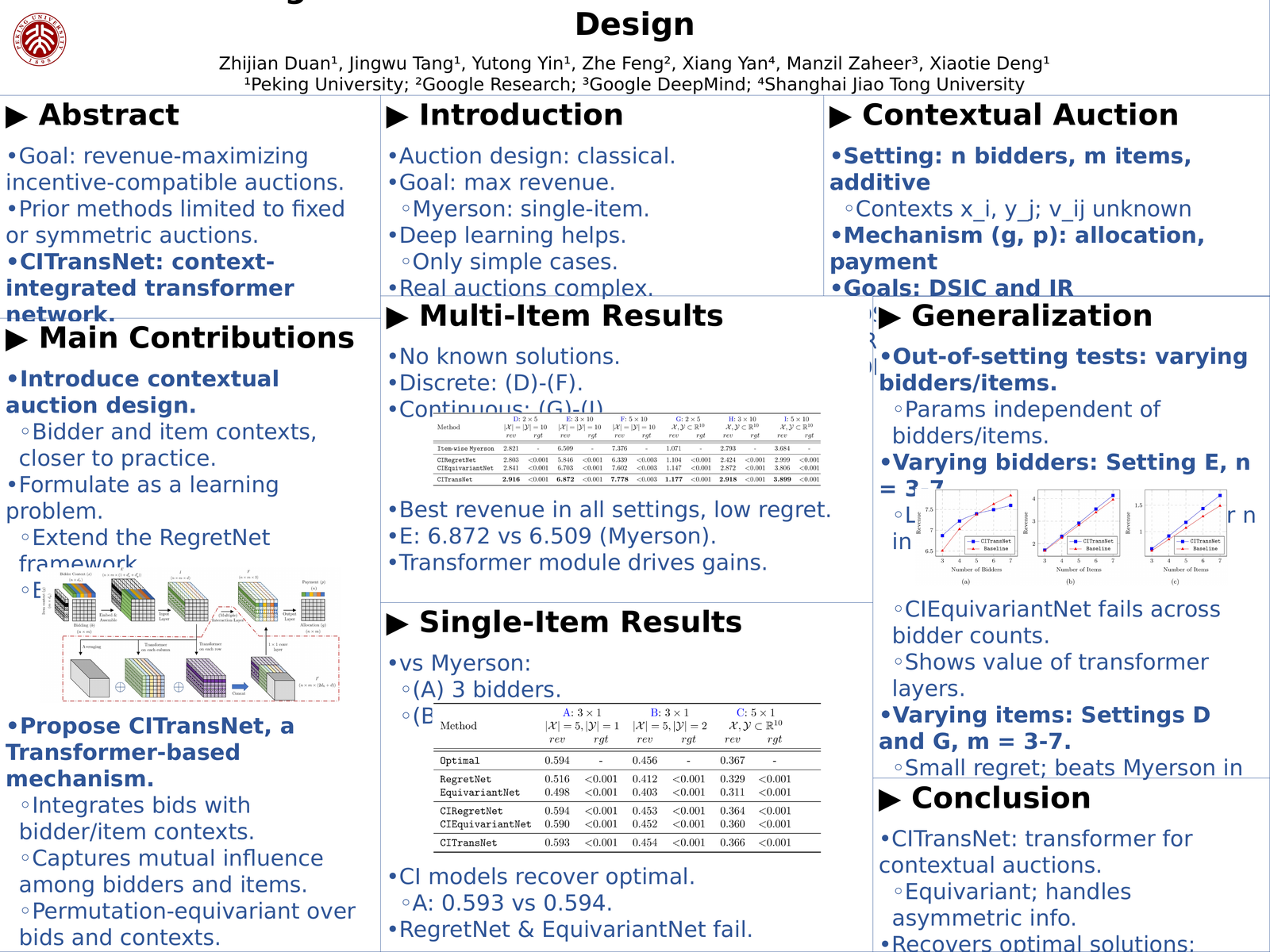}\\[1pt]{\scriptsize Paper2Poster Tool~\citep{pang2026paper2poster}}\\[5pt]
\includegraphics[width=\linewidth]{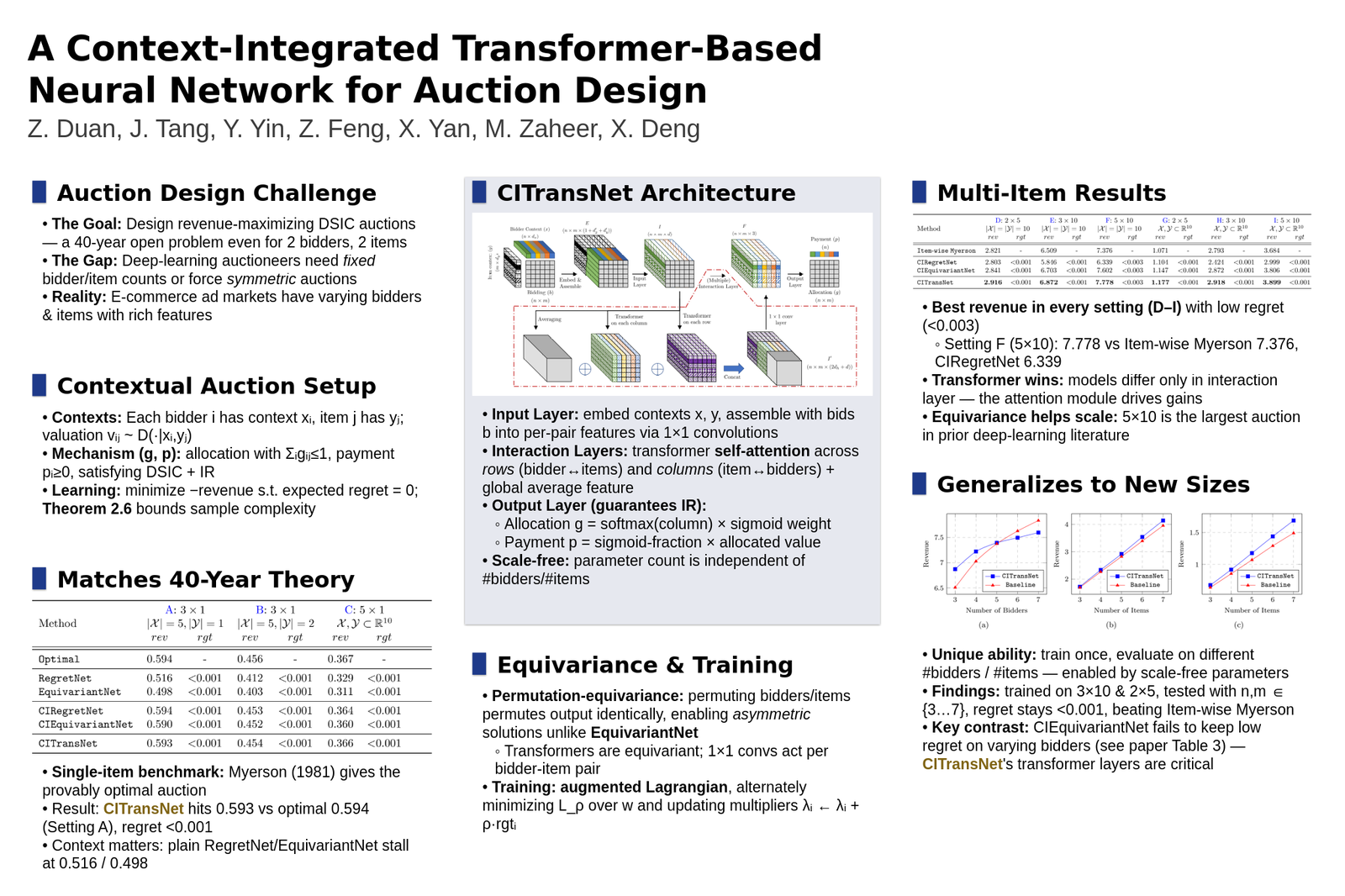}\\[1pt]{\scriptsize PosterGen~\citep{zhang2025postergen}}
\end{minipage}\hfill
\begin{minipage}[b]{0.33\linewidth}\centering
\includegraphics[width=\linewidth]{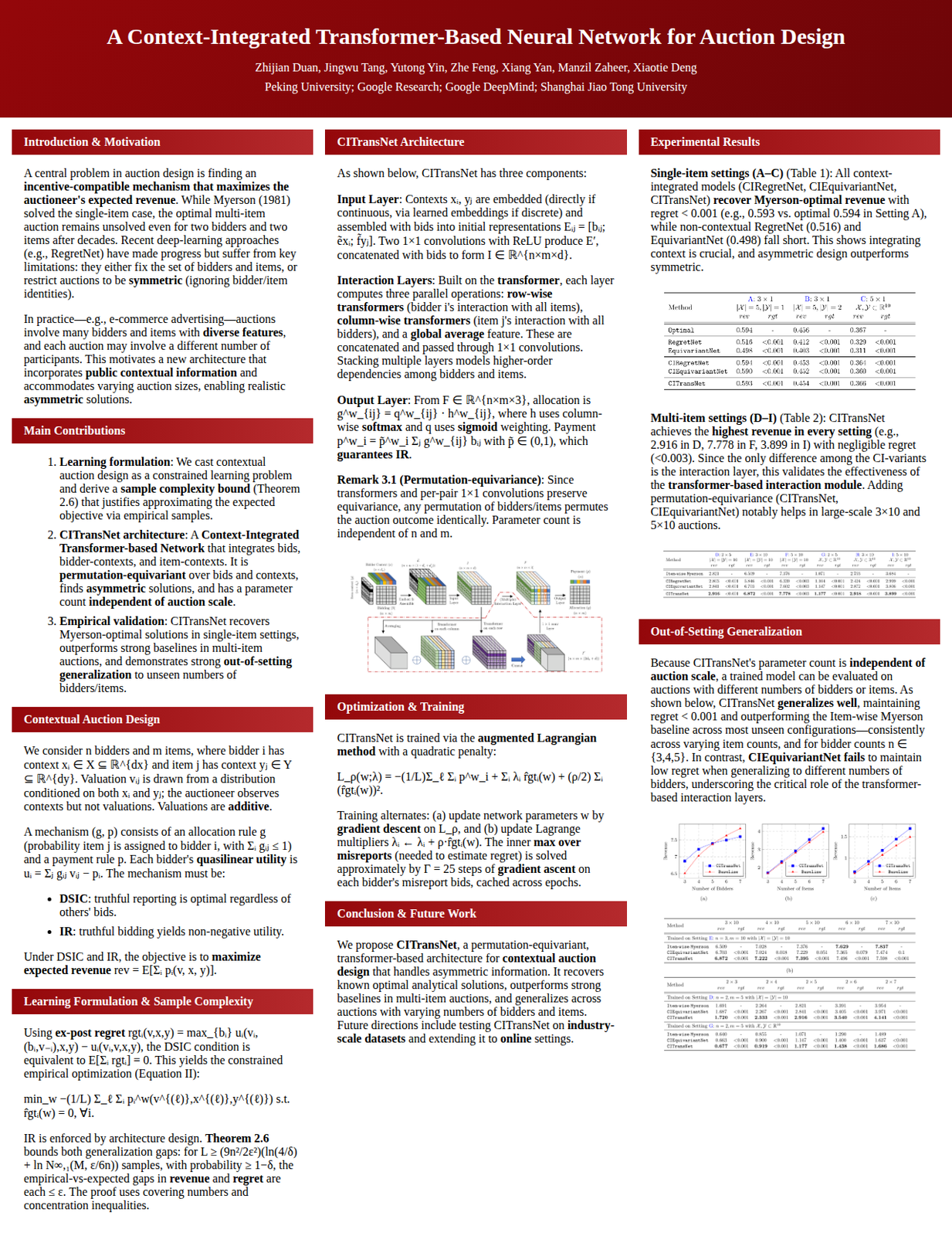}\\[1pt]{\scriptsize P2P~\citep{sun2025p2p}}
\end{minipage}

\vspace{6pt}
\begin{minipage}[b]{0.32\linewidth}\centering
\includegraphics[width=\linewidth]{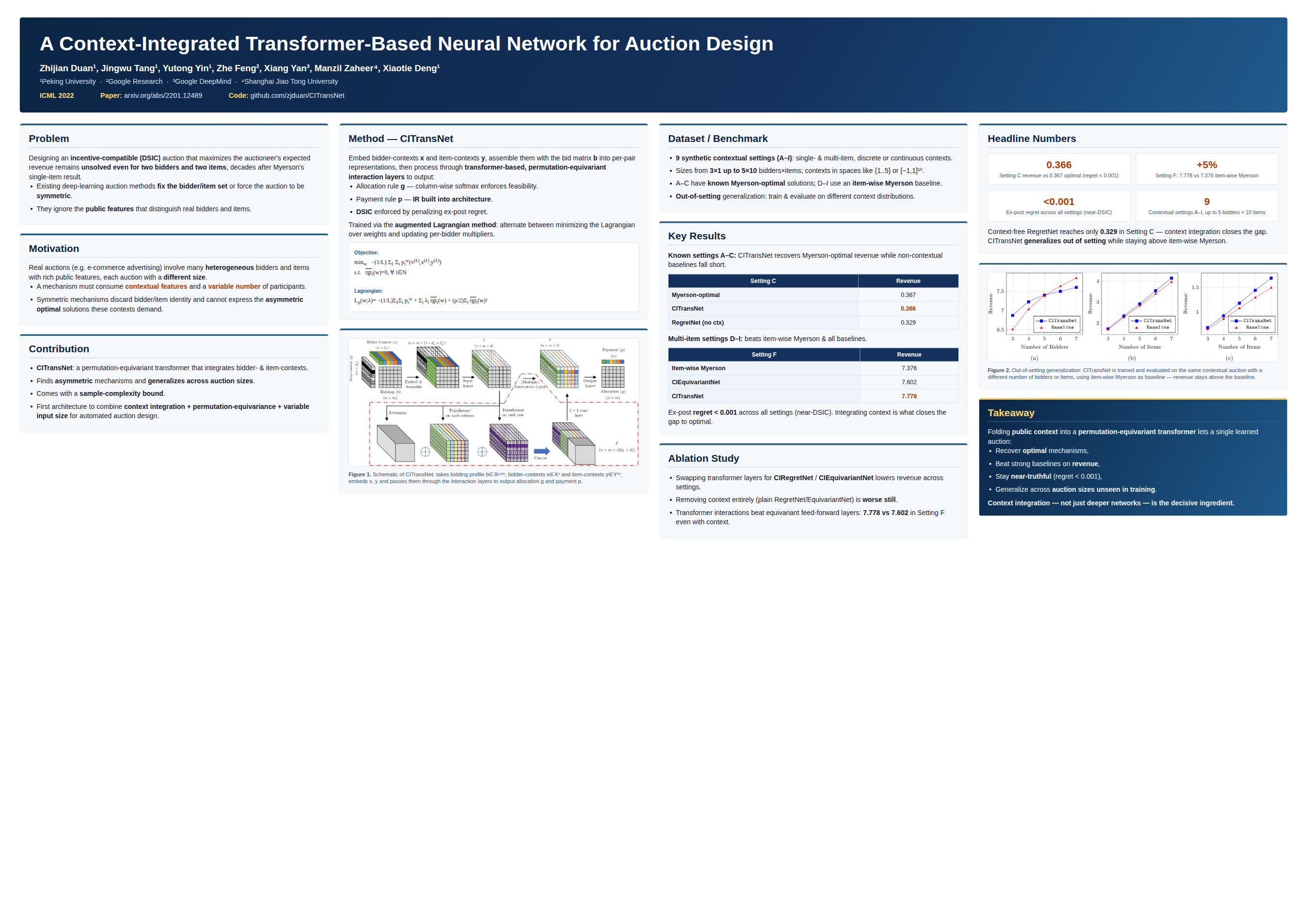}\\[1pt]{\scriptsize Claude-4.8 Opus (single-shot)}
\end{minipage}\hfill
\begin{minipage}[b]{0.32\linewidth}\centering
\includegraphics[width=\linewidth]{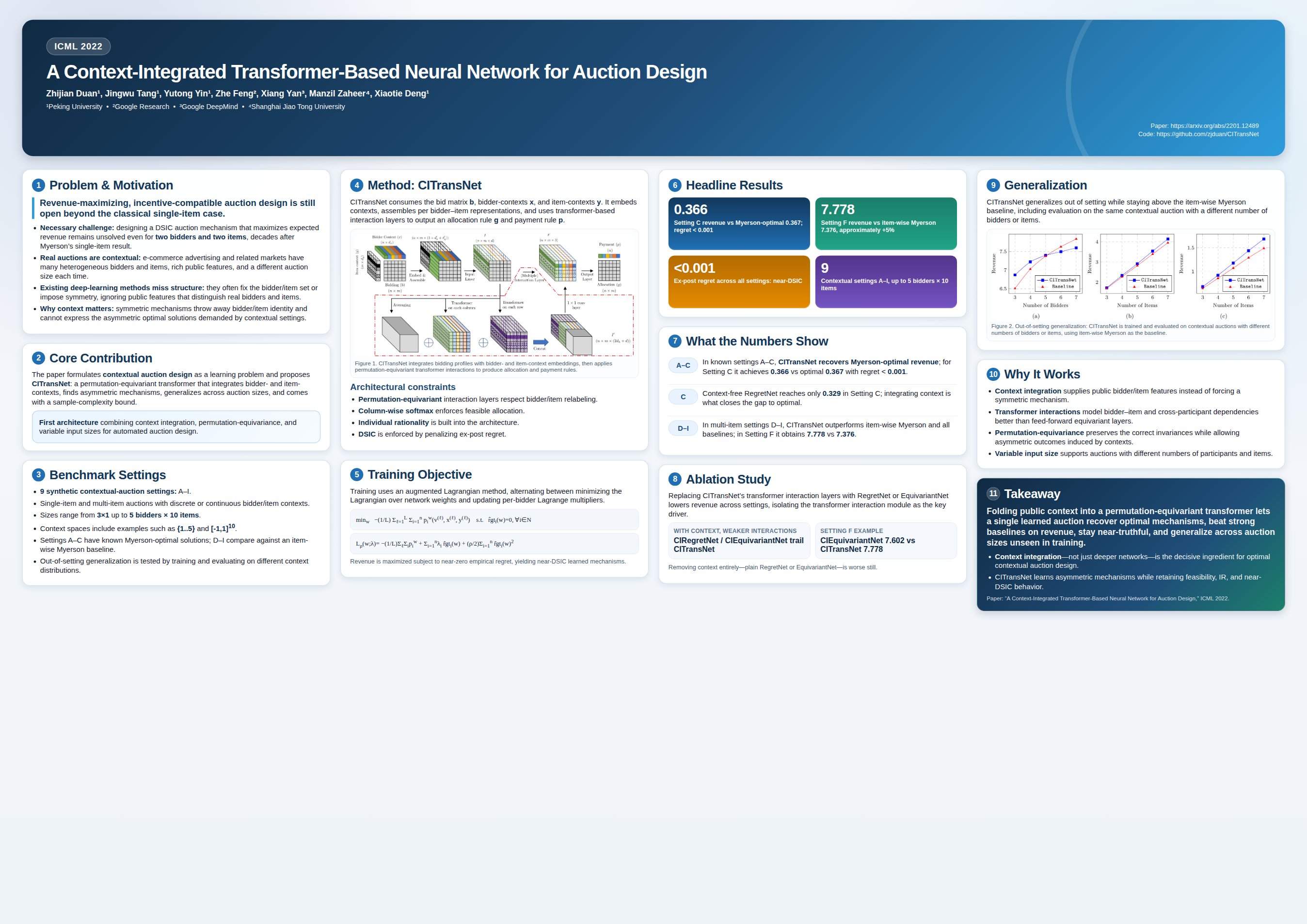}\\[1pt]{\scriptsize GPT-5.5 (single-shot)}
\end{minipage}\hfill
\begin{minipage}[b]{0.32\linewidth}\centering
\includegraphics[width=\linewidth]{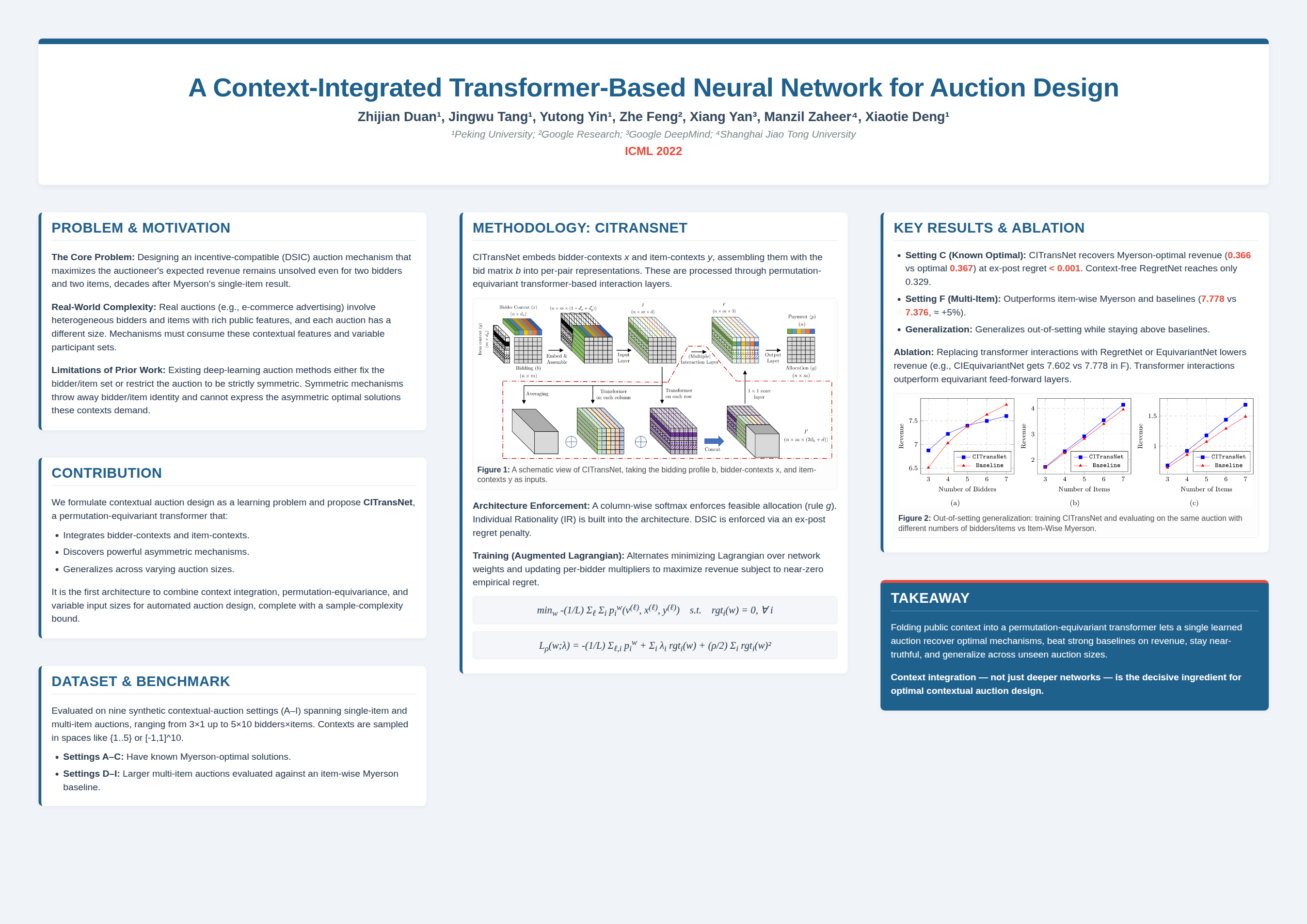}\\[1pt]{\scriptsize Gemini-3.1 Pro (single-shot)}
\end{minipage}
\caption{\textbf{Poster comparison on one benchmark paper} (``A Context-Integrated Transformer-Based Neural Network for Auction Design''). Every system renders from the same source PDF: our poster and the human-made (authors') poster (left, largest), the Paper2Poster Tool~\citep{pang2026paper2poster} and PosterGen~\citep{zhang2025postergen} baselines (center), the portrait P2P~\citep{sun2025p2p} baseline (right), and three single-shot LLM baselines (Claude-4.8 Opus, GPT-5.5, Gemini-3.1 Pro; bottom row). The single-shot posters carry comparable information but a markedly weaker visual layout than ours.}
\label{fig:poster-compare}
\end{figure}

On a single benchmark paper we place our poster beside the three baseline systems and the authors' ground-truth poster, all rendered from the same source PDF (Figure~\ref{fig:poster-compare}); the panel widths are uneven only so the columns bottom-align.

\textbf{Observation.} The single-shot LLM baselines are not held back by content: Claude-4.8 Opus, GPT-5.5, and Gemini-3.1 Pro each recover the paper's title, contributions, and headline numbers and pick sensible figures, so their PaperQuiz accuracy in Table~\ref{tbl:main} is close to ours. The gap is almost entirely in layout: emitting a full A0 poster in one pass, they fall back on a rigid grid with unequal columns, long undifferentiated text, and figures placed without regard to whitespace, so the sheet reads flat at poster scale. Our compose-then-fill loop instead balances column density and promotes key results into typed widgets, which is where the Layout and aesthetic margins in Table~\ref{tbl:main} come from; the reproduced poster pipelines (Paper2Poster Tool, PosterGen, and the portrait P2P) sit in between.

\section{Poster Gallery Showcase}\label{app:gallery}
Posters generated end-to-end by Paper2Poster from the paper PDF alone, for the 100-paper benchmark~\cite{pang2026paper2poster}; per-paper titles and the operational cost analysis appear in \S\ref{sec:experiments}. We show 10 posters sampled uniformly at random from the 100-paper set, two per row. Each poster is rendered in a distinct accent color to illustrate the template's theming range and to keep the thumbnails visually separable; the column layout, visual style, and header are sampled per paper by the composer.

\begingroup\setlength{\fboxsep}{0pt}\centering
\begin{minipage}[t]{0.47\linewidth}\centering\includegraphics[width=\linewidth]{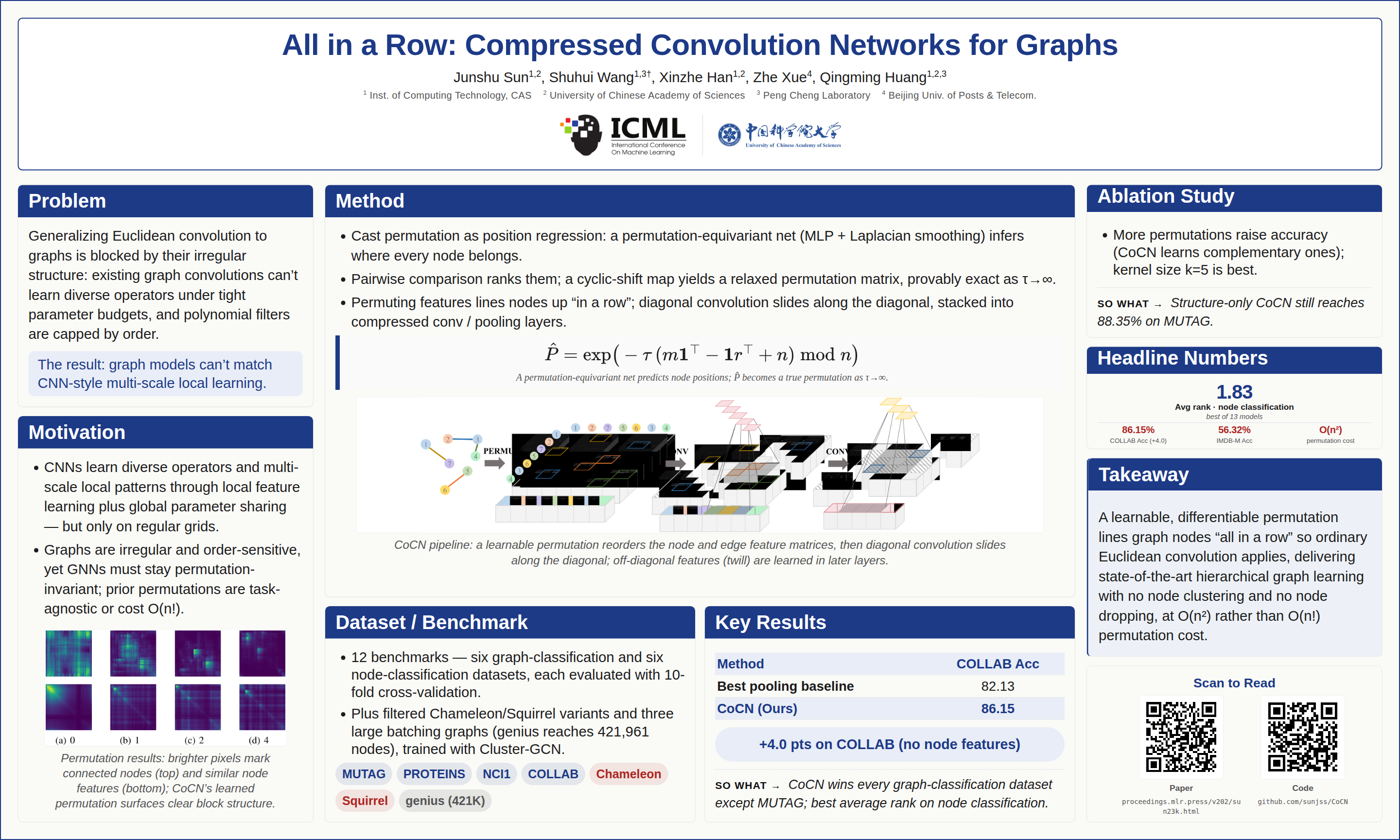}\end{minipage}\hfill\begin{minipage}[t]{0.47\linewidth}\centering\includegraphics[width=\linewidth]{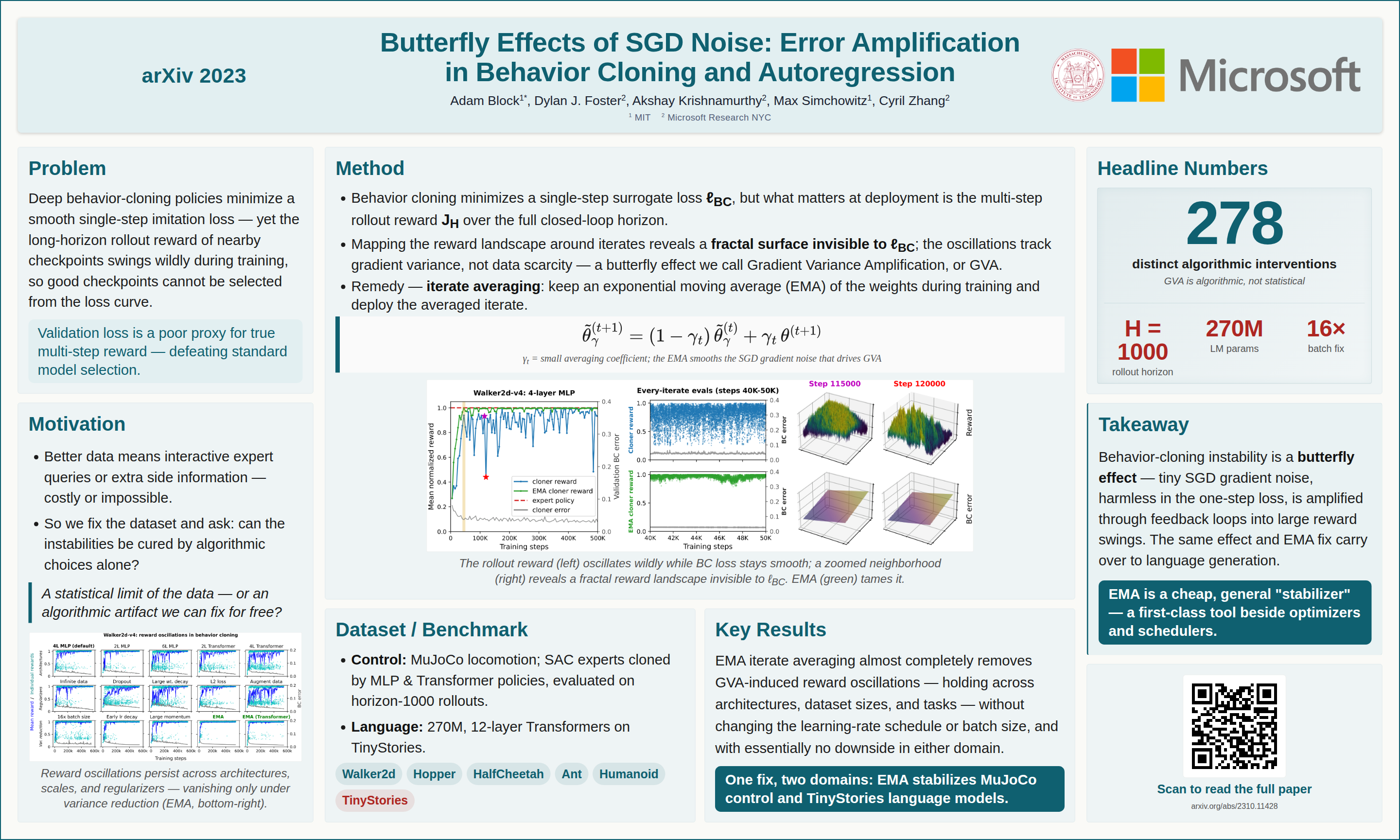}\end{minipage}\par\vspace{5pt}
\begin{minipage}[t]{0.47\linewidth}\centering\includegraphics[width=\linewidth]{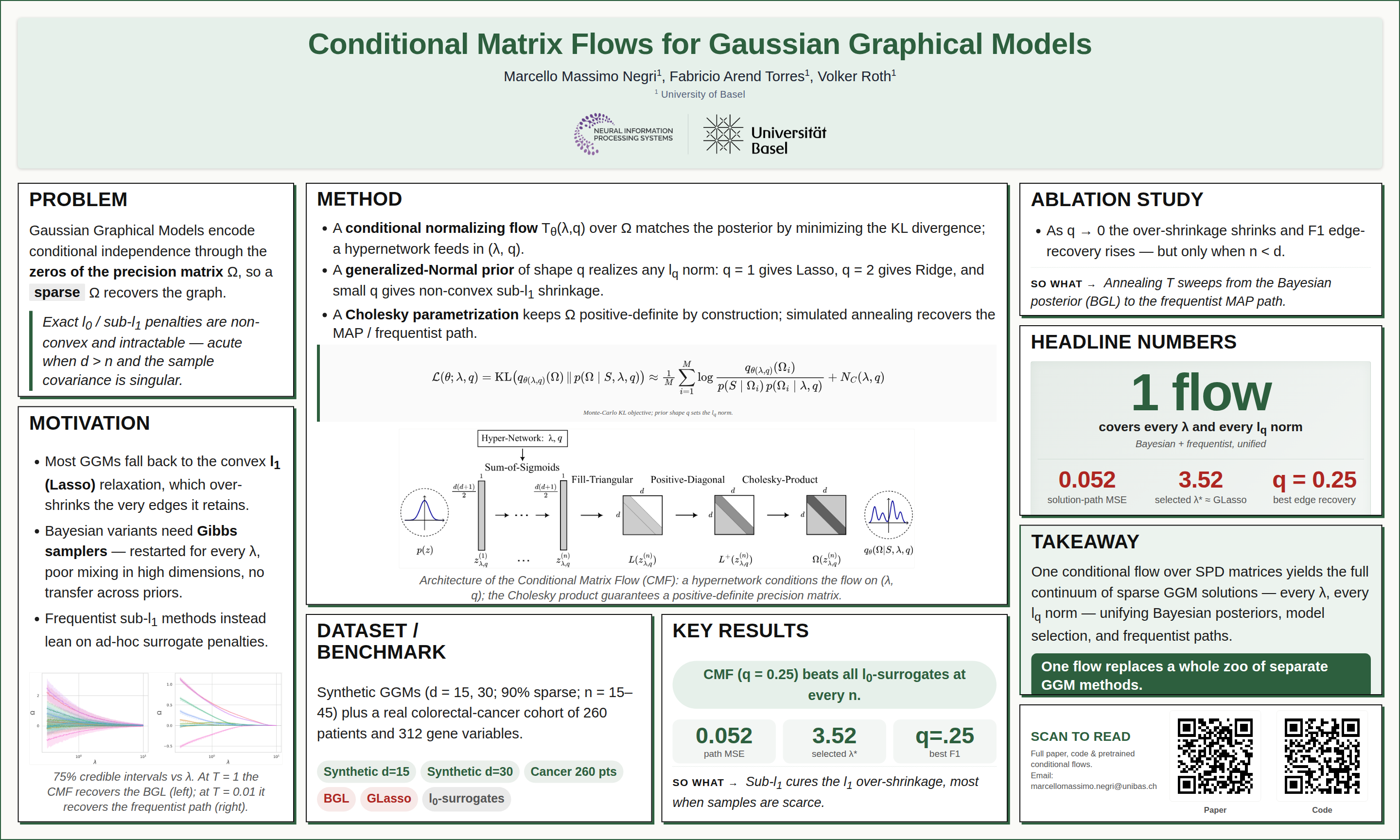}\end{minipage}\hfill\begin{minipage}[t]{0.47\linewidth}\centering\includegraphics[width=\linewidth]{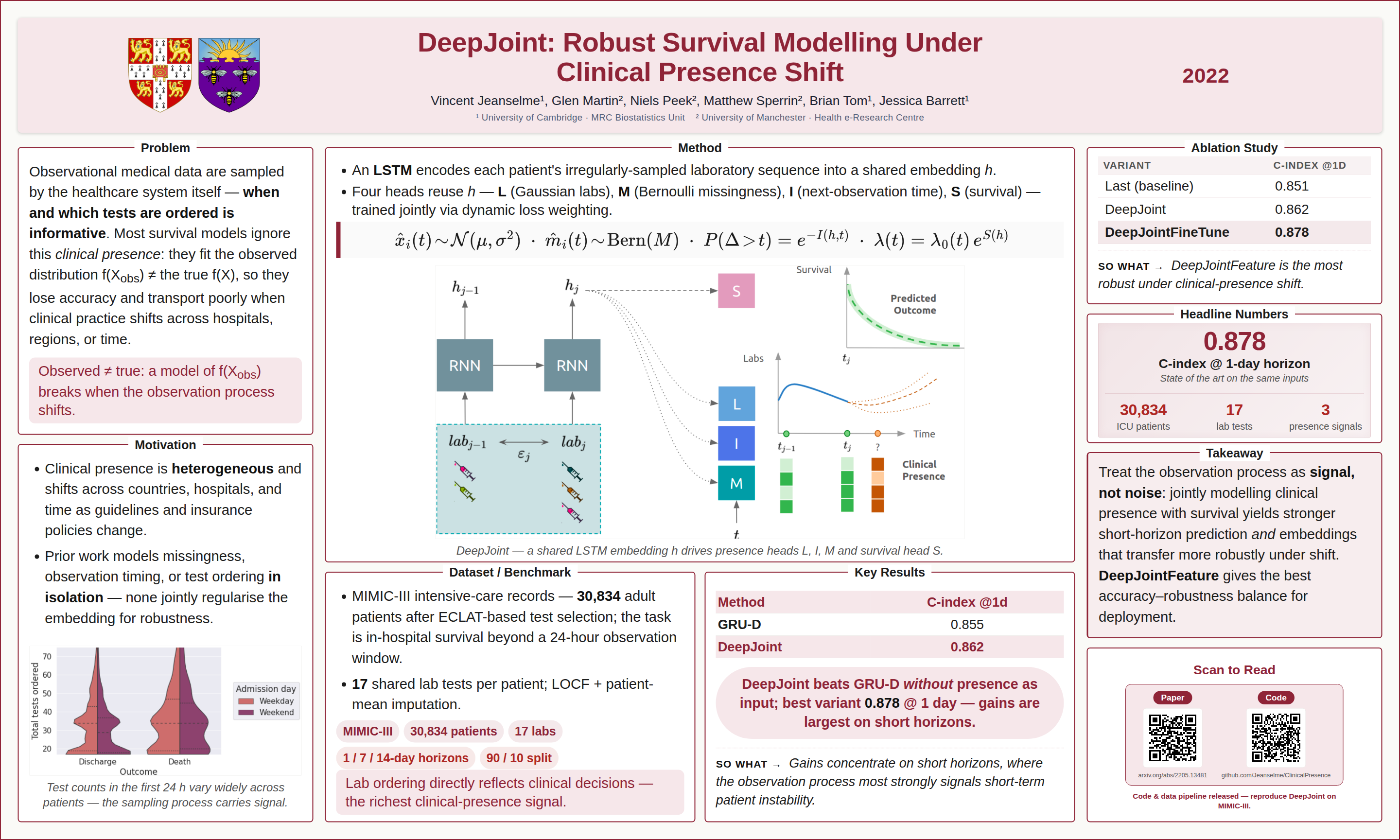}\end{minipage}\par\vspace{5pt}
\begin{minipage}[t]{0.47\linewidth}\centering\includegraphics[width=\linewidth]{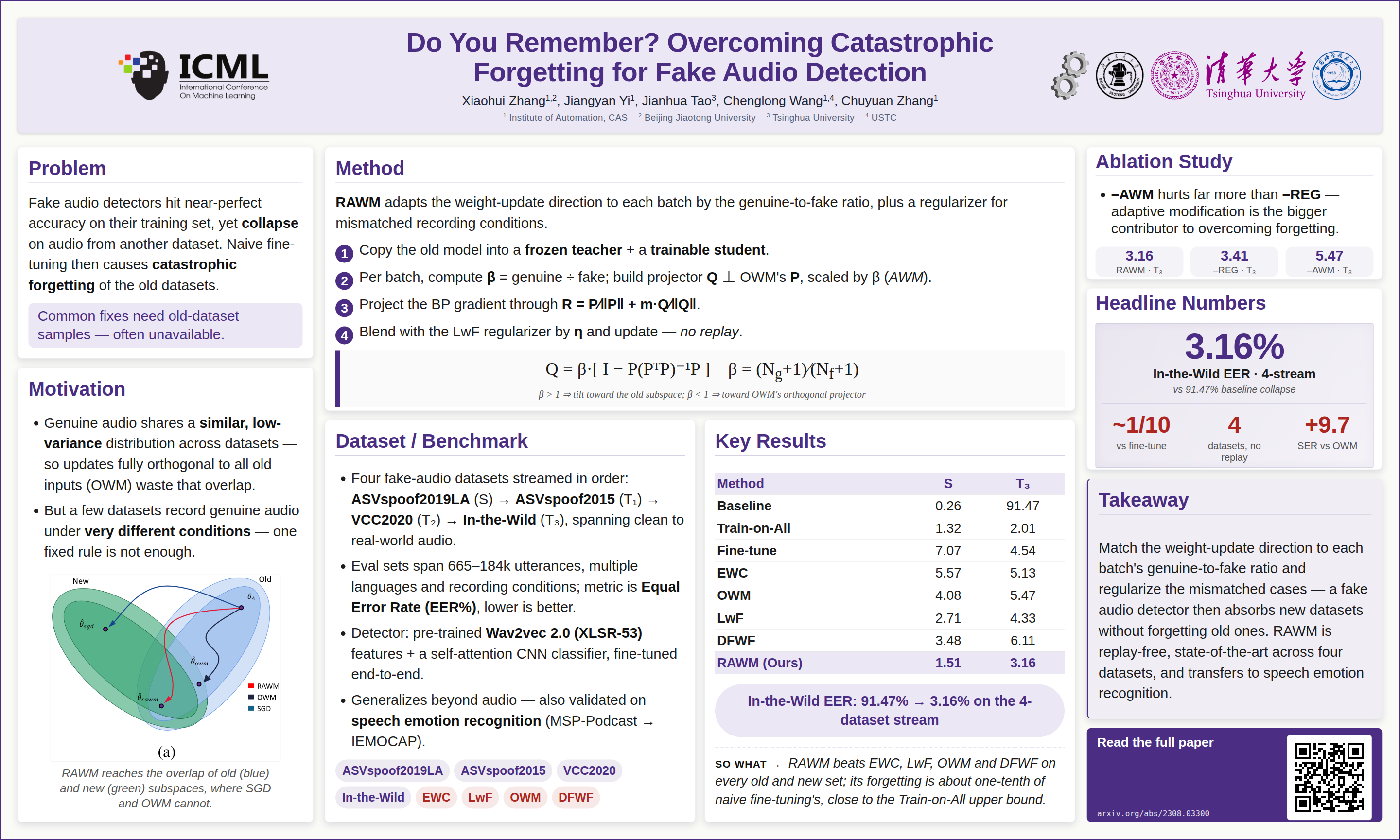}\end{minipage}\hfill\begin{minipage}[t]{0.47\linewidth}\centering\includegraphics[width=\linewidth]{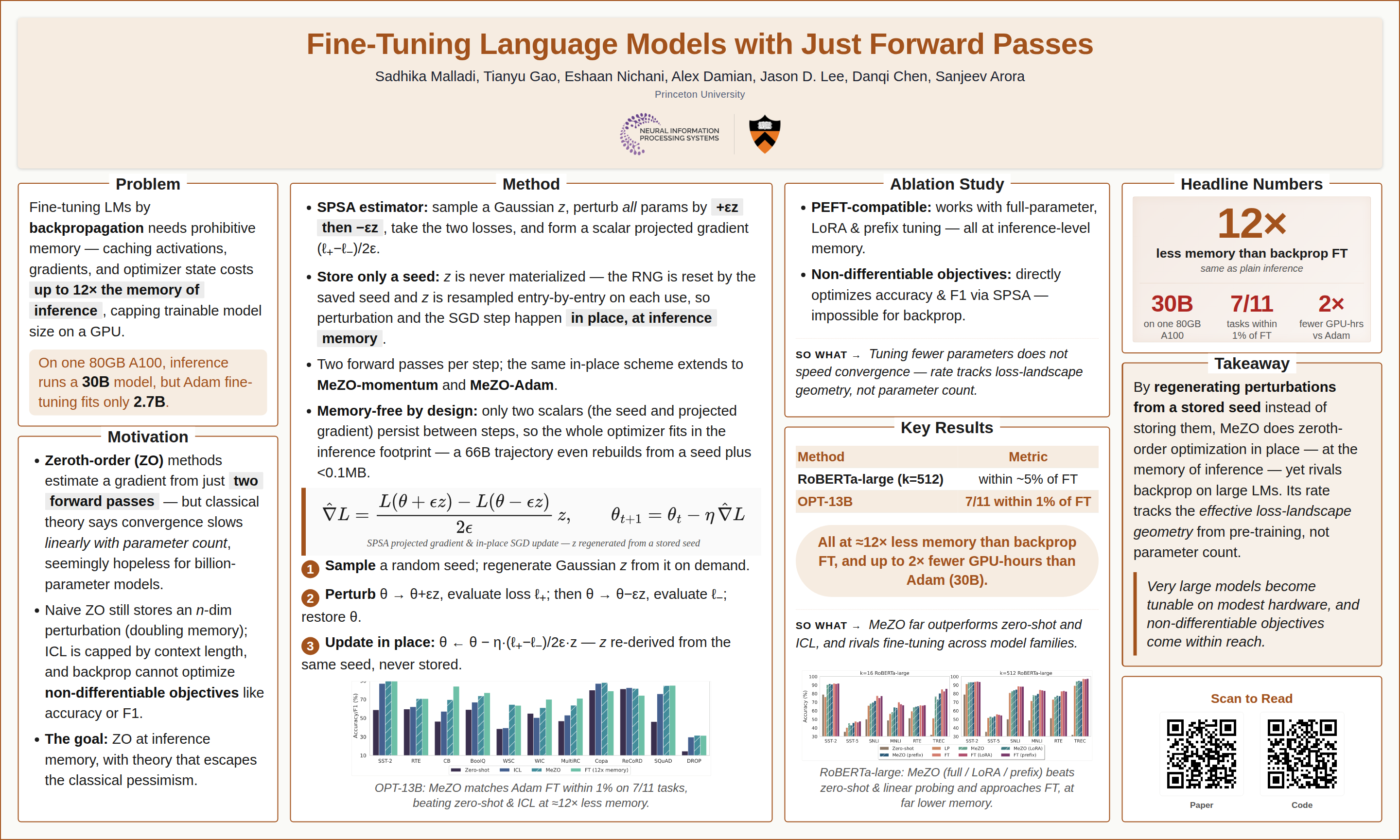}\end{minipage}\par\vspace{5pt}
\begin{minipage}[t]{0.47\linewidth}\centering\includegraphics[width=\linewidth]{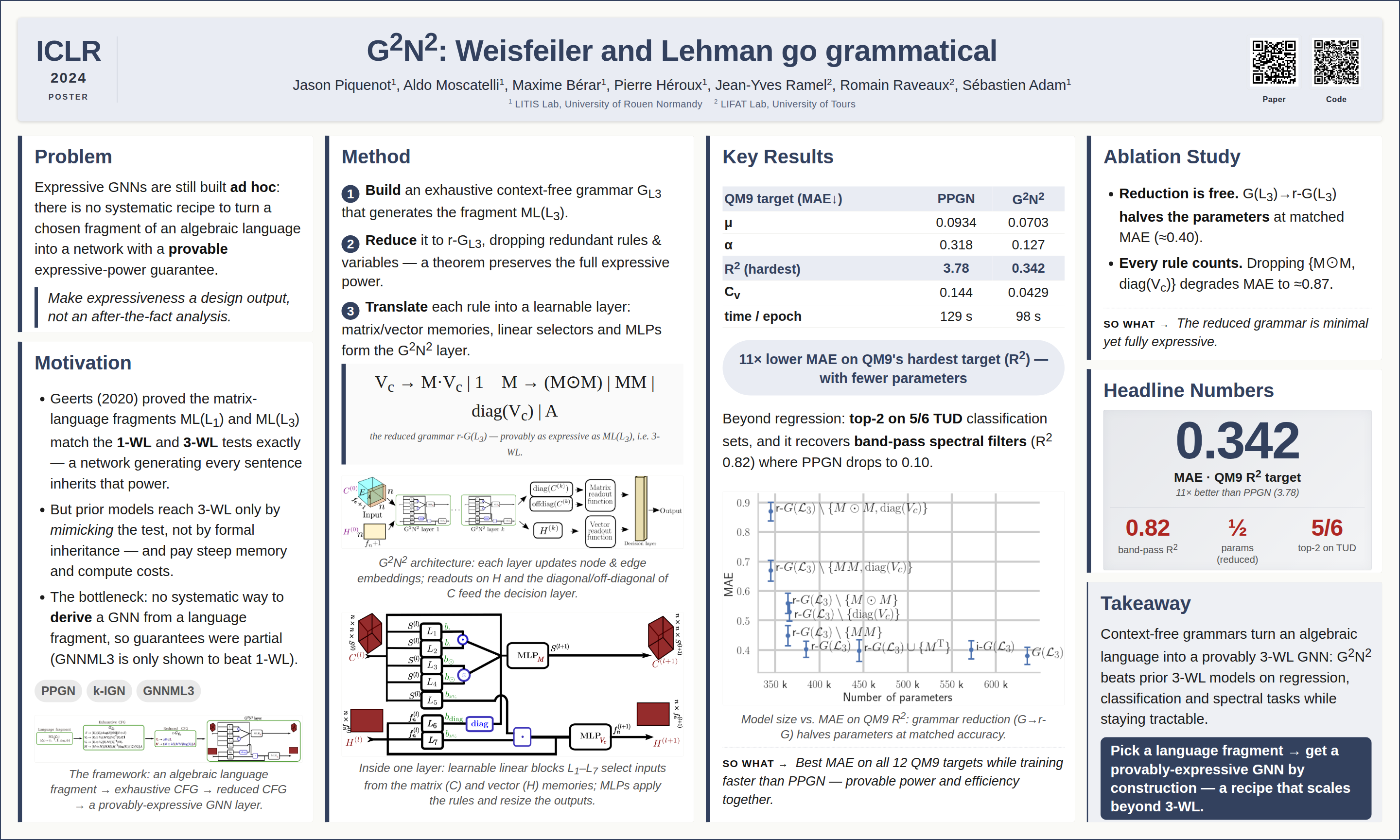}\end{minipage}\hfill\begin{minipage}[t]{0.47\linewidth}\centering\includegraphics[width=\linewidth]{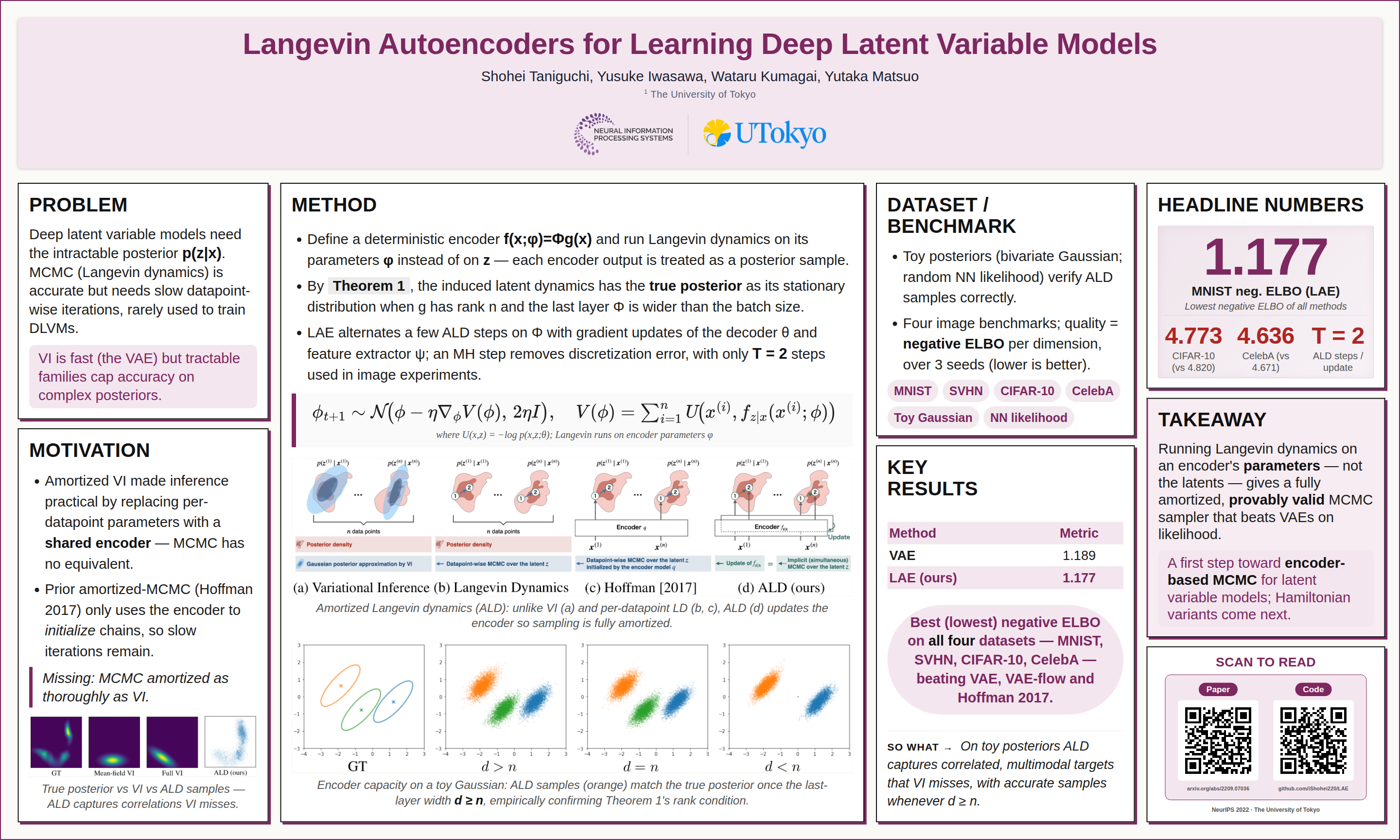}\end{minipage}\par\vspace{5pt}
\begin{minipage}[t]{0.47\linewidth}\centering\includegraphics[width=\linewidth]{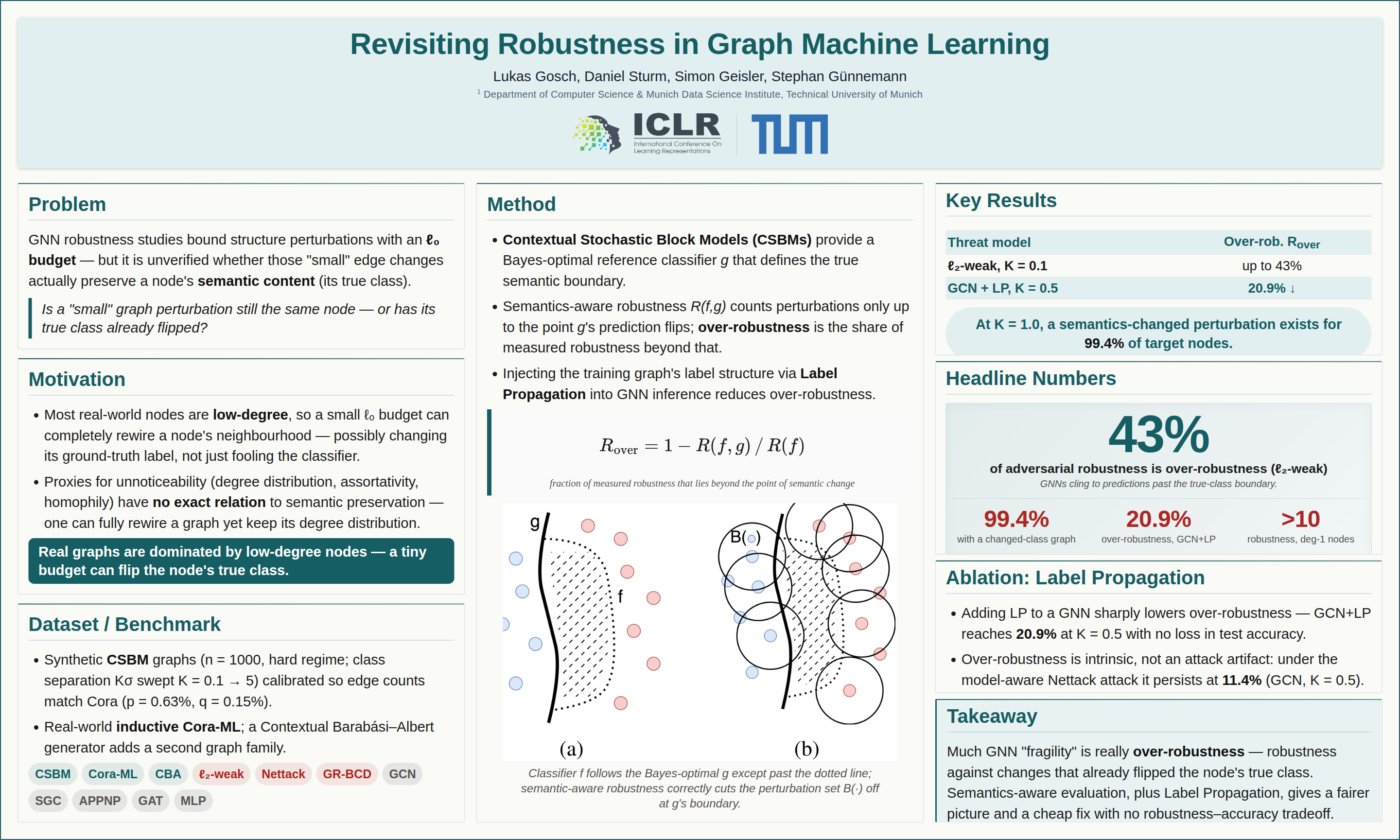}\end{minipage}\hfill\begin{minipage}[t]{0.47\linewidth}\centering\includegraphics[width=\linewidth]{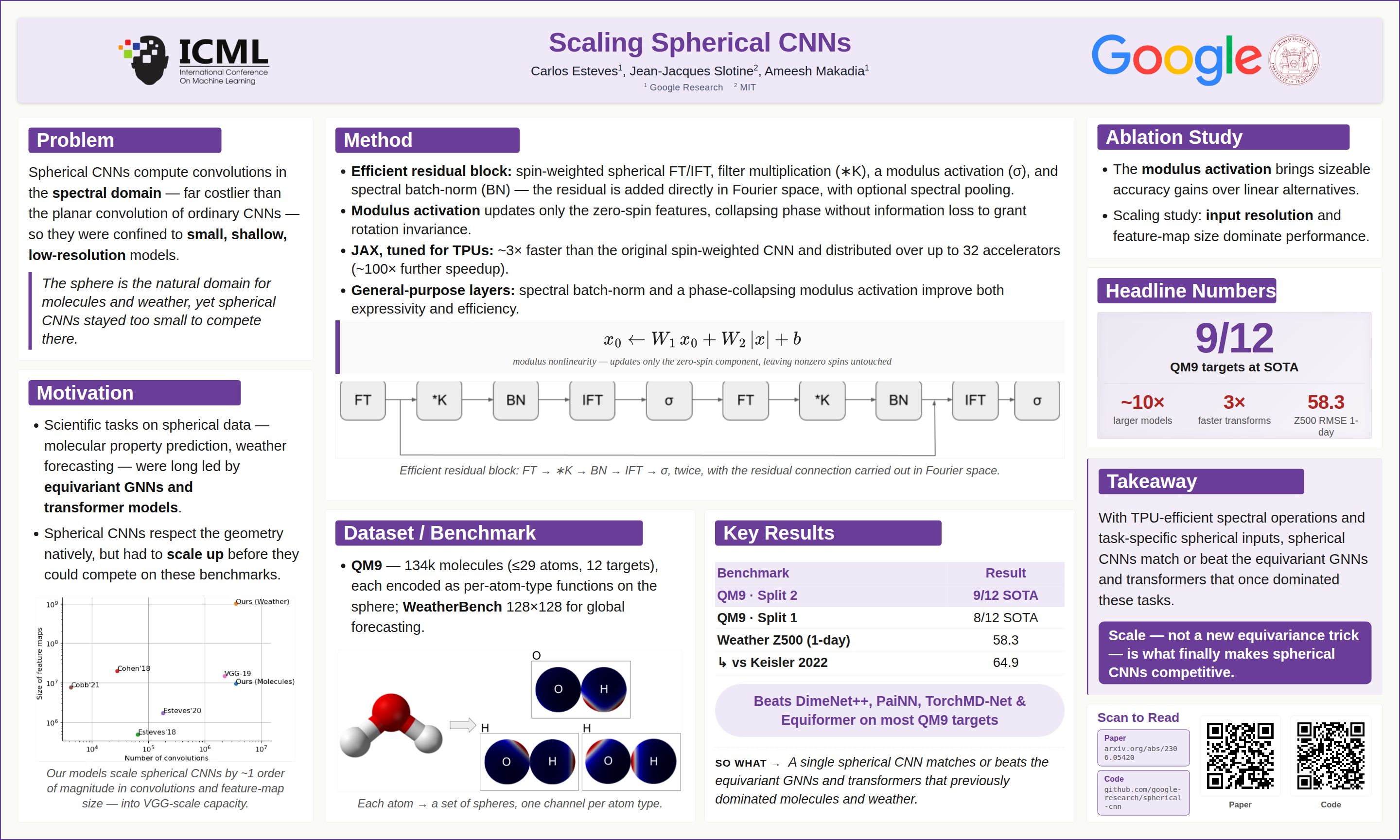}\end{minipage}\par\vspace{5pt}
\endgroup

\clearpage

\section{Single-Shot LLM Baseline Prompt}\label{app:llm-prompt}
The three single-shot LLM baselines in Table~\ref{tbl:main} (Claude-4.8 Opus, GPT-5.5, Gemini-3.1 Pro) share the one fixed prompt reproduced below; only the model identifier changes between runs, so the comparison isolates raw model capability rather than prompt engineering. Each model receives the paper's structured summary (the \texttt{paper\_spec.md} emitted by Paper2Assets, with audio-narration lines removed), its metadata, and the list of available figure filenames with captions, and must return a single self-contained A0-landscape poster HTML in one call: no fill loop and no multi-agent pipeline. That HTML is then rendered to A0 by the same \texttt{render\_poster.py} the full pipeline uses.

\paragraph{System prompt} (identical for all three models).
\begin{lstlisting}
You are an expert academic conference poster designer. Given a paper's structured summary and its figures, produce ONE complete, self-contained, print-ready HTML file for an A0 LANDSCAPE conference poster (1189mm wide by 841mm tall).

Hard requirements:
- Output ONLY the HTML, wrapped in a single ```html code block. No prose before or after.
- A0 landscape canvas: include `@page { size: 1189mm 841mm; margin: 0; }` and make the root container exactly 1189mm wide by 841mm tall.
- Clean multi-column academic layout (3 or 4 columns) with a prominent title bar (title, authors, institutes, venue), clearly separated titled sections, and large-format readable type (title very large; section headings roughly 30-40pt; body roughly 18-24pt at print scale).
- Faithfully present the paper: Problem/Motivation, Method, Key Results (WITH the exact numbers provided), and a Takeaway. Do NOT invent numbers, results, or citations beyond what is provided.
- Include the paper's figures using EXACTLY the provided filenames: `<img src="figures/FILENAME">`. Place each figure near the relevant section with a short caption. Never reference a figure filename that is not in the provided list.
- All CSS inline in one `<style>` block. No external stylesheets, no web fonts, no JavaScript. System-safe fonts only (Helvetica, Arial, Georgia, Times).
- Produce a polished, visually balanced poster suitable for a top-tier conference.
\end{lstlisting}

\paragraph{User prompt} (per paper; the \texttt{\{...\}} placeholders are filled from each paper's Paper2Assets bundle).
\begin{lstlisting}
TITLE: {title}
AUTHORS: {authors}
INSTITUTES: {institutes}
VENUE: {venue}

STRUCTURED SUMMARY (authoritative content; the numbers below are the real results to display):
{spec}

FIGURES AVAILABLE (use these exact filenames; place each with a short caption near the matching section):
{figs}

Produce the complete A0-landscape poster HTML now.
\end{lstlisting}

\end{document}